\documentclass{article}

\usepackage{microtype}
\usepackage{graphicx}
\usepackage{subfig}
\usepackage{booktabs} 

\usepackage{amssymb}
\usepackage{amsmath}
\usepackage{amsthm}

\usepackage{hyperref}

\usepackage[accepted]{icml2020}

\icmltitlerunning{Subspace Fitting Meets Regression}

\renewcommand{\vec}[1]{\mathbf{#1}}
\newcommand{\vecgreek}[1]{\boldsymbol{#1}}
\newcommand{\mtx}[1]{\mathbf{#1}}

\DeclareMathOperator*{\argmax}{arg\,max}
\DeclareMathOperator*{\argmin}{arg\,min}

\DeclareMathOperator{\Tr}{Tr}
\newcommand{\mtxtrace}[1]{\Tr\left\{{#1}\right\}}

\newtheorem{proposition}{Proposition}[section]
\newtheorem{corollary}{Corollary}[section]

\newtheorem{remark}{Remark}[section]

\theoremstyle{definition}
\newtheorem{definition}{Definition}[section]

\usepackage{color}

\usepackage{chngcntr}

\begin{document}

\twocolumn[
\icmltitle{Subspace Fitting Meets Regression: The Effects of Supervision and  Orthonormality Constraints on Double Descent of Generalization Errors}



\icmlsetsymbol{equal}{*}

\begin{icmlauthorlist}
\icmlauthor{Yehuda Dar}{Rice}
\icmlauthor{Paul Mayer}{Rice}
\icmlauthor{Lorenzo Luzi}{Rice}
\icmlauthor{Richard G. Baraniuk}{Rice}
\end{icmlauthorlist}

\icmlaffiliation{Rice}{ECE Department, Rice University, Houston, TX, USA}
\icmlcorrespondingauthor{Yehuda Dar}{ydar@rice.edu}

\icmlkeywords{Subspace fitting, regression, linear models, overparameterization, double descent, semi-supervised learning}

\vskip 0.3in
]



\printAffiliationsAndNotice{}  

\begin{abstract}
We study the linear subspace fitting problem in the \textit{overparameterized} setting, where the estimated subspace can perfectly \textit{interpolate} the training examples. 
Our scope includes the least-squares solutions to subspace fitting tasks with varying levels of supervision in the training data (i.e., the proportion of input-output examples of the desired low-dimensional mapping) and orthonormality of the vectors defining the learned operator.
This flexible family of problems connects standard, unsupervised subspace fitting that enforces strict orthonormality with a corresponding regression task that is fully supervised and does not constrain the linear operator structure.
This class of problems is defined over a \textit{supervision-orthonormality plane}, where each coordinate induces a problem instance with a unique pair of supervision level and softness of orthonormality constraints.
We explore this plane and show that the generalization errors of the corresponding subspace fitting problems follow \textit{double descent} trends as the settings become more supervised and less orthonormally constrained. 
\end{abstract}

\section{Introduction}
\label{sec:introduction}

Learning processes are naturally limited by the amount of data available for making inferences according to the chosen model. 
In particular, the interplay between the number of training examples and the complexity of the model (the number of parameters) is fundamental to successful learning in terms of generalization ability. 

\begin{figure}[ht]
\vskip 0.2in
\begin{center}
\centerline{\includegraphics[width=\columnwidth]{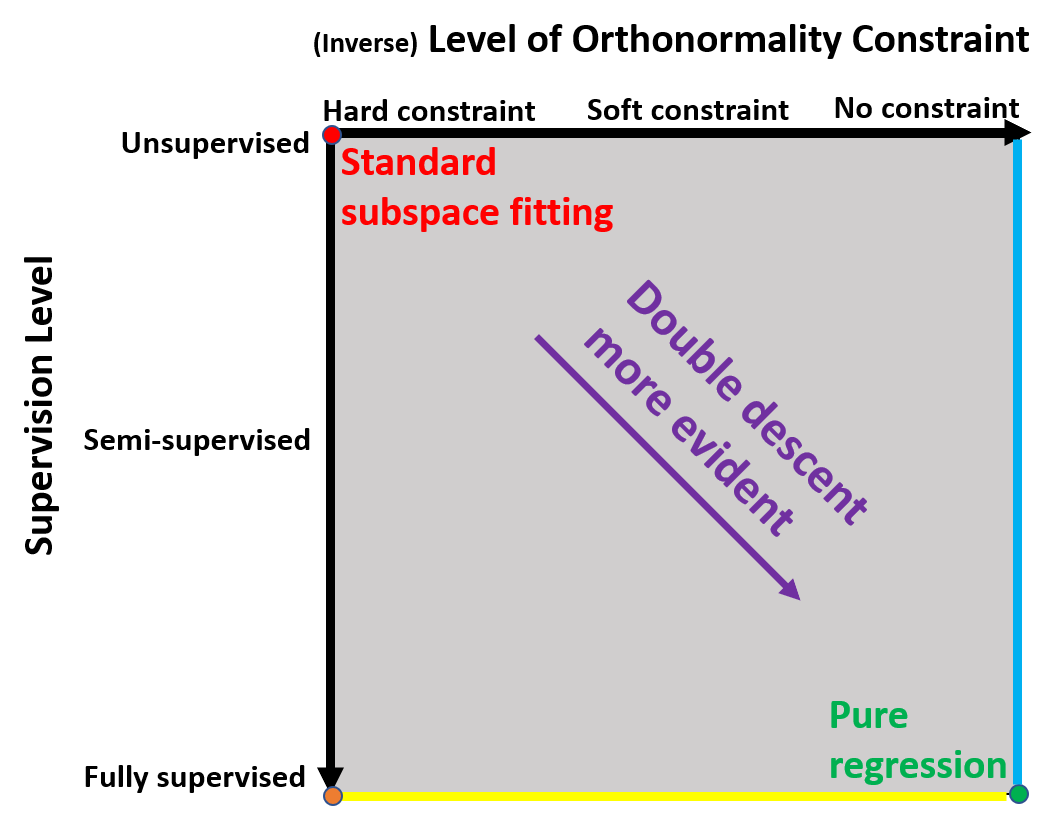}}
\caption{The supervision-orthonormality plane of subspace fitting problems.}
\label{fig:The supervision-orthonormality plane}
\end{center}
\vskip -0.3in
\end{figure}

The classical problem of linear regression, where one learns a linear mapping from a given set of input-output pairs, has been addressed for many years from the bias-variance tradeoff perspective. This design approach requires the number of parameters of the learned mapping to be sufficiently high, to avoid errors due to model bias, yet sufficiently low, to prevent overfitting to the training data. The established guideline \cite{breiman1983many} is that highly parameterized models, which lead to very low (or even zero) training error, are bad design choices that result in poor generalization performance.

The incredible success of highly overparameterized, deep neural networks has recently revived scientific interest in understanding the generalization errors induced by overparameterized models that are learned without explicit regularization mechanisms. One such prominent research line in \cite{spigler2018jamming,geiger2019scaling,belkin2019reconciling} shows that the generalization error actually decreases as the learned model is more overparameterized, even though all such models perfectly interpolate the training data (i.e., have zero training error). This generalization-error behavior (as a function of the number of model parameters) has been termed \textit{double descent}, due to the second decrease in the generalization error after entering the range of interpolating models. This finding has motivated an impressive series of mathematical studies, e.g., \cite{belkin2019two,hastie2019surprises,xu2019number,mei2019generalization} that formulate the double-descent phenomenon for the least-squares forms of various linear regression problems. 

In this paper, we extend the study of overparameterized models to the realm of \textit{dimensionality reduction}. We begin by considering the standard \textit{subspace fitting problem}, where one estimates an underlying linear operator (in the form of a matrix with orthonormal columns) that generates a given set of noisy examples. For this unsupervised subspace fitting problem, we define and explore the meaning of interpolating solutions and their generalization errors. We show that, while overparameterization is beneficial, the generalization error follows a \textit{single descent} trend throughout the entire range of parameterization levels, differing from the double descent shape that appears in (fully supervised) regression problems. 

Pushing further, \textit{we bridge the tasks of subspace fitting and regression} using a flexible optimization framework that generates a family of learning problems, with member of the family aiming to recover the same underlying subspace under a different setting. Specifically, we develop a general problem structure with two adjustable aspects. The first is the \textit{supervision level}, referring to the relative proportion between input-output and input-only examples given for learning. This essentially covers the range of problems from unsupervised, through semi-supervised, to fully supervised. The second adjustable aspect involves the structure of the learned linear operator that characterizes the fitted subspace, specifically, the \textit{degree of orthonormality} required of the columns of the estimated matrix. This provides a continuum of optimization forms, from unconstrained to strictly constrained, including intermediate settings with soft constraints. 

We interpret this entire class of problems as residing over a \textit{supervision-orthonormality plane} (see Fig.~\ref{fig:The supervision-orthonormality plane}), where each coordinate instantiates a distinct problem with its own coupled levels of supervision and orthonormality constraints. The two extreme, diagonal corners of the plane correspond to the standard subspace fitting and regression problems.

Since the non-standard problems on the  supervision-orthonormality plane do not have closed-form solutions, we explore them by developing iterative optimization procedures based on the projected gradient descent (PGD) technique. 
Interestingly, the soft orthonormality constraints reduce to thresholding operations applied to the singular values of the evolving solutions. Equipped with these PGD-based solutions, we empirically explore the generalization errors induced by the subspace estimation settings throughout the supervision-orthonormality plane. {\em Our results clearly demonstrate that the double-descent phenomenon emerges as the problems become increasingly supervised and less orthonormally constrained.}

\subsection{Related Work}
As explained above, our study directly relates to the recent research line on the double descent phenomenon \cite{belkin2019two,hastie2019surprises,xu2019number,mei2019generalization}. In addition, our study includes learning problems in settings that may resemble concepts available in the existing literature described next. 

In Section \ref{sec:unsupervised subspace learning} we address the linear subspace fitting problem via principal component analysis (PCA) that considers only a subset of coordinates of the given data vectors (as this design enables us to determine the number of parameters in the learned model, see Section \ref{subsec:parameterization level intro}). 
Interestingly, the study of dimensionality reduction mechanisms applied on a subset of the available input variables (or features) dates back to \cite{jolliffe1972discarding,jolliffe1973discarding}, where PCA was improved and/or made more computationally efficient. 
The approach of variable selection was developed further into sparse variable PCA methods, e.g., the transform-based preprocessing \cite{johnstone2009consistency} and expectation-maximization based approach \cite{ulfarsson2011vector}. These motivated corresponding studies of PCA in overparameterized settings under asymptotic assumptions, e.g., \cite{paul2007asymptotics,johnstone2009consistency,shen2016general}.

To motivate our supervised and semi-supervised settings in Sections \ref{sec:Supervised Subspace Fitting} and \ref{sec:Semi-Supervised Subspace Fitting}, we refer to \cite{yang2006semi}, where dimensionality reduction is improved using a subset of high-dimensional data examples and their corresponding exact low-dimensional representations. Beyond that, dimensionality reduction applied on multi-class data can be improved by supervised examples of class-labeled input data, e.g., \cite{sugiyama2006local,zhang2007semi,nie2010flexible}.

\subsection{Paper Outline}
This paper is organized as follows. In Section \ref{sec:Basic settings}, we describe the subspace fitting data model and its related definitions. 
In Section \ref{sec:unsupervised subspace learning}, we study the standard, unsupervised subspace fitting problem; this problem  lies at the red point in the supervision-orthonormality plane in Fig.~\ref{fig:The supervision-orthonormality plane}.
In Section \ref{sec:Supervised Subspace Fitting}, we explore a range of fully supervised learning problems that aim to recover the underlying subspace; 
these problems reside along the yellow axis of the supervision-orthonormality plane in Fig.~\ref{fig:The supervision-orthonormality plane} and include the green point of pure, standard regression.
In Section \ref{sec:Semi-Supervised Subspace Fitting}, we define a general optimization framework that supports any level of supervision and orthonormality constraint. This enables us to explore problems residing throughout the supervision-orthonormality plane. As two representative sets of problems, we evaluate the range of unconstrained settings (marked in blue in Fig.~\ref{fig:The supervision-orthonormality plane}) and the diagonal trajectory connecting the standard subspace fitting with pure regression (the direction of the purple arrow in Fig.~\ref{fig:The supervision-orthonormality plane}).
We conclude with a discussion of our findings in Section \ref{sec:conclusion}.
All proofs plus additional experimental details are provided in the Appendices included in the Supplementary Materials.

\section{Basic Settings}
\label{sec:Basic settings}
\subsection{Data Model}
\label{subsec:data model}

Consider a 
vector $\vec{x}\in\mathbb{R}^d$ that satisfies a noisy linear model in the form of 
\begin{equation} 
\label{eq:data model - x}
    \vec{x}={\mtx{U}}_{m} \vec{z} + \vec{\epsilon}
\end{equation}
where $\mtx{U}_m \in \mathbb{R}^{d\times m}$ is a matrix consisting of $m < d$ orthonormal column vectors $\lbrace \vec{u}^{(i)} \rbrace_{i=1}^{m} \in \mathbb{R}^{d}$ that span a rank-$m$ linear subspace. 
The underlying  $m$ coefficients, organized in $ \vec{z} \in \mathbb{R}^m $, are independent and identically distributed (i.i.d.) and standard Gaussian:  $ \vec{z} \sim \mathcal{N}\left( \vec{0}, \mtx{I}_{m} \right)$, where $\mtx{I}_{m}$ is the $m\times m$ identity matrix.  The random vector $\vec{\epsilon} \sim \mathcal{N} \left( \vec{0}, \sigma^{2}_{\epsilon} \mtx{I}_{d} \right) $, which is independent of $\vec{z}$, plays the role of a Gaussian noise vector in $\mathbb{R}^d$. Thus, $\vec{x}$ is zero mean with covariance matrix 
\begin{equation} 
\label{eq:data model - x - covariance matrix}
    \mtx{C}_{\vec{x}} = \mtx{U}_m \mtx{U}_m^T + \sigma^2_{\epsilon} \mtx{I}_d . 
\end{equation}

The problems in this paper are defined for learning settings where the data model (\ref{eq:data model - x}) is unknown, and only a dataset $\mathcal{D} \triangleq \left \{ \vec{x}^{(\ell)}\right\}_{\ell=1}^n \in \mathbb{R}^d$ of $n$ i.i.d.\ examples of data vectors satisfying (\ref{eq:data model - x}) is available. The vectors in $\mathcal{D}$ are centered with respect to their sample mean.
Note that $\mathcal{D}$, as defined here, enables unsupervised learning. In a later stage in the paper, where we discuss supervised and semi-supervised learning problems, the formulation of $\mathcal{D}$ will be extended.

\subsection{Learning Mappings with Desired Parameterization Levels}
\label{subsec:parameterization level intro}

Our interest is in learning tasks that infer mappings ${f:\mathbb{R}^d \rightarrow \mathbb{R}^{k}}$ to be applied on $d$-dimensional data and provide ${k}$-dimensional results, where $k <d$. The learned mapping is used for computing $f(\vec{x})$ for $\vec{x}\in\mathbb{R}^d$ realizing (\ref{eq:data model - x}) beyond the examples in $\mathcal{D}$. The common case of a very high dimension $d$ induces complex and highly parameterized instances of $f$, which are usually more difficult to learn. This challenge can be addressed by simplifying the learned mapping using the following design. 
A single set $\mathcal{S}$ of $p$ out of $d$ coordinates is determined arbitrarily (i.e., without any adaptation to the data). Specifically, $\mathcal{S}=\lbrace s_1, ..., s_p\rbrace$, where $1\leq s_1 < s_2 < ... < s_p \leq d$. The $p$-dimensional \textit{feature vector} of $\vec{x}\in\mathbb{R}^d$ is defined as $\vec{x}_{\mathcal{S}} \triangleq \left[ x_{s_1}, x_{s_2},...,x_{s_p} \right]^T$, where $x_{s_j}$ is the $s_j$-th component of $\vec{x}$. Then, the overall mapping is defined as 
$f(\vec{x})=f_{\mathcal{S}}(\vec{x}_{\mathcal{S}})$, where $f_{\mathcal{S}}: \mathbb{R}^p \rightarrow \mathbb{R}^{k}$ is a learned mapping that requires fewer parameters (than $f$) due to the lower dimension of its inputs. This simple approach lets us determine the actual number of parameters in the learned mappings by choosing the size of $\mathcal{S}$ (i.e., $p$). 

We consider procedures that learn $f_{\mathcal{S}}$ using only {$p$-dimensional} subvectors, specified by ${\mathcal{S}}$, of the vectors in $\mathcal{D}$ (a similar approach was used in \cite{belkin2019two} for non-asymptotic analysis of linear regression). Accordingly, the dataset of the $p$-dimensional feature vectors used for the learning process is denoted by $\mathcal{D}_{\mathcal{S}} \triangleq \left \{ \vec{x}^{(\ell)}_{\mathcal{S}} \right\}_{\ell=1}^n$. 

\section{Linear Subspace Fitting: The Standard, Unsupervised Setting}
\label{sec:unsupervised subspace learning}
\subsection{Problem Definition}
\label{subsec:unsupervised subspace learning - problem definition}
The goal is to find the linear subspace of rank $k$ that provides the best approximation ability, in the squared-error sense, of the data satisfying (\ref{eq:data model - x}). Recall that $m$, the true rank of the underlying linear part in (\ref{eq:data model - x}), is unknown and, hence, $k$ is not necessarily equal to $m$. The subspace estimate is formed based on the dataset $\mathcal{D}_{\mathcal{S}}$ of $p$-dimensional feature vectors. Nevertheless, the desired representation ability is for $d$-dimensional vectors beyond the dataset $\mathcal{D}$, namely, the out-of-sample squared error with respect to the data model in (\ref{eq:data model - x}) is the performance criterion of interest. 

A simple approach to address the subspace fitting problem, for $k\le p$, in a way conforming to the guidelines given in Section \ref{subsec:parameterization level intro}, is as follows. The first stage is to define the linear subspace $\widehat{\mathcal{U}}_{k,\mathcal{S}}$ that has rank $k$ and resides in $\mathbb{R}^p$, that minimizes the Euclidean distance between the data vectors in $\mathcal{D}_{\mathcal{S}}$ and their corresponding orthogonal projections onto the estimated subspace. Denote the $k$ orthonormal vectors spanning $\widehat{\mathcal{U}}_{k,\mathcal{S}}$ by $\widehat{\vec{u}}^{(1)}_{\mathcal{S}}, \dots, \widehat{\vec{u}}^{(k)}_{\mathcal{S}} \in \mathbb{R}^p$; organize them into the columns of a $p \times k$ matrix ${\widehat{\mtx{U}}_{k,\mathcal{S}}\triangleq \left[ \widehat{\vec{u}}^{(1)}_{\mathcal{S}},  \dots, \widehat{\vec{u}}^{(k)}_{\mathcal{S}}  \right]}$.
Note that, for $k < p$, $\widehat{\mtx{U}}_{k,\mathcal{S}}^T \widehat{\mtx{U}}_{k,\mathcal{S}} = \mtx{I}_{k}$, whereas $\widehat{\mtx{U}}_{k,\mathcal{S}}\widehat{\mtx{U}}_{k,\mathcal{S}}^T  \ne \mtx{I}_p$. 
Then, the closest point in $\widehat{\mathcal{U}}_{k,\mathcal{S}}$ to an arbitrary vector $\vec{v} \in \mathbb{R}^p$ is $\widehat{\vec{v}} =\widehat{\mtx{U}}_{k,\mathcal{S}}\widehat{\mtx{U}}_{k,\mathcal{S}}^T \vec{v}$. This produces the standard form of the subspace fitting problem, namely, 
\begin{align*} 
\label{eq:linear subspace fitting optimization}
    \widehat{\mtx{U}}_{k,\mathcal{S}} & =  \argmin_{\mtx{W}\in\mathbb{R}^{p\times k}:~ \mtx{W}^T \mtx{W} = \mtx{I}_{k}} \frac{1}{n} \sum_{\ell=1}^n \left \Vert{  \left({ \mtx{I}_{p} - \mtx{W} \mtx{W}^T }\right)\vec{x}^{(\ell)}_{\mathcal{S}}} \right\Vert _2^2
    \\
    & = \argmin_{\mtx{W}\in\mathbb{R}^{p\times k}:~ \mtx{W}^T \mtx{W} = \mtx{I}_{k}} \frac{1}{n} \left \Vert { \left({\mtx{I}_{p} - \mtx{W} \mtx{W}^T }\right) \mtx{X}_{\mathcal{S}} } \right \Vert _F^2
\end{align*}
where $ \mtx{X}_{\mathcal{S}} \triangleq \left[ \vec{x}^{(1)}_{\mathcal{S}} , \dots, \vec{x}^{(n)}_{\mathcal{S}} \right] \in \mathbb{R}^{p \times n}$ is the data matrix having the examples in $\mathcal{D}_{\mathcal{S}}$ as its columns. As is commonly known, the last optimization form is equivalent to  
\begin{equation} 
\label{eq:linear subspace fitting optimization - trace maximization form}
    \widehat{\mtx{U}}_{k,\mathcal{S}} = \argmax_{\mtx{W}\in\mathbb{R}^{p\times k}:~ \mtx{W}^T \mtx{W} = \mtx{I}_{k}} \mtxtrace{ \mtx{W}^T \mtx{X}_{\mathcal{S}} \mtx{X}_{\mathcal{S}}^T  \mtx{W} },
\end{equation}
which can be solved via a principal component analysis (PCA) procedure. Specifically, the orthonormal columns of $\widehat{\mtx{U}}_{k,\mathcal{S}}$ are the  eigenvectors corresponding to the first $k$ principal components of the sample covariance matrix induced by $\mathcal{D}_{\mathcal{S}}$. 

The learned rank-$k$ subspace $\widehat{\mathcal{U}}_{k,\mathcal{S}} \subset \mathbb{R}^p $ is extended to a rank-$k$ subspace $\widehat{\mathcal{U}}_{k}$ that resides in $\mathbb{R}^d$ and is spanned by $k$ orthonormal vectors, denoted as $\widehat{\vec{u}}^{(1)}, \dots, \widehat{\vec{u}}^{(k)} \in \mathbb{R}^d$. The suggested construction defines $\widehat{\vec{u}}^{(j)}$ (for $j=1,...,k$) such that its subvector corresponding to its coordinates in $\mathcal{S}$ is the learned $\widehat{\vec{u}}^{(j)}_{\mathcal{S}}$, and the rest of its $d-p$ components are zeros. Organizing these orthonormal vectors as the columns of a $d \times k$ matrix $\widehat{\mtx{U}}_{k}\triangleq \left[ \widehat{\vec{u}}^{(1)},  \dots, \widehat{\vec{u}}^{(k)} \right]$ provides a linear operator that, as required, creates $k$-dimensional representations for $d$-dimensional inputs. Namely, \begin{equation} 
\label{eq:unsupervised low dimensional representation}
    \widehat{\vec{v}} = \widehat{\mtx{U}}_{k}^T \vec{x}
\end{equation}
for $\vec{x}\in\mathbb{R}^d$ satisfying the data model (\ref{eq:data model - x}). 

Consider the case of $k=m$ and note that the unsupervised learning is defined to minimize $d$-dimensional reconstruction errors and, therefore, the columns of $\widehat{\mtx{U}}_{m}$ do not necessarily match in their indices to their closest columns of the true matrix ${\mtx{U}}_{m}$. Hence, the vector $\widehat{\vec{v}}$ is not a straightforward estimate of the underlying $\vec{z}\in\mathbb{R}^m$ that generates the given $\vec{x}$. 
This leads to the test error evaluation metric that is described next.

While $\widehat{\mtx{U}}_{k}$ is optimized to approximate the given sample $\mathcal{D}_{\mathcal{S}}$, the real interest is in representing arbitrary realizations of the model in (\ref{eq:data model - x}). Hence, the quality of $\widehat{\mtx{U}}_{k}$ should be evaluated for test data, $\vec{x}_{\rm test}\in\mathbb{R}^d$, randomly drawn from the probability distribution $P_\vec{x}$ induced by (\ref{eq:data model - x}). This provides the out-of-sample error of interest  
\begin{equation} 
\label{eq:linear subspace fitting - out of sample error}
\begin{split}
    & \mathcal{E}_{\rm out} ^{\rm unsup}  \left( \widehat{\mtx{U}}_{k} \right) \triangleq \mathbb{E} \left \Vert { \left ( \mtx{I}_d - \widehat{\mtx{U}}_{k} \widehat{\mtx{U}}_{k}^T \right ) \vec{x}_{\rm test}} \right \Vert _2^2 
    \\ 
    & = \mtxtrace{ \left( \mtx{I}_{d} - \widehat{\mtx{U}}_{k} \widehat{\mtx{U}}_{k}^T \right ) \mtx{C}_\vec{x} \left ( \mtx{I}_d - \widehat{\mtx{U}}_{k} \widehat{\mtx{U}}_{k}^T \right )^T } 
\end{split}
\end{equation}
where the expectation is for $\vec{x}_{\rm test} \sim P_\vec{x}$, and $\mtx{C}_\vec{x}$ is the covariance matrix from (\ref{eq:data model - x - covariance matrix}). Naturally, the formula for $\mathcal{E}_{\rm out} ^{\rm unsup} $ has an empirical counterpart defined for a set of test data vectors. 

Another metric useful for studying properties of learned subpaces is the in-sample approximation error of $\mathcal{D}$ 
\begin{align} 
\label{eq:linear subspace fitting - in sample error}
 &\mathcal{E}_{\rm in} ^{\rm unsup}  \left( \widehat{\mtx{U}}_{k} \right) \triangleq
 \\ \nonumber
 &\mtxtrace{ \left ( \mtx{I}_{d} - \widehat{\mtx{U}}_{k} \widehat{\mtx{U}}_{k}^T \right ) \widehat{\mtx{C}}_\vec{x}^{(n)} \left ( \mtx{I}_d - \widehat{\mtx{U}}_{k} \widehat{\mtx{U}}_{k}^T \right )^T }. 
\end{align}
Here $\widehat{\mtx{C}}_\vec{x}^{(n)}\triangleq \frac{1}{n} \mtx{X}\mtx{X}^T $ is the $d \times d$ sample covariance matrix corresponding to the $n$ examples provided in $\mathcal{D}$ (recall that the data is centered). 

Since the actual learning in the proposed construction of $\widehat{\mtx{U}}_{k}$ involves an actual learning only with respect to $\mathcal{D}_{\mathcal{S}}$, we define an additional in-sample approximation error as 
\begin{equation} 
\label{eq:linear subspace fitting - in sample error - actual learning}
\begin{split}
 &\mathcal{E}_{{\rm in},\mathcal{S}}^{\rm unsup}  \left( \widehat{\mtx{U}}_{k,\mathcal{S}} \right) \triangleq
 \\
 &\mtxtrace{ \left ( \mtx{I}_{p} - \widehat{\mtx{U}}_{k,\mathcal{S}} \widehat{\mtx{U}}_{k,\mathcal{S}}^T \right ) \widehat{\mtx{C}}_\vec{x,\mathcal{S}}^{(n)} \left ( \mtx{I}_p - \widehat{\mtx{U}}_{k,\mathcal{S}} \widehat{\mtx{U}}_{k,\mathcal{S}}^T \right )^T } 
 \end{split}
\end{equation}
where $\widehat{\mtx{C}}_\vec{x,\mathcal{S}}^{(n)}\triangleq \frac{1}{n} \mtx{X}_{\mathcal{S}}\mtx{X}_{\mathcal{S}}^T$ is a $p\times p$ sample-covariance matrix corresponding to $\mathcal{D}_{\mathcal{S}}$. 
Note that 
\begin{equation} 
\label{eq:linear subspace fitting - in sample error - relation}
 \mathcal{E}_{\rm in} ^{\rm unsup}  \left( \widehat{\mtx{U}}_{k} \right) = \mathcal{E}_{{\rm in},\mathcal{S}}^{\rm unsup}  \left( \widehat{\mtx{U}}_{k,\mathcal{S}} \right) + \frac{1}{n}\left \Vert \mtx{X}_{\mathcal{S}_c}  \right \Vert _F^2
\end{equation}
where ${\mathcal{S}_c \triangleq  \left\lbrace 1,...,d \right\rbrace \setminus \mathcal{S}}$ is the subset of coordinates excluded from the actual learning process, and $ {\mtx{X}_{\mathcal{S}_c} \triangleq \left[ \vec{x}^{(1)}_{\mathcal{S}_c} , \dots, \vec{x}^{(n)}_{\mathcal{S}_c} \right] \in \mathbb{R}^{(d-p) \times n}}$ includes the corresponding subvectors from the dataset as its columns. 
Accordingly, the term $\left \Vert \mtx{X}_{\mathcal{S}_c}  \right \Vert _F^2$ in (\ref{eq:linear subspace fitting - in sample error - relation}) is a quantity stemming from $\mathcal{S}$ and the number of parameters $p$, but independent of the specific subspace estimate. 

\subsection{Interpolating Subspaces}
\label{subsec:unsupervised subspace learning - interpolating subspaces}
We now turn to define two central concepts in our analysis. 
\theoremstyle{definition}
\begin{definition}
A subspace estimate $\widehat{\mathcal{U}}_{k}$, constructed based on the learning of $\widehat{\mathcal{U}}_{k,\mathcal{S}}$, is $\mathcal{S}$-\textit{interpolating} if $\mathcal{E}_{{\rm in},\mathcal{S}}^{\rm unsup}  \left( \widehat{\mtx{U}}_{k,\mathcal{S}} \right)=0$. 
\end{definition}
That is, an $\mathcal{S}$-interpolating subspace is able to perfectly represent the information embodied in $\mathcal{D}_{\mathcal{S}}$.

\theoremstyle{definition}
\begin{definition}
A subspace estimate $\widehat{\mathcal{U}}_{k}$, constructed based on learning $\widehat{\mathcal{U}}_{k,\mathcal{S}}$, is \textit{overparameterized} if ${p \in \lbrace {n+1},...,d \rbrace}$
and 
\textit{rank-overparameterized} if ${p \in \lbrace {n+1},...,d \rbrace}$ and $ {k  \in \lbrace n,...,p \rbrace}$. 
\end{definition}

\begin{remark}
A rank-overparameterized estimate of a subspace is also overparameterized.
\end{remark}

Recall that $\widehat{\mtx{C}}_\vec{x,\mathcal{S}}^{(n)}$ is a $p\times p$ matrix constructed from $n$ centered samples. 

\begin{corollary}
An overparameterized subspace estimate  $\widehat{\mathcal{U}}_{k}$ is formed based on a rank-deficient sample covariance matrix $\widehat{\mtx{C}}_\vec{x,\mathcal{S}}^{(n)}$ of rank $\rho \triangleq rank\left\lbrace \widehat{\mtx{C}}_\vec{x,\mathcal{S}}^{(n)} \right\rbrace \leq n-1$. If the subspace estimate $\widehat{\mathcal{U}}_{k}$ is also rank-overparameterized, then the rank-deficiency of $\widehat{\mtx{C}}_\vec{x,\mathcal{S}}^{(n)}$ affects $\widehat{\mathcal{U}}_{k}$. 
\end{corollary}

\begin{corollary}
\label{corollary:rank-overparameterized estimate construction}
A rank-overparameterized subspace estimate (of rank $k$) is spanned by the $\rho$ eigenvectors of $\widehat{\mtx{C}}_\vec{x,\mathcal{S}}^{(n)}$ corresponding to all the nonzero eigenvalues. The additional $k- \rho$ orthonormal vectors can be  arbitrarily chosen from the $p- \rho$ eigenvectors of $\widehat{\mtx{C}}_\vec{x,\mathcal{S}}^{(n)}$ that match to its zero eigenvalues.
\end{corollary}

\begin{remark}
Corollary \ref{corollary:rank-overparameterized estimate construction} provides a suggested construction for a rank-overparameterized subspace estimate. In general, the additional $k- \rho$ orthonormal vectors defined above can be any set spanning a rank-${(k- \rho)}$ subspace of the null space of the sample covariance $\widehat{\mtx{C}}_\vec{x,\mathcal{S}}^{(n)}$. 
\end{remark}

This means that the PCA procedure required for solving (\ref{eq:linear subspace fitting optimization - trace maximization form}) reduces to a significantly simpler task.
The following is proved in Appendix \ref{appsec:Proofs and Explanations for Section 3}. 

\begin{proposition}
\label{proposition:rank-overparameterization implies interpolation}
A rank-overparameterized subspace estimate is also an $\mathcal{S}$-interpolating subspace. 
\end{proposition}

\subsection{Generalization Error vs.\ Parameterization Level}

We now turn to characterize the benefits of overparameterized solutions to the unsupervised subspace fitting problem. 

\begin{proposition}
\label{proposition:unsupervised out-of-sample formula}
The out-of-sample error (\ref{eq:linear subspace fitting - out of sample error}) can be expressed as
\begin{equation} 
\label{eq:linear subspace fitting - out of sample error - theorem formula}
    \mathcal{E}_{\rm out} ^{\rm unsup}  \left( \widehat{\mtx{U}}_{k} \right) = \sum_{i=1}^{d} {\lambda^{(i)}} - \sum_{i\in\widehat{\mathcal{S}}_{\rm{max}}^{(k)}}{ \sum_{j=1}^{p} {\lambda^{(j)}_{\mathcal{S}}} 
    \left| \left\langle {\vecgreek{\psi}}_{\mathcal{S}}^{(j)}, \widehat{\vecgreek{\psi}}_{\mathcal{S}}^{(i)} \right\rangle \right|^2 }
\end{equation}
where ${\lambda^{(i)}}$ is the $i^{th}$ eigenvalue of ${\mtx{C}}_\vec{x}$, the eigenvalues $\left\{ {{\lambda^{(j)}_{\mathcal{S}}}}\right\}_{j=1}^{p}$ and eigenvectors $\left\{ {{\vecgreek{\psi}}_{\mathcal{S}}^{(j)} }\right\}_{j=1}^{p}$ correspond to the true covariance matrix of the $p$-dimensional feature vectors ${\mtx{C}}_{\vec{x},\mathcal{S}}$, and  $\widehat{\vecgreek{\psi}}_{\mathcal{S}}^{(j)}$ is the $j^{th}$ eigenvector of the sample covariance  $\widehat{\mtx{C}}_\vec{x,\mathcal{S}}^{(n)}$. Also,  $\widehat{\mathcal{S}}_{\rm{max}}^{(k)}$ is the set of indices corresponding to the $k$ maximal eigenvalues of $\widehat{\mtx{C}}_\vec{x,\mathcal{S}}^{(n)}$. 
\end{proposition}

\begin{remark}
In case the subspace estimate is rank-overparameterized, then the definition of $\widehat{\mathcal{S}}_{\rm{max}}^{(k)}$ in Proposition \ref{proposition:unsupervised out-of-sample formula} assumes the construction suggested in Corollary \ref{corollary:rank-overparameterized estimate construction}. This means that when $k>\rho$,  the set $\hat{S}_{\rm max}^{(k)}$ includes $k-\rho$ indices that correspond to $k-\rho$ out of the $p-\rho$ zero eigenvalues of the sample covariance matrix $\widehat{\mtx{C}}_\vec{x,\mathcal{S}}^{(n)}$. 
\end{remark}

There are two axes along which to study how $\mathcal{E}_{\rm out} ^{\rm unsup}  \left( \widehat{\mtx{U}}_{k} \right)$ decays:  along $k$ and along $p$.
For $k$, we can state the following (see the proof in Appendix \ref{appsec:Proofs and Explanations for Section 3}).

\begin{proposition}
\label{proposition:unsupervised out-of-sample error reduces with k}
A subspace estimate induces an out-of-sample error $\mathcal{E}_{\rm out} ^{\rm unsup}  \left( \widehat{\mtx{U}}_{k} \right)$  that monotonically decreases as ${k  \in \lbrace 1,...,p \rbrace}$ increases and $\widehat{\mtx{U}}_{k}$ is gradually extended.
\end{proposition}

For $p$, the situation is more delicate.  A rigorous proof has so far eluded us, possibly due to our non-asymptotic setting that hinders the important characterization of the sample covariance eigenvectors (e.g., as provided in the asymptotic frameworks in \cite{paul2007asymptotics,shen2016general}). Yet, the results of extensive simulations indicate that, on average with respect to $\mathcal{S}$ that is uniformly chosen at random,  $\mathcal{E}_{\rm out} ^{\rm unsup}  \left( \widehat{\mtx{U}}_{k} \right)$ decays monotonically in $p$ as well (see Fig.~\ref{fig:unsupervised - out-of-sample errors} and the additional results provided in Appendix \ref{appsec:Proofs and Explanations for Section 3}).

To summarize what we have learned so far, increased overparameterization and/or rank-overparameterization of unsupervised subspace estimates provide lower generalization errors. Moreover, the overall trend induced by increasing the number of features, $p$, significantly differs from the double-descent behavior arising in regression problems (see, e.g., \cite{belkin2019two}).

\begin{figure}[t]
\begin{center}
\subfloat[]{\includegraphics[width=0.75\columnwidth]{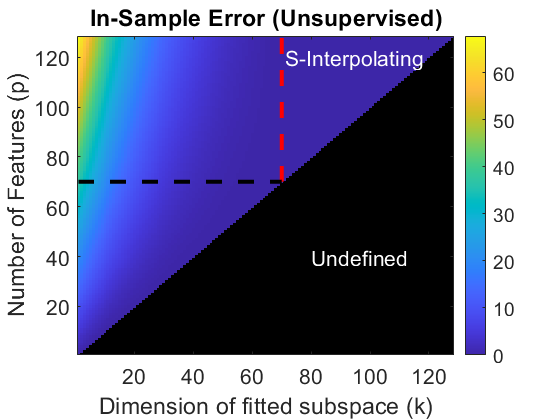}\label{fig:unsupervised - in-sample errors}}\\
\subfloat[]{\includegraphics[width=0.75\columnwidth]{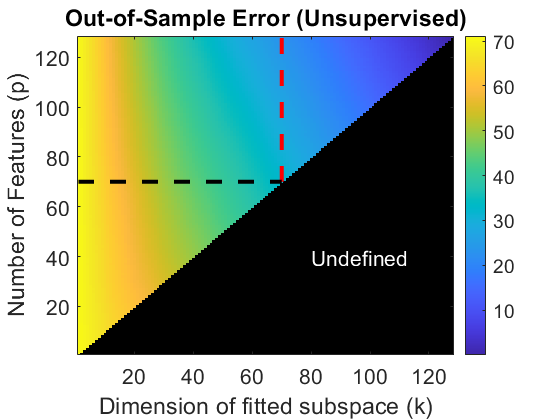} \label{fig:unsupervised - out-of-sample errors}}
\caption{Evaluation of unsupervised learning  at various parameterization settings. (a) The in-sample errors, $\mathcal{E}_{\rm in} ^{\rm unsup}  \left( \widehat{\mtx{U}}_{k} \right)$. (b)~The out-of-sample errors, $\mathcal{E}_{\rm out} ^{\rm unsup}  \left( \widehat{\mtx{U}}_{k} \right)$. 
The border lines of the overparamaeterization and rank-overparameterization regions are marked with black and red dashed lines, respectively.}
\end{center}
\vspace*{-5mm}
\end{figure}

\subsection{Empirical Demonstrations}

We now present results for unsupervised learning settings, where $d=128$, $n=70$, and the $m=40$ columns of $\mtx{U}_m$ are set as the first $40$ normalized columns of the Hadamard matrix of order $128$. Figure \ref{fig:unsupervised - in-sample errors} shows the in-sample error, $\mathcal{E}_{{\rm in},\mathcal{S}}^{\rm unsup}  \left( \widehat{\mtx{U}}_{k,\mathcal{S}} \right)$, obtained for the various parameterization combinations of $p$ and $k$ (recall that $k\le p$, and this is the reason for the undefined regions in
Figs.~\ref{fig:unsupervised - in-sample errors}--\ref{fig:unsupervised - out-of-sample errors}). 
Figure \ref{fig:unsupervised - out-of-sample errors}  demonstrates the out-of-sample errors, $\mathcal{E}_{\rm out} ^{\rm unsup}  \left( \widehat{\mtx{U}}_{k} \right)$, that are empirically evaluated using a test set of 1000 out-of-sample realizations of data vectors $\vec{x}$ satisfying (\ref{eq:data model - x}). The border lines of the overparamaeterization and rank-overparameterization regions are marked with black and red dashed lines, respectively. The monotonic decrease of the out-of-sample error with the increase in $p$ and/or $k$ is evident (see Fig. \ref{fig:unsupervised - out-of-sample errors}). The fact that rank-overparameterization induces $\mathcal{S}$-interpolating subspace estimates is also visible in Fig.~\ref{fig:unsupervised - in-sample errors}.

\section{Supervised Subspace Fitting}
\label{sec:Supervised Subspace Fitting}

The previous section demonstrated the behavior of the generalization error with respect to the number of features $p$ for the unsupervised subspace fitting setting.
We now turn to define \textit{fully supervised} forms that are related to the above defined problem (and reside along the bottom, yellow-colored border line of the supervision-orthonormality plane in Fig.~\ref{fig:The supervision-orthonormality plane}). Our main goal is to study how the aspects of supervision and constraints affect the trends of generalization errors observed for the unsupervised setting. 

The data model remains the same as in Section \ref{subsec:data model}. The only exception, here, is that the provided dataset is ${\mathcal{D}^{\rm sup}  \triangleq \left \{ \left( \vec{x}^{(\ell)},\vec{z}^{(\ell)} \right) \right\}_{\ell=1}^n \in \mathbb{R}^d \times \mathbb{R}^m}$ of $n$ i.i.d.\ samples of $(\vec{x},\vec{z})$ pairs satisfying (\ref{eq:data model - x}). Note that the examples given for the low-dimensional representations $\vec{z}$ reflect the true dimension of the linear subspace underlying  the noisy data. Hence, the learning is to be defined for establishing a mapping that provides $m$-dimensional representations. This contrasts the unsupervised case, where $m$ is unknown and, thus, the assumed low-dimension $k$ is possibly incorrect.

\subsection{Supervised Learning with Strict Orthonormality Constraints}

This subsection examines the problem induced at the lower-left corner of the supervision-orthonormality plane (see orange-colored coordinate in Fig.~\ref{fig:The supervision-orthonormality plane}). 
We employ the approach described in Section \ref{subsec:parameterization level intro} for setting a parameterization level of interest. Again, the subset of $p$ coordinates specified in $\mathcal{S}$ is used to subsample  the $\vec{x}$ vectors, corresponding to the data elements that the learned mapping should be applied on. Note that the $\vec{z}$ vectors remain in their full forms. Accordingly, the dataset used for the supervised learning is $\mathcal{D}^{\rm sup} _{\mathcal{S}} \triangleq \left \{ \left( \vec{x}^{(\ell)}_{\mathcal{S}},\vec{z}^{(\ell)} \right) \right\}_{\ell=1}^n \in \mathbb{R}^p \times \mathbb{R}^m$, where $p\ge m$. The optimization problem for establishing the orthonormal set of $m$ vectors spanning the subspace is 
\begin{align} 
\label{eq:linear subspace fitting optimization - supervised with constraints}
    \widehat{\mtx{U}}_{m,\mathcal{S}} & =  \argmin_{\mtx{W}\in\mathbb{R}^{p\times m}:~ \mtx{W}^T \mtx{W} = \mtx{I}_{m}} \frac{1}{n} \left \Vert  \mtx{W}\mtx{Z} - \mtx{W} \mtx{W}^T \mtx{X}_{\mathcal{S}}  \right \Vert _F^2
    \nonumber \\
    & =  \argmin_{\mtx{W}\in\mathbb{R}^{p\times m}:~ \mtx{W}^T \mtx{W} = \mtx{I}_{m}} \frac{1}{n} \left \Vert  \mtx{Z} - \mtx{W}^T \mtx{X}_{\mathcal{S}}  \right \Vert _F^2
\end{align}
where $ {\mtx{X}_{\mathcal{S}} \triangleq \left[ \vec{x}^{(1)}_{\mathcal{S}} , \dots, \vec{x}^{(n)}_{\mathcal{S}} \right] \in \mathbb{R}^{p \times n}}$ and $ {\mtx{Z} \triangleq \left[ \vec{z}^{(1)} , \dots, \vec{z}^{(n)} \right] \in \mathbb{R}^{m \times n}}$. 

The optimization problem in (\ref{eq:linear subspace fitting optimization - supervised with constraints}) is related to the orthonormal Procrustes problem \cite{gower2004procrustes}. However, here the optimization variable is a rectangular, instead of a square, matrix and therefore we do not have a closed-form solution. This motivates us to address (\ref{eq:linear subspace fitting optimization - supervised with constraints}) by a projected gradient descent approach (see Algorithm \ref{alg:Supervised Subspace Fitting via Projected Gradient Descent}, where $t$ is the iteration index, $\mu$ is the gradient step size, and $T_{\rm hard}$ is defined next). 

In this case, the constraint-projection stage reduces to an operator applied on the singular values of the evolving solution. Specifically, consider a matrix $\mtx{W}^{(\text{\rm in} )}\in\mathbb{R}^{p\times m}$ (where $p\ge m$), with the SVD $\mtx{W}^{(\text{\rm in} )}=\mtx{\Omega} \mtx{\Sigma}^{({\rm in})} \mtx{\Theta}^{T}$, where $\mtx{\Omega}$ and $\mtx{\Theta}$ are $p\times p$ and $m\times m$ real orthonormal matrices, respectively, and $\mtx{\Sigma}^{({\rm in})}$ is a $p\times m$ real diagonal matrix with $m$ singular values $\lbrace{ \sigma_{i}\left( \mtx{W}^{(\text{\rm in} )} \right) }\rbrace_{i=1}^{m}$ on its main diagonal. Then, projecting $\mtx{W}^{(\text{\rm in} )}$ onto the hard-orthonormality constraint via
\begin{equation} 
\label{eq:hard ortho constraint projection - optimization problem form}
{\mtx{W}^{(\text{\rm out} )}} = \argmin_{\mtx{W}\in\mathbb{R}^{p\times m}:~ \mtx{W}^T \mtx{W} = \mtx{I}_{m}} \left \Vert  \mtx{W} - {\mtx{W}^{(\text{\rm in} )}} \right \Vert _F^2
\end{equation}
induces the mapping  ${\mtx{W}^{(\text{\rm out} )}}\triangleq T_{\text{hard}}\left({\mtx{W}^{(\text{\rm in} )}}\right)$, where ${\mtx{W}^{(\text{\rm out} )}}=\mtx{\Omega} \mtx{\Sigma}^{({\rm out})} \mtx{\Theta}^{T}$ and the singular values along the main diagonal of $\mtx{\Sigma}^{({\rm out})}$ are ${\sigma_{i}\left( \mtx{W}^{(\text{\rm out} )} \right)=1}$ for ${i=1,\dots,m}$. See Appendix \ref{appsec:Proofs and Additional Details for Section 4} for the proof. 

\begin{algorithm}[t]
   \caption{Supervised Subspace Fitting via Projected Gradient Descent: Strict Orthonormality Constraints}
   \label{alg:Supervised Subspace Fitting via Projected Gradient Descent}
\begin{algorithmic}
   \STATE {\bfseries Input:} dataset $\mathcal{D}^{\rm sup} _{\mathcal{S}} = \left \{ \left( \vec{x}^{(\ell)}_{\mathcal{S}},\vec{z}^{(\ell)} \right) \right\}_{\ell=1}^n$ and a coordinate subset $\mathcal{S}$
   \STATE {\bfseries Initialize} $\mtx{W}^{(t=0)}= T_{\text{hard}}\left( \left( \mtx{Z} \mtx{X}_{\mathcal{S}}^{+} \right)^T \right)$, $t=0$
   \REPEAT
   \STATE $t \leftarrow t+1$
   \STATE $\mtx{Y}^{(t)} = \mtx{W}^{(t-1)} - \mu \mtx{X}_{\mathcal{S}} \left(  \left(\mtx{W}^{(t-1)}\right)^T \mtx{X}_{\mathcal{S}} - \mtx{Z}\right)^T$
   \STATE $\mtx{W}^{(t)} = T_{\text{hard}}\left( { \mtx{Y}^{(t)} } \right) $ 
   \UNTIL{stopping criterion is satisfied}
   \STATE Set $\widehat{\mtx{U}}_{m,\mathcal{S}} = \mtx{W}^{(t)}$
   \STATE Create $\widehat{\mtx{U}}_{m}$ based on $\widehat{\mtx{U}}_{m,\mathcal{S}}$ and zeros at rows correponding to $\mathcal{S}_c$
   \STATE {\bfseries Output:} $\widehat{\mtx{U}}_{m}$
   
\end{algorithmic}
\end{algorithm}

Unlike the unsupervised settings in Section \ref{sec:unsupervised subspace learning}, the supervised learning procedures defined here provide estimates $\widehat{\mtx{U}}_{m}$ that approximate the mapping from $\vec{x}\in\mathbb{R}^d$ to $\vec{z}\in\mathbb{R}^m$. This enables us to define the following \textit{supervised} evaluation metrics, considering the in-sample squared error (with respect to the dataset $\mathcal{D}^{\rm sup} _{\mathcal{S}}$)
\begin{equation} 
\label{eq:linear subspace fitting - supervised - in sample error}
 \mathcal{E}_{\rm in} ^{\rm sup}  \left( \widehat{\mtx{U}}_{m} \right) \triangleq \frac{1}{n} \sum_{\ell=1}^{n} {\left\Vert { \vec{z}^{(\ell)} - \widehat{\mtx{U}}_{m}^T \vec{x}^{(\ell)} }\right\Vert _2^2} 
\end{equation}
and the out-of-sample squared error
\begin{equation} 
\label{eq:linear subspace fitting - supervised - out of sample error}
 \mathcal{E}_{\rm out} ^{\rm sup}  \left( \widehat{\mtx{U}}_{m} \right) \triangleq \mathbb{E} \left\Vert {  \vec{z}_{\rm test} - \widehat{\mtx{U}}_{m}^T \vec{x}_{\rm test} }\right\Vert _2^2 
\end{equation}
where the expectation is over $\left( \vec{x}_{\rm test}, \vec{z}_{\rm test}\right) \sim P_{\vec{x},\vec{z}}$ as induced by (\ref{eq:data model - x}). 

\begin{figure}[t]
\centering
\subfloat[]{\includegraphics[width=0.8\columnwidth]{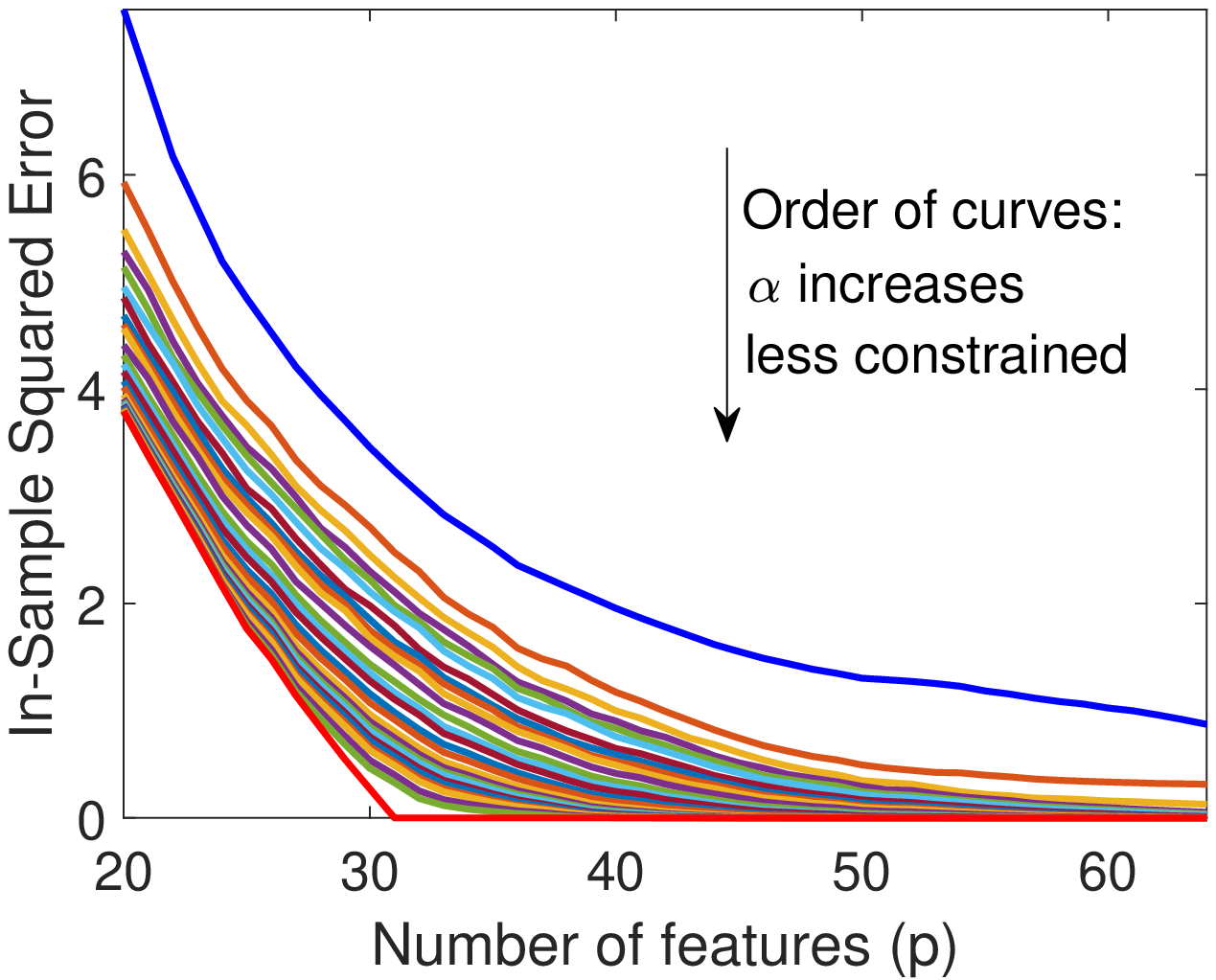}\label{fig:fully supervised with soft ortho constraints - in-sample errors - AVERAGE}}
\\
\subfloat[]{\includegraphics[width=0.8\columnwidth]{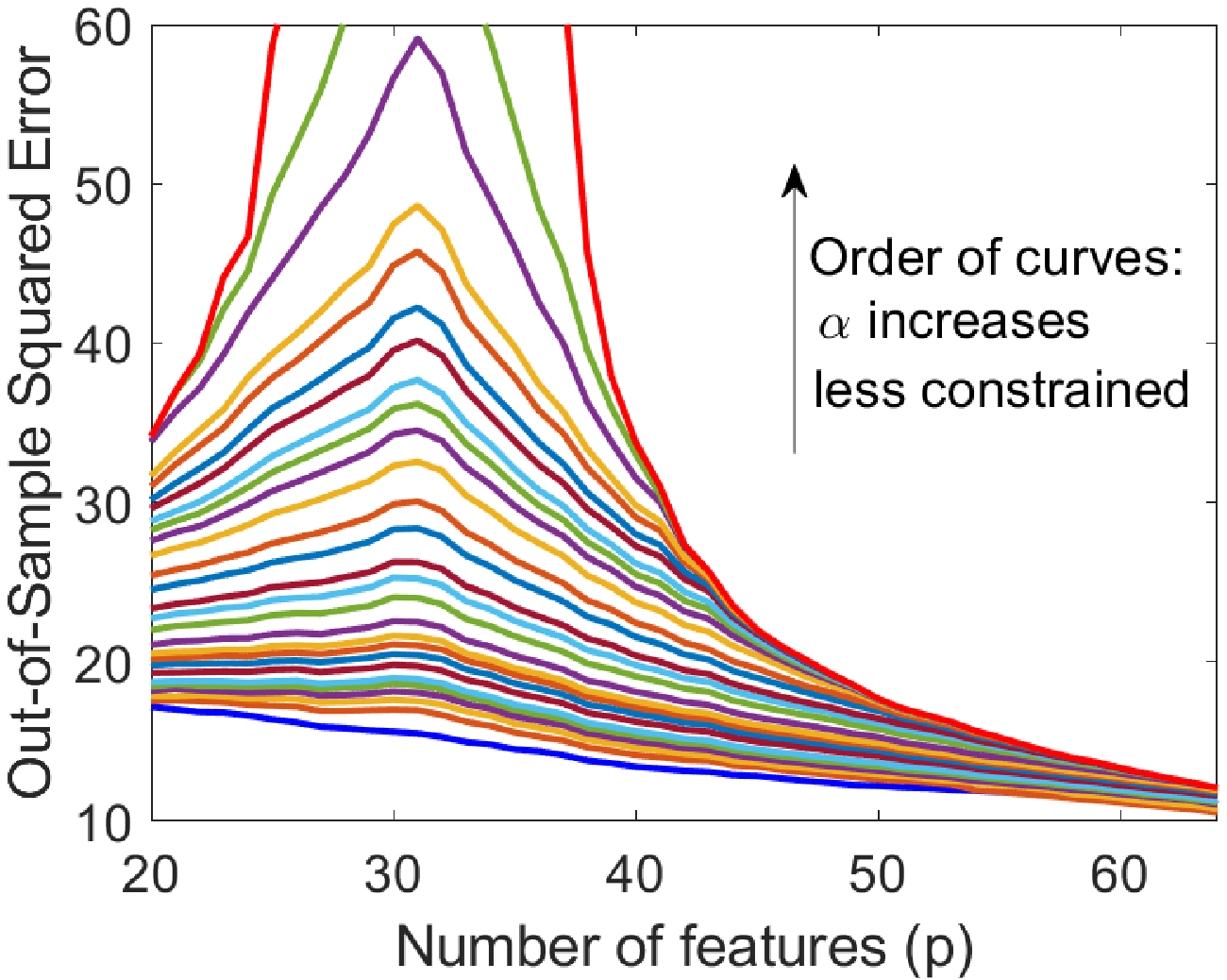}\label{fig:fully supervised with soft ortho constraints - out-of-sample errors - AVERAGE}}
\caption{The \textbf{(a) in-sample}  errors $\mathcal{E}_{\rm in} ^{\rm sup}  \left( \widehat{\mtx{U}}_{m} \right)$ and \textbf{(b) out-of-sample}  errors $\mathcal{E}_{\rm out} ^{\rm sup}  \left( \widehat{\mtx{U}}_{m} \right)$ of fully-supervised learning versus the number of parameters $p$. The errors are \textbf{averaged} over 10 experiments with different sequential orders of adding coordinates to $\mathcal{S}$. Here $d=64$, $m=20$ and $n=32$. Each curve presents the results for a different level $\alpha$ of orthonormality constraints. The results here correspond to problems located along the yellow-colored border line in Fig.~\ref{fig:The supervision-orthonormality plane}. The colors of the curves in this figure are arbitrary and not related to the colors in Fig.~\ref{fig:The supervision-orthonormality plane}.}
\label{fig:fully supervised with soft ortho constraints}
\vskip -0.1in
\end{figure}

Our results (see the bottom blue-colored curve of out-of-sample errors in Fig.~\ref{fig:fully supervised with soft ortho constraints - out-of-sample errors - AVERAGE} and Appendix \ref{appsec:Proofs and Additional Details for Section 4} for more details) show that there is no double-descent behavior in this setting, despite the fact the learning is fully supervised. Moreover, the corresponding in-sample error curve (see the upper blue-colored curve in Fig.~\ref{fig:fully supervised with soft ortho constraints - in-sample errors - AVERAGE}) shows that, under strict orthonormality constraints, interpolation is not achieved, even not by solutions corresponding to $p>n$. 

\subsection{The Regression Approach: A Supervised, Unconstrained Setting}

The problem defined in (\ref{eq:linear subspace fitting optimization - supervised with constraints}) recalls the usual regression form, except for the constraint on the matrix estimate. This motivates us to extend the range of problems we consider to include a standard regression problem for the purpose of estimating $\mtx{U}_m$ without constraining its structure. This problem is located at the green coordinate in the corner of the supervision-orthonormality plane in Fig.~\ref{fig:The supervision-orthonormality plane}. This setting is simply obtained by removing the constraint from (\ref{eq:linear subspace fitting optimization - supervised with constraints}), namely, 
\begin{equation} 
\label{eq:linear subspace fitting optimization - supervised unconstrained}
    \widehat{\mtx{U}}_{m,\mathcal{S}} =  \argmin_{\mtx{W}\in\mathbb{R}^{p\times m}} \frac{1}{n} \left \Vert  \mtx{Z} - \mtx{W}^T \mtx{X}_{\mathcal{S}}  \right \Vert _F^2
\end{equation}
which has a closed-form solution  $  \widehat{\mtx{U}}_{m,\mathcal{S}} = \left( \mtx{Z} \mtx{X}_{\mathcal{S}}^{+} \right)^T $, 
where $\mtx{X}_{\mathcal{S}}^{+}$ is the pseudoinverse of $\mtx{X}_{\mathcal{S}}$. Similar to the previous settings, the matrix $\widehat{\mtx{U}}_{m}$ is formed based on $\widehat{\mtx{U}}_{m,\mathcal{S}}$ in addition to zeros at the rows corresponding to indices in $\mathcal{S}_c$. Again, the relevant evaluation metrics are $\mathcal{E}_{\rm in} ^{\rm sup}  \left( \widehat{\mtx{U}}_{m} \right)$ and $\mathcal{E}_{\rm out} ^{\rm sup}  \left( \widehat{\mtx{U}}_{m} \right)$ as defined in (\ref{eq:linear subspace fitting - supervised - in sample error}) and (\ref{eq:linear subspace fitting - supervised - out of sample error}), respectively. 

Note that in this setting, which does not include strict orthonormality constraints on the columns of $\widehat{\mtx{U}}_{m}$, one can construct estimates also for $p < m$. However, since our scope includes also problems with strict or soft orthonormality constraints, all the results in this paper are presented only for $p \ge m$. 

Our results (see the upper red-colored curve in Fig.~\ref{fig:fully supervised with soft ortho constraints - out-of-sample errors - AVERAGE} and Appendix \ref{appsec:Proofs and Additional Details for Section 4} for more details) demonstrate that the generalization error follows a double-descent behavior. Note that the ``first descent" in the underparameterized range is missing due to the constructions from Section \ref{subsec:parameterization level intro} (this is also the case in \cite{belkin2019two}). 
The corresponding in-sample error curve (see the bottom red-colored curve in Fig.~\ref{fig:fully supervised with soft ortho constraints - in-sample errors - AVERAGE}) shows that all the \textit{unconstrained} overparameterized solutions interpolate, i.e., zero in-sample error is achieved for $p\ge n-1$ (this range is defined by $n-1$ and not $n$ due to data centering).
This specific result is a consequence of the pure regression setting we examine in this subsection. In our next steps below we explore settings that are not standard regression problems and, for them, studying the existence of double descent phenomena is of interest.

\subsection{Supervised Learning with Soft Orthonormality Constraints}

The two supervised problems defined in (\ref{eq:linear subspace fitting optimization - supervised with constraints}) and (\ref{eq:linear subspace fitting optimization - supervised unconstrained}) correspond to the extreme cases of strict orthonormality constraints and no constraints at all, respectively. We observed that, while the unconstrained problem yields generalization errors following the double-descent behavior, the strictly constrained problem does not (despite the fact it is also fully supervised). This motivates us to explore the entire range of supervised problems connecting (\ref{eq:linear subspace fitting optimization - supervised with constraints}) and (\ref{eq:linear subspace fitting optimization - supervised unconstrained}) via orthonormality constraints that can be progressively softened. This range of problems is denoted by the yellow line in Fig.~\ref{fig:The supervision-orthonormality plane}. 

The following constructions rely on the fact that a tall (rectangular) matrix has orthonormal columns if and only if all of its singular values equal 1. This statement is proved in Appendix \ref{appsec:Proofs and Additional Details for Section 4}. Accordingly, we formulate the soft-constraint problem (for $p\ge m$) as  
\begin{align} 
\label{eq:linear subspace fitting optimization - supervised with soft constraints}
    &\widehat{\mtx{U}}_{m,\mathcal{S}} =   \argmin_{\mtx{W}\in\mathbb{R}^{p\times m}} \frac{1}{n} \left \Vert  \mtx{Z} - \mtx{W}^T \mtx{X}_{\mathcal{S}}  \right \Vert _F^2
    \\ \nonumber
    &\text{subject to}~\lvert{  \sigma_{i}^{2}\left( \mtx{W} \right) - 1 }\rvert \le \alpha ~~\text{for }i=1,...,m
\end{align}
where $ \sigma_{i}\left( \mtx{W} \right)$ is the $i^{th}$ singular value of $\mtx{W}$, and the constant $\alpha\ge 0$ defines the softness of the constraints. 
Note that for $\alpha = 0$ the demand becomes a hard constraint of orthonormality and, then, (\ref{eq:linear subspace fitting optimization - supervised with soft constraints}) reduces to (\ref{eq:linear subspace fitting optimization - supervised with constraints}). When $\alpha \rightarrow \infty$ the problem converges to the unconstrained regression form of (\ref{eq:linear subspace fitting optimization - supervised unconstrained}).

Due to the constraints, the problem (\ref{eq:linear subspace fitting optimization - supervised with soft constraints}) does not have a closed-form solution. Hence, we propose again a procedure based on the projected gradient descent technique.  Nicely, the constraint-projection step takes the form of a thresholding operation applied on the singular values of the evolving solution, as explained next (see details in Appendix \ref{appsec:Proofs and Additional Details for Section 4}). Consider a matrix $\mtx{W}^{(\text{\rm in} )}\in\mathbb{R}^{p\times m}$ (where $p\ge m$), with the SVD $\mtx{W}^{(\text{\rm in} )}=\mtx{\Omega} \mtx{\Sigma}^{({\rm in})} \mtx{\Theta}^{T}$, where $\mtx{\Omega}$ and $\mtx{\Theta}$ are $p\times p$ and $m\times m$ real orthonormal matrices, respectively, and $\mtx{\Sigma}^{({\rm in})}$ is a $p\times m$ real diagonal matrix with $m$ singular values $\lbrace{ \sigma_{i}\left( \mtx{W}^{(\text{\rm in} )} \right) }\rbrace_{i=1}^{m}$ on its main diagonal (recall that, by definition, singular values are non-negative). Projecting $\mtx{W}^{(\text{\rm in} )}$ on the soft-orthonormality constraints via 
\begin{align} 
\label{eq:soft ortho constraint projection - optimization problem form}
    &{\mtx{W}^{(\text{\rm out} )}} = \argmin_{\mtx{W}\in\mathbb{R}^{p\times m}} \left \Vert  \mtx{W} - {\mtx{W}^{(\text{\rm in} )}} \right \Vert _F^2
    \\ \nonumber
    &\text{subject to}~~\lvert{  \sigma_{i}^{2}\left( \mtx{W} \right) - 1 }\rvert \le \alpha ~~\text{for }i=1,...,m
\end{align}
is equivalent to the thresholding mapping  ${{\mtx{W}^{(\text{\rm out} )}}\triangleq T_{\alpha}\left({\mtx{W}^{(\text{\rm in} )}}\right)}$ where ${\mtx{W}^{(\text{\rm out} )}}=\mtx{\Omega} \mtx{\Sigma}^{({\rm out})} \mtx{\Theta}^{T}$ and the singular values along the main diagonal of $\mtx{\Sigma}^{({\rm out})}$ are 
\begin{align}
\label{eq:linear subspace fitting optimization - supervised with soft constraints - singular values thresholding}
    &\sigma_{i}\left( \mtx{W}^{(\text{\rm out} )} \right)=
    \\
    &\begin{cases}
      \sigma_{i}\left( \mtx{W}^{(\text{\rm in} )} \right), \qquad \text{if}\ \sigma_{i}\left( \mtx{W}^{(\text{\rm in} )} \right) \in \left[\tau_{\alpha}^{\rm low},\tau_{\alpha}^{\rm high}\right] 
      \\
       {\tau_{\alpha}^{\rm low}} , ~~~\qquad\qquad \text{if}\ {\sigma_{i}\left( \mtx{W}^{(\text{\rm in} )} \right)} <  {\tau_{\alpha}^{\rm low}} 
      \\
      {\tau_{\alpha}^{\rm high}} , ~~\qquad\qquad \text{if}\ {\sigma_{i}\left( \mtx{W}^{(\text{\rm in} )} \right)} >  {\tau_{\alpha}^{\rm high}}
    \end{cases} \nonumber
\end{align}
for $i=1,...,m$, where the threshold levels are defined by $\tau_{\alpha}^{\rm low} \triangleq {\sqrt{\max{\lbrace{0, 1-\alpha} \rbrace }}} $ and $\tau_{\alpha}^{\rm high} \triangleq {\sqrt{1+\alpha}} $. 
The entire optimization process is like in Algorithm \ref{alg:Supervised Subspace Fitting via Projected Gradient Descent}, except that the projections onto the constraint are done using the soft thresholding $T_{\alpha}$ defined using (\ref{eq:linear subspace fitting optimization - supervised with soft constraints - singular values thresholding}) (instead of the hard thresholding $T_{\rm hard})$. See Appendix \ref{appsec:Proofs and Additional Details for Section 4} for details.

The empirical demonstration in Fig.~\ref{fig:fully supervised with soft ortho constraints - out-of-sample errors - AVERAGE} shows the generalization errors (as function of $p$) corresponding to a range of problem settings where $\alpha$ gradually increases from 0 (i.e., strictly  constrained setting) to $\infty$ (i.e., practically unconstrained, standard regression problem). This demonstrates that the double-descent trend emerges in the fully supervised setting as the orthonormality constraints are relaxed (and eventually removed). The evolution of the corresponding in-sample error curves in Fig.~\ref{fig:fully supervised with soft ortho constraints - in-sample errors - AVERAGE} shows that the range of interpolating solutions gradually increases as the orthonormality constraints are relaxed. Specifically, for a given $\alpha$, the interpolation occurs for $p\ge p_{\alpha}$ where ${p_{\alpha} \ge n-1}$ is a threshold that monotonically decreases together with the increase in the constraint level $\alpha$. Eventually, when the orthonormality constraint is completely removed (i.e., $\alpha\rightarrow\infty$), the range of interpolating solutions becomes the full range of overparameterized solutions (i.e., $p\ge n-1$). Interestingly, the peaks of the double descent trends of the out-of-sample error curves are still obtained at ${p=n-1}$ even if ${p_{\alpha} > n-1}$. In the few supervised settings where the orthonormality is nearly or exactly strictly constrained, the curves do not arrive to accurate interpolation ability and accordingly the double descent shape is not apparent (or apparent in very weak forms) in the matching out-of-sample error curves.

Our findings for fully supervised settings with varying orthonormality constraints can be also examined in the future for other formulations of the optimization cost and constraints, and different optimization techniques.

\section{Semi-Supervised Subspace Fitting}
\label{sec:Semi-Supervised Subspace Fitting}

The fully supervised problem (\ref{eq:linear subspace fitting optimization - supervised with soft constraints}), enabling flexible orthonormality constraint levels, demonstrated the important dependency of the double-descent behavior on the constraints. Now we turn to explore the supervision level as the additional crucial factor for the existence of double descent in subspace estimation tasks. Here, we essentially establish the ability to explore estimation problems induced \textit{anywhere} on the supervision-orthonormality plane (Fig.~\ref{fig:The supervision-orthonormality plane}).

We define a learning problem with an arbitrary level of supervision, implemented as described next. 
The data model is again as specified in Section \ref{subsec:data model}. However, now, the provided dataset of $n$ examples is $\mathcal{D}^{\rm semisup} \triangleq {\widetilde{\mathcal{D}}^{\rm sup} } \cup {\widetilde{\mathcal{D}}^{\rm unsup} }$, where ${\widetilde{\mathcal{D}}^{\rm sup} } \triangleq \left \{ \left( \vec{x}^{(\ell)},\vec{z}^{(\ell)} \right) \right\}_{\ell=1}^{n^{\rm sup} } \in \mathbb{R}^d \times \mathbb{R}^m$ is a set of $n^{\rm sup} \in\{0,\dots,n\}$ i.i.d.\ samples of $(\vec{x},\vec{z})$ pairs satisfying (\ref{eq:data model - x}), and ${\widetilde{\mathcal{D}}^{\rm unsup} }\triangleq \left \{  \vec{x}^{(\ell)} \right\}_{\ell=n^{\rm sup}  + 1}^{n} \in \mathbb{R}^d$ contains additional ${n^{\rm unsup}  \triangleq n-n^{\rm sup} }$ i.i.d.~samples of $\vec{x}$. Again, the learning goal is to estimate a linear operator $\widehat{\mtx{U}}_{m}$, where only the $p$ features (specified in $\mathcal{S}$) of $\vec{x}$ are used in the actual learning. Note the extreme cases of $n^{\rm sup} =0$ and $n^{\rm sup} =n$ where the setting reduces to unsupervised and fully-supervised forms, respectively. For any $n^{\rm sup} \in\{1,...,n-1\}$, the problem is semi-supervised at a level that grows with $n^{\rm sup} $. 

\begin{figure*}[t]
\centering
\subfloat[]{\includegraphics[width=0.8\columnwidth]{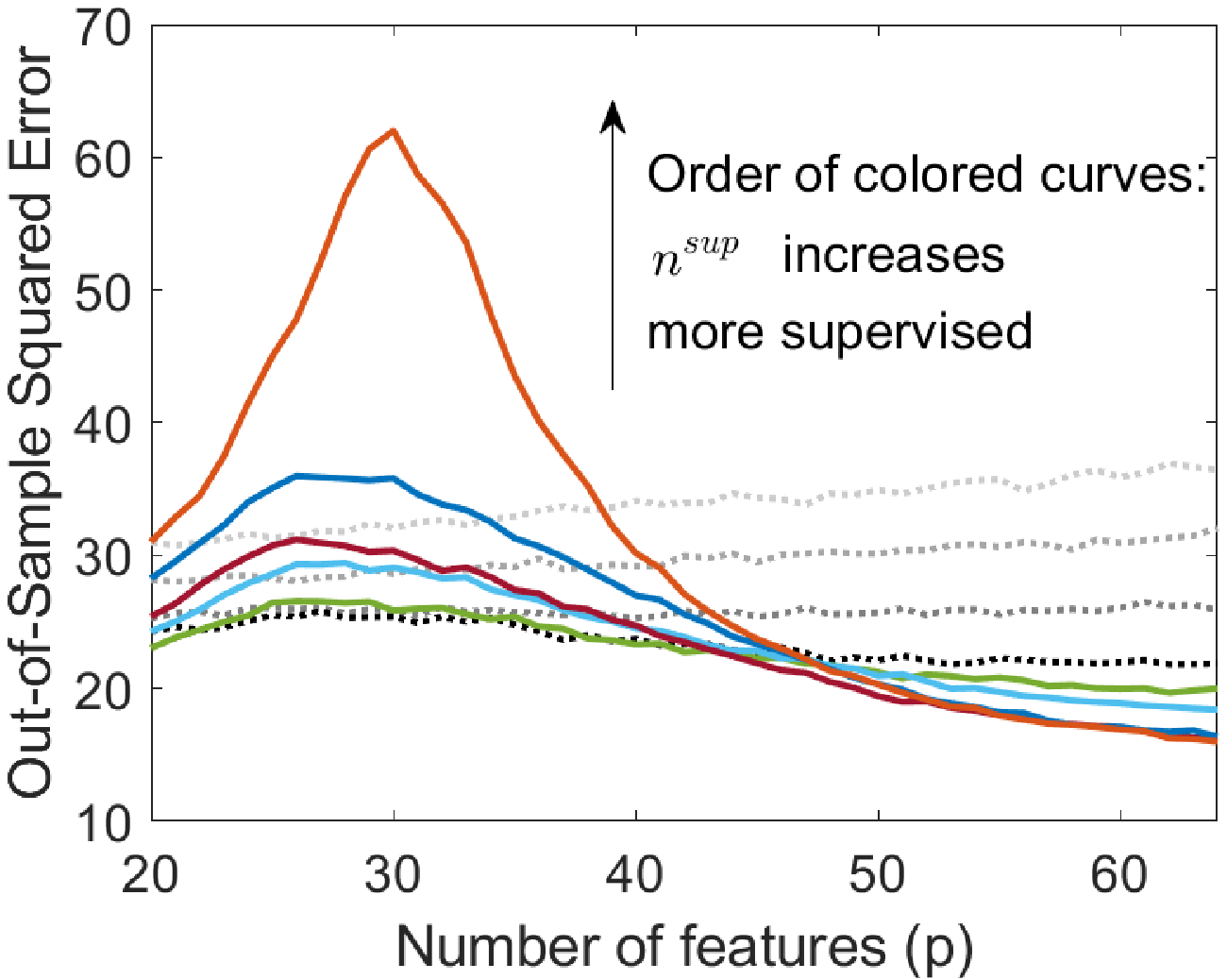}\label{fig:semi supervised with soft ortho constraints - out-of-sample errors - AVERAGE}}
\subfloat[]{\includegraphics[width=0.8\columnwidth]{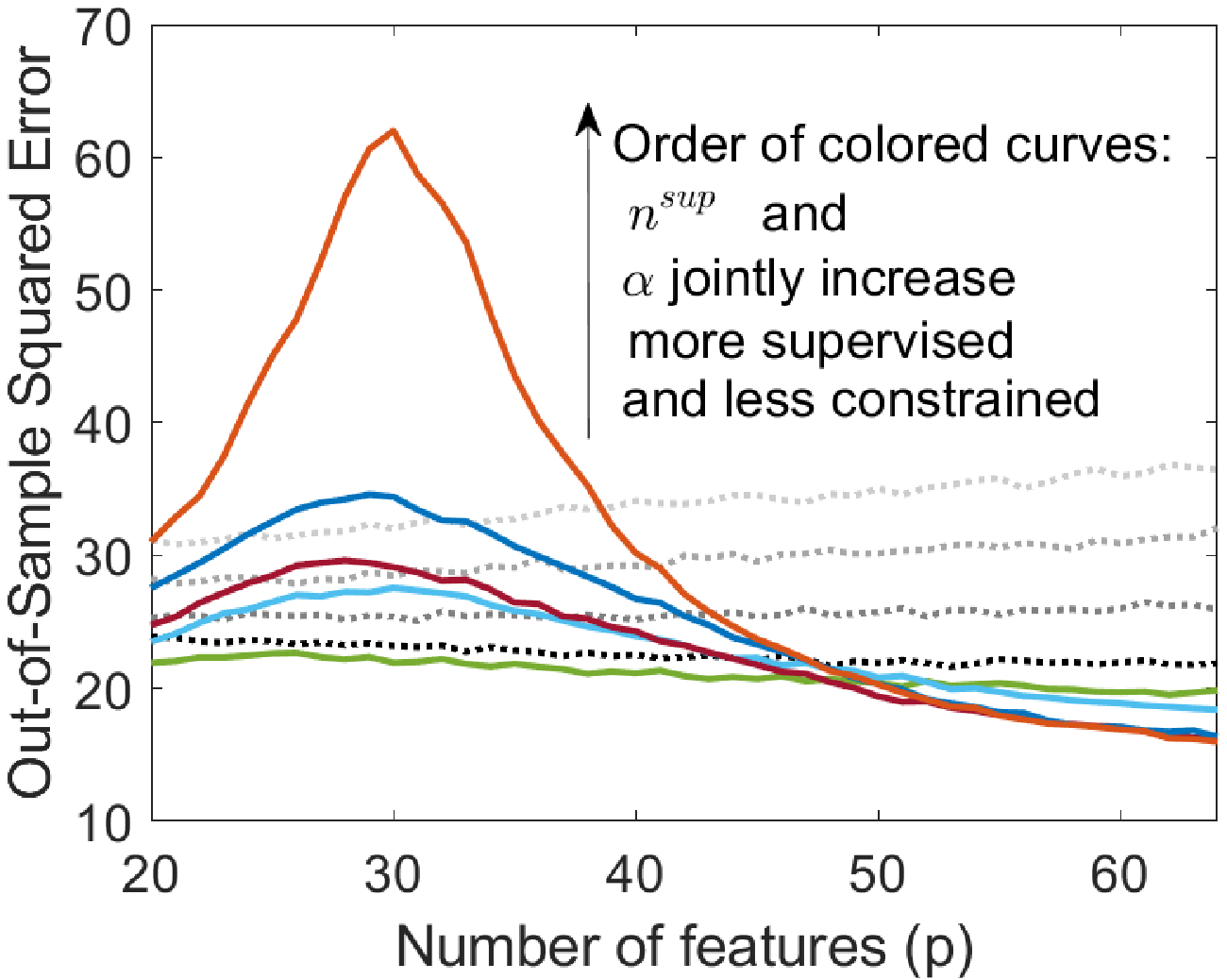}\label{fig:diagonal direction - out-of-sample errors - AVERAGE}}
\caption{The out-of-sample errors, $\mathcal{E}_{\rm out} ^{\rm sup}  \left( \widehat{\mtx{U}}_{m} \right)$ versus the number of parameters $p$. The errors are averaged over 25 experiments with different sequential orders of adding coordinates to $\mathcal{S}$. Here $d=64$, $m=20$ and $n=32$. 
\textbf{(a) Unconstrained} settings ($\alpha\rightarrow \infty$):  Each curve presents the results for a different supervision level, $n^{\rm sup} \in \left\{ {0,4,8,12, 16,20,24,28,n=32} \right\}$.
\textbf{(b)} Problems residing at the supervision-orthonormality plane along the \textbf{diagonal trajectory} connecting the standard subspace fitting and the pure regression. Each curve presents the results for a different pair of supervision and orthonormality constraint levels that jointly increase.
In both subfigures, the gray dotted curves correspond to $n^{\rm sup} \in \left\{ {0,4,8,12} \right\}$.}
\end{figure*}

We define the learning task by extending (\ref{eq:linear subspace fitting optimization - supervised with soft constraints}) into 
\begin{align} 
\label{eq:linear subspace fitting optimization - semi-supervised with soft constraints}
    &\widehat{\mtx{U}}_{m,\mathcal{S}} =  \argmin_{\mtx{W}\in\mathbb{R}^{p\times m}} \left\{ \left \Vert  \mtx{Z}^{\rm sup}  - \mtx{W}^T \mtx{X}_{\mathcal{S}}^{\rm sup}   \right \Vert _F^2  \right.
    \nonumber \\ \nonumber 
    &\quad\quad\quad\quad\quad\quad\quad \left. + \left \Vert  \left( {\mtx{I}_{p} - \mtx{W} \mtx{W}^T } \right) \mtx{X}_{\mathcal{S}}^{\rm unsup}  \right \Vert _F^2 \right\}
    \\ 
    &\text{subject to} ~\lvert{  \sigma_{i}^{2}\left( \mtx{W} \right) - 1 }\rvert \le \alpha ~~\text{for }i=1,...,m
\end{align}
where $ \mtx{X}_{\mathcal{S}}^{\rm sup}  \triangleq \left[ \vec{x}^{(1)}_{\mathcal{S}} , \dots, \vec{x}^{(n^{\rm sup} )}_{\mathcal{S}} \right]$, $ \mtx{Z}^{\rm sup}  \triangleq \left[ \vec{z}^{(1)} , \dots, \vec{z}^{(n^{\rm sup} )} \right]$, $ \mtx{X}_{\mathcal{S}}^{\rm unsup}  \triangleq \left[ \vec{x}^{(n^{\rm sup}  + 1)}_{\mathcal{S}} , \dots, \vec{x}^{n}_{\mathcal{S}} \right]$, and $\alpha$ determines the orthonormality constraint level. The optimization cost in (\ref{eq:linear subspace fitting optimization - semi-supervised with soft constraints}) naturally blends the supervised and unsupervised metrics in proportions induced by the ${n^{\rm sup} }$ to ${n^{\rm unsup} }$ ratio. 
We address (\ref{eq:linear subspace fitting optimization - semi-supervised with soft constraints}) using a projected gradient descent approach. Since (\ref{eq:linear subspace fitting optimization - semi-supervised with soft constraints}) extends (\ref{eq:linear subspace fitting optimization - supervised with soft constraints}) only with respect to the optimization cost, the current optimization procedure extends Algorithm \ref{alg:Supervised Subspace Fitting via Projected Gradient Descent} by using the soft-threshold projection $T_{\alpha}$ from (\ref{eq:linear subspace fitting optimization - supervised with soft constraints - singular values thresholding}), replacing the gradient descent stage with the one suitable to the new cost function in (\ref{eq:linear subspace fitting optimization - semi-supervised with soft constraints}), and 
initializing the optimization process with a random matrix (of i.i.d.~Gaussian components with zero mean and variance $1/p$) that is projected onto the relevant orthonormality constraint.
See Appendix \ref{appsec:Additional Details for Section 5} for the detailed development of the algorithm.

At this stage, equipped with the problem defined in (\ref{eq:linear subspace fitting optimization - semi-supervised with soft constraints}), we are able to generate a subspace estimation problem at any point of the supervision-orthonormality plane (recall Fig.~\ref{fig:The supervision-orthonormality plane}) and empirically evaluate the corresponding generalization errors as function of the number of features $p$ used in the actual learning. 
We start by evaluating the range of problems that are unconstrained (i.e., $\alpha\rightarrow \infty$) and their supervision level gradually varies from unsupervised ($n^{\rm sup} =0$) to fully supervised ($n^{\rm sup} =n$). This set of problems is located along the right, blue-colored  border line of the supervision-orthonormality plane in Fig.~\ref{fig:The supervision-orthonormality plane}.
Figure~\ref{fig:semi supervised with soft ortho constraints - out-of-sample errors - AVERAGE} clearly demonstrates the emergence of the double descent trend together with the increase in supervision level. This shows that double descent can occur in problems that are semi-supervised and deviate from the ordinary regression form. 
Our concluding demonstration evaluates the range of problems on the diagonal trajectory (on the supervision-orthonormality plane) connecting the standard subspace fitting and the pure regression settings (see the purple-colored trajectory in Fig.~\ref{fig:The supervision-orthonormality plane}). Here we simultaneously increase $\alpha$ (from $0$ to $\infty$) and $n^{\rm sup}$ (from $0$ to $n$). The observed generalization errors (Fig. \ref{fig:diagonal direction - out-of-sample errors - AVERAGE}) clearly exhibit the rise of the double descent phenomena together with the joint increase in supervision level and decrease in orthonormality level.

\section{Conclusions}
\label{sec:conclusion}

In this work we have opened up a new avenue of research on linear subspace estimation problems. 
We defined a family of linear subspace estimation problems that reside over a supervision-orthonormality plane (where each coordinate induces a unique problem setting). This class of problems connects the standard subspace fitting and the pure regression problems. We proposed an optimization procedure, based on the projected gradient descent technique, to evaluate any problem instance on the supervision-orthonormality plane. Then, we explored problems defined along various trajectories of the supervision-orthonormality plane, and showed that the double-descent phenomena is more evident as the problems are more supervised and less orthonormally constrained. 
We believe that our findings open a new direction of theoretical and practical research of the generalization ability of overparameterized models learned in diverse supervision levels (i.e., including semi-supervised settings) and various optimization constraints.

\section*{Acknowledgments}
This work was supported by 
NSF grants CCF-1911094, IIS-1838177, and IIS-1730574; 
ONR grants N00014-18-12571 and N00014-17-1-2551;
AFOSR grant FA9550-18-1-0478; 
DARPA grant G001534-7500; and a 
Vannevar Bush Faculty Fellowship, ONR grant N00014-18-1-2047.

\renewcommand{\theequation}{\thesection.\arabic{equation}}

\renewcommand\thefigure{\thesection.\arabic{figure}}    
\setcounter{figure}{0}  

\renewcommand{\thealgorithm}{\thesection.\arabic{algorithm}} 
\setcounter{algorithm}{1}

\counterwithin*{equation}{section}
\counterwithin*{figure}{section}
\counterwithin*{algorithm}{section}

\appendix 

\section*{Appendices}
\label{sec:Appendix Outline}
These appendices support the main paper in the following ways. Appendix \ref{appsec:Proofs and Explanations for Section 3} provides proofs and various explanations to the statements provided in Section 3 of the main text. In particular, in Appendix Section \ref{appendixsubsec:monotonic decrease justification}, we provide mathematical analysis and experimental justification for the claim regarding the \textit{on average} decrease of the out-of-sample error $\mathcal{E}_{\rm out} ^{\rm unsup}  \left( \widehat{\mtx{U}}_{k} \right)$ with the number of features $p$. 
In Appendix \ref{appsec:Proofs and Additional Details for Section 4} we refer to Section 4 of the paper, prove the specific projection operators used in our projected gradient descent algorithms, and provide additional details on the experiments for the supervised settings. 
In Appendix \ref{appsec:Additional Details for Section 5} we elaborate on the semi-supervised subspace fitting method presented in Section 5 of the main text. 
Appendix \ref{appsec:Unsupervised Subspace Fitting with Soft Orthonormality Constraints} provides the details on the range of unsupervised problems with soft orthonormality constraints. 

Note that the indexing of equations and figures in the Appendices below is prefixed with the letter of the corresponding Appendix. Other references correspond to the main paper.

\section{Proofs and Explanations for Section 3}
\label{appsec:Proofs and Explanations for Section 3}

\subsection{Explanation for Corollaries 3.1 and 3.2}
One should note that any overparameterized subspace estimate $\widehat{\mathcal{U}}_{k}$ is induced by a rank-deficient sample covariance matrix $\widehat{\mtx{C}}_\vec{x,\mathcal{S}}^{(n)}$ of rank $\rho \triangleq {\rm rank}\left\lbrace \widehat{\mtx{C}}_\vec{x,\mathcal{S}}^{(n)} \right\rbrace \leq n-1$. This is simply because $\widehat{\mtx{C}}_\vec{x,\mathcal{S}}^{(n)}$ is formed based on $n$ centered samples of $p$-dimensional feature vectors where, as implied from the definition of overparameterization, $p>n$. This is also the case for rank-overparameterized subspace estimates (which are a particular type of overparameterized subspace estimates). However, the point that Corollary 3.1 emphasizes is that rank-overparameterized subspace estimates are guaranteed to be affected by the rank-deficiency of $\widehat{\mtx{C}}_\vec{x,\mathcal{S}}^{(n)}$. Accordingly, the construction provided in Corollary 3.2 shows that, due to the insufficient number of nonzero eigenvalues of $\widehat{\mtx{C}}_\vec{x,\mathcal{S}}^{(n)}$, a rank-overparameterized estimate has freedom in setting $k-\rho$ out of its $k$ spanning orthonormal vectors.

\subsection{Proof of Proposition 3.1}
Since $p>n$ (due to overparameterization), the sample covariance matrix $\widehat{\mtx{C}}_{\vec{x},\mathcal{S}}^{(n)}$ has size $p\times p$ and rank ${\rho \triangleq {\rm rank}\left\{{ \widehat{\mtx{C}}_{\vec{x},\mathcal{S}}^{(n)} }\right\} \le n-1}$. Hence, the eigendecomposition $\widehat{\mtx{C}}_{\vec{x},\mathcal{S}}^{(n)} = \widehat{\mtx{\Psi}}_{\mathcal{S}} \widehat{\mtx{\Lambda}}\widehat{\mtx{\Psi}}_{\mathcal{S}}^{*}$ corresponds to a $p\times p$ unitary matrix $\widehat{\mtx{\Psi}}_{\mathcal{S}} \triangleq\left[ \widehat{\vecgreek{\psi}}_{\mathcal{S}}^{(1)},\dots,\widehat{\vecgreek{\psi}}_{\mathcal{S}}^{(p)} \right]$ and a diagonal matrix $\widehat{\mtx{\Lambda}}\triangleq\text{diag}\left\{{ \widehat{\lambda}^{(1)}, \dots, \widehat{\lambda}^{(p)} }\right\}$ with only $\rho$ nonzero eigenvalues $\widehat{\lambda}^{(h_1)}, \dots, \widehat{\lambda}^{(h_{\rho})}$, where ${1\le h_1 < h_2 < \dots <h_{\rho} \le p }$. Therefore, the eigenvectors $\widehat{\vecgreek{\psi}}_{\mathcal{S}}^{(h_1)},\dots,\widehat{\vecgreek{\psi}}_{\mathcal{S}}^{(h_{\rho})} $ are those associated with the nonzero eigenvalues. Here $\mtx{\Psi}^{*}$ denotes the conjugate transpose of the matrix $\mtx{\Psi}$.

The subspace estimate is rank-overparameterized (recall Definition 3.2), thus, ${p>n}$ and $k\in\lbrace n,\dots,p \rbrace$. Then, $\widehat{\mtx{U}}_{k,\mathcal{S}}$ is a $p\times k$ matrix with $k$ orthonormal columns, where the first $\rho$ of them satisfy $\widehat{\vec{u}}_{\mathcal{S}}^{(i)}=\widehat{\vecgreek{\psi}}^{(h_i)}$ for $i=1,...,\rho$. The additional $k-\rho$ columns $\widehat{\vec{u}}_{\mathcal{S}}^{(\rho + 1)},\dots,\widehat{\vec{u}}_{\mathcal{S}}^{(k)}$  are chosen arbitrarily from the $p-\rho$ columns of $\widehat{\mtx{\Psi}}_{\mathcal{S}}$ corresponding to zero eigenvalues. 
Namely, $\widehat{\vec{u}}_{\mathcal{S}}^{(\rho + i)}=\widehat{\vecgreek{\psi}}^{(r_i)}$ for ${i=1,...,k-\rho}$ and $\lbrace r_1, \dots, r_{k-\rho} \rbrace$ is an arbitrary subset of ${\lbrace 1, \dots, p \rbrace \setminus \lbrace h_1, \dots, h_{\rho} \rbrace}$.
This construction satisfies the orthonormality demand for the $k$ columns of $\widehat{\mtx{U}}_{k,\mathcal{S}}$.

Here, the in-sample error of interest is (7), namely, \begin{equation}
\label{eq:linear subspace fitting - in sample error - actual learning - proposition 3.1 proof}
\begin{split}
 &\mathcal{E}_{{\rm in},\mathcal{S}}^{\rm unsup}  \left( \widehat{\mtx{U}}_{k,\mathcal{S}} \right) =
 \\
 &\mtxtrace{ \left ( \mtx{I}_{p} - \widehat{\mtx{U}}_{k,\mathcal{S}} \widehat{\mtx{U}}_{k,\mathcal{S}}^* \right ) \widehat{\mtx{\Psi}}_{\mathcal{S}} \widehat{\mtx{\Lambda}}\widehat{\mtx{\Psi}}_{\mathcal{S}}^{*} \left ( \mtx{I}_p - \widehat{\mtx{U}}_{k,\mathcal{S}} \widehat{\mtx{U}}_{k,\mathcal{S}}^* \right )^* }
\end{split}
\end{equation}
Note that, by the construction of $\widehat{\mtx{U}}_{k,\mathcal{S}}$, the eigendecomposition of the $p\times p$ projection operator $\widehat{\mtx{U}}_{k,\mathcal{S}}\widehat{\mtx{U}}_{k,\mathcal{S}}^*$ satisfies 
\begin{equation}
\label{eq:proposition 3.1 proof - eigendecomposition of projection operator}
\widehat{\mtx{U}}_{k,\mathcal{S}}\widehat{\mtx{U}}_{k,\mathcal{S}}^* = \widehat{\mtx{\Psi}}_{\mathcal{S}} {\widehat{\mtx{\Lambda}}_{\mathcal{S},\text{{\rm in}d}[k]}} \widehat{\mtx{\Psi}}_{\mathcal{S}}^{*}
\end{equation}
where $\widehat{\mtx{\Psi}}_{\mathcal{S}}$ is the $p\times p$ unitary matrix that diagonalizes $\widehat{\mtx{C}}_{\vec{x},\mathcal{S}}^{(n)}$, and ${\widehat{\mtx{\Lambda}}_{\mathcal{S},\text{{\rm in}d}[k]}}$ is a $p\times p$ diagonal matrix with ones at the coordinates ${\lbrace (h_1,h_1), \dots, (h_{\rho},h_{\rho}) \rbrace \cup \lbrace (r_1,r_1), \dots, (r_{k-\rho},r_{k-\rho}) \rbrace}$ and zeros elsewhere. 
Therefore, 
\begin{align}
\label{eq:linear subspace fitting - in sample error - actual learning - proposition 3.1 proof - cont.}
 &\mathcal{E}_{{\rm in},\mathcal{S}}^{\rm unsup}  \left( \widehat{\mtx{U}}_{k,\mathcal{S}} \right) = \text{Tr} \left\{ \widehat{\mtx{\Psi}}_{\mathcal{S}} \left ( \mtx{I}_{p} - {\widehat{\mtx{\Lambda}}_{\mathcal{S},\text{{\rm in}d}[k]}} \right )  \times \right.
 \nonumber \\ & \qquad\qquad \left. \widehat{\mtx{\Psi}}_{\mathcal{S}}^{*}\widehat{\mtx{\Psi}}_{\mathcal{S}} \widehat{\mtx{\Lambda}}\widehat{\mtx{\Psi}}_{\mathcal{S}}^{*} \widehat{\mtx{\Psi}}_{\mathcal{S}} \left( \mtx{I}_p - {\widehat{\mtx{\Lambda}}_{\mathcal{S},\text{{\rm in}d}[k]}} \right ) \widehat{\mtx{\Psi}}_{\mathcal{S}}^{*} \right\}
 \nonumber  \\
 &=\mtxtrace{\left( \mtx{I}_{p} - {\widehat{\mtx{\Lambda}}_{\mathcal{S},\text{{\rm in}d}[k]}} \right)  \widehat{\mtx{\Lambda}}\left ( \mtx{I}_p - {\widehat{\mtx{\Lambda}}_{\mathcal{S},\text{{\rm in}d}[k]}} \right ) }
 \nonumber \\
 & = \mtx{0}
\end{align}
This proves that a rank-overparameterized subspace estimate formed by the construction in Corollary 3.2 is $\mathcal{S}$-interpolating. 

One can extend the last proof to the general form of rank-overparameterized subspace estimates, where the additional arbitrary $k-\rho$ orthonormal vectors can be any $(k-\rho)$-dimensional subspace of the $(p-\rho)$-dimensional null space of $\widehat{\mtx{C}}_{\vec{x},\mathcal{S}}^{(n)}$.

~
\subsection{Proof of Proposition 3.2} 

Let us denote the eigenvalues of the true covariance matrix, ${\mtx{C}}_{\vec{x}}$, as $\lambda^{(1)},\dots,\lambda^{(d)}$. The covariance matrix of the $p$-dimensional feature vectors is denoted as ${\mtx{C}}_{\vec{x},\mathcal{S}}$, and its eigendecomposition satisfies ${{\mtx{C}}_{\vec{x},\mathcal{S}} = {\mtx{\Psi}_{\mathcal{S}}} {\mtx{\Lambda}_{\mathcal{S}}}{\mtx{\Psi}_{\mathcal{S}}^{*}}}$ where ${\mtx{\Psi}_{\mathcal{S}}}$ is a $p\times p$ unitary matrix, and ${{{\mtx{\Lambda}_{\mathcal{S}}}=\text{diag}\left\{{ \lambda_{\mathcal{S}}^{(1)}, \dots , \lambda_{\mathcal{S}}^{(p)}}\right\}}}$ is a diagonal matrix containing the eigenvalues of ${\mtx{C}}_{\vec{x},\mathcal{S}}$. Similar to the construction in (\ref{eq:proposition 3.1 proof - eigendecomposition of projection operator}) we have here ${\widehat{\mtx{U}}_{k,\mathcal{S}}\widehat{\mtx{U}}_{k,\mathcal{S}}^{*} = \widehat{\mtx{\Psi}}_{\mathcal{S}} {\widehat{\mtx{\Lambda}}_{\mathcal{S},\text{{\rm in}d}[k]}} \widehat{\mtx{\Psi}}_{\mathcal{S}}^{*}}$, where ${\widehat{\mtx{\Lambda}}_{\mathcal{S},\text{{\rm in}d}[k]}}$ is a diagonal matrix with ones at the main-diagonal coordinates corresponding to columns of $\widehat{\mtx{\Psi}}_{\mathcal{S}}$ chosen to define $\widehat{\mtx{U}}_{k,\mathcal{S}}$ and zeros elsewhere. 
Then, the expression for the unsupervised out-of-sample error is developed as follows.  
\begin{align} 
\label{eq:linear subspace fitting - out of sample error - proof of proposition 3.2}
    & \mathcal{E}_{\rm out} ^{\rm unsup}  \left( \widehat{\mtx{U}}_{k} \right) = \mathbb{E} \left \Vert { \left ( \mtx{I}_d - \widehat{\mtx{U}}_{k} \widehat{\mtx{U}}_{k}^* \right ) \vec{x}_{\rm test}} \right \Vert _2^2 
    \nonumber \\ 
    &  = \mathbb{E} \left \Vert \vec{x}_{\rm test} \right \Vert _2^2 - \mathbb{E} \left \Vert {  \widehat{\mtx{U}}_{k} \widehat{\mtx{U}}_{k}^* \vec{x}_{\rm test}} \right \Vert _2^2 
    \nonumber \\ 
    & = \mtxtrace{{\mtx{C}}_{\vec{x}}} - \mathbb{E} \left \Vert {  \widehat{\mtx{U}}_{k,\mathcal{S}} \widehat{\mtx{U}}_{k,\mathcal{S}}^* \vec{x}_{{\rm test},\mathcal{S}}} \right \Vert _2^2
    \nonumber \\ 
    & = \mtxtrace{{\mtx{C}}_{\vec{x}}} - \mtxtrace{ \widehat{\mtx{U}}_{k,\mathcal{S}} \widehat{\mtx{U}}_{k,\mathcal{S}}^* {\mtx{C}}_{\vec{x},\mathcal{S}}    \widehat{\mtx{U}}_{k,\mathcal{S}} \widehat{\mtx{U}}_{k,\mathcal{S}}^*  }
    \nonumber \\ 
    & = \mtxtrace{{\mtx{C}}_{\vec{x}}} 
    \nonumber \\
    &~~~ - \mtxtrace{  \widehat{\mtx{\Psi}}_{\mathcal{S}} {\widehat{\mtx{\Lambda}}_{\mathcal{S},\text{{\rm in}d}[k]}} \widehat{\mtx{\Psi}}_{\mathcal{S}}^{*}
    {\mtx{\Psi}_{\mathcal{S}}} {\mtx{\Lambda}_{\mathcal{S}}}{\mtx{\Psi}_{\mathcal{S}}^{*}}
    \widehat{\mtx{\Psi}}_{\mathcal{S}} {\widehat{\mtx{\Lambda}}_{\mathcal{S},\text{{\rm in}d}[k]}} \widehat{\mtx{\Psi}}_{\mathcal{S}}^{*}}
    \nonumber \\ 
    & = \mtxtrace{{\mtx{C}}_{\vec{x}}} - \mtxtrace{   {\widehat{\mtx{\Lambda}}_{\mathcal{S},\text{{\rm in}d}[k]}} \widehat{\mtx{\Psi}}_{\mathcal{S}}^{*}
    {\mtx{\Psi}_{\mathcal{S}}} {\mtx{\Lambda}_{\mathcal{S}}}{\mtx{\Psi}_{\mathcal{S}}^{*}}
    \widehat{\mtx{\Psi}}_{\mathcal{S}} }
    \nonumber \\ 
    & = \sum_{i=1}^{d} {\lambda^{(i)}} - \sum_{i\in\mathcal{S}} {{\lambda_{\mathcal{S},\text{{\rm in}d}[k]}^{(i)}} \sum_{j=1}^{p} { {\lambda_{\mathcal{S}}^{(j)}} {\left| \left\langle {\vecgreek{\psi}}_{\mathcal{S}}^{(j)}, \widehat{\vecgreek{\psi}}_{\mathcal{S}}^{(i)} \right\rangle \right|}^2 } }
    \nonumber \\
    & = \sum_{i=1}^{d} {\lambda^{(i)}} - \sum_{i\in\widehat{\mathcal{S}}_{\rm{max}}^{(k)}}{ \sum_{j=1}^{p} {\lambda_{\mathcal{S}}^{(j)}} {\left| \left\langle {\vecgreek{\psi}}_{\mathcal{S}}^{(j)}, \widehat{\vecgreek{\psi}}_{\mathcal{S}}^{(i)} \right\rangle \right| }^2 }
\end{align}
where $\widehat{\mathcal{S}}_{\rm{max}}^{(k)} \subset \lbrace{1,\dots,p}\rbrace$ is the set of $k$ indices corresponding to the columns of $\widehat{\mtx{\Psi}}_{\mathcal{S}}$ used for the construction of $\widehat{\mtx{U}}_{k,\mathcal{S}}$. 
This means that the indices in $\widehat{\mathcal{S}}_{\rm{max}}^{(k)}$ correspond to the $k$ maximal eigenvalues of $\widehat{\mtx{C}}_\vec{x,\mathcal{S}}^{(n)}$. If $k>\rho$,  then $k-\rho$ of the indices in $\widehat{\mathcal{S}}_{\rm{max}}^{(k)}$ correspond to zero eigenvalues.

~

~

~
\subsection{Proof of Proposition 3.3} 

The error expression provided in Proposition 3.2  has the property that 
\begin{equation}
\label{eq:proposition 3.3 proof}
\begin{split}
&\mathcal{E}_{\rm out} ^{\rm unsup}  \left( \widehat{\mtx{U}}_{k+1} \right) =
\\
& \mathcal{E}_{\rm out} ^{\rm unsup}  \left( \widehat{\mtx{U}}_{k} \right) -  \sum_{j=1}^{p} {\lambda_{\mathcal{S}}^{(j)}} {\left| \left\langle {\vecgreek{\psi}}_{\mathcal{S}}^{(j)}, \widehat{\vecgreek{\psi}}_{\mathcal{S}}^{(i_{\text{added}})} \right\rangle \right|}^2
\end{split}
\end{equation}
where ${i_{\text{added}}\in \lbrace{1,\dots,p}\rbrace \setminus \widehat{\mathcal{S}}_{\rm{max}}^{(k)}}$ is the index of the column of $\widehat{\mtx{\Psi}}_{\mathcal{S}}$ that is joined to $\widehat{\mtx{U}}_{k}$ as the $(k+1)$-th column that yields $\widehat{\mtx{U}}_{k+1}$.
Note that ${\lambda_{\mathcal{S}}^{(j)}}\ge 0$ for any $j$, as these are eigenvalues of a covariance matrix. Hence, 
\begin{equation}
\label{eq:proposition 3.3 proof - non-negativity of the subtracted sum}
 \sum_{j=1}^{p} {\lambda_{\mathcal{S}}^{(j)}} {\left| \left\langle {\vecgreek{\psi}}_{\mathcal{S}}^{(j)}, \widehat{\vecgreek{\psi}}_{\mathcal{S}}^{(i_{\text{added}})} \right\rangle \right|}^2 \ge 0 .
\end{equation}
This implies that $\mathcal{E}_{\rm out} ^{\rm unsup}  \left( \widehat{\mtx{U}}_{k+1} \right) \le \mathcal{E}_{\rm out} ^{\rm unsup}  \left( \widehat{\mtx{U}}_{k} \right)$, proving that the unsupervised out-of-sample error is monotonic decreasing in $k$ (for a subspace construction that is sequential in $k$ as described above).

\subsection{On the Monotonic Decrease of $\mathcal{E}_{\rm out} ^{\rm unsup}  \left( \widehat{\mtx{U}}_{k} \right)$ with $p$}
\label{appendixsubsec:monotonic decrease justification}

We now justify our statement regarding the monotonic decrease of $\mathcal{E}_{\rm out} ^{\rm unsup}  \left( \widehat{\mtx{U}}_{k} \right)$ as the number of features, $p$, increases (and $k$ is kept fixed).

The following definitions and notations will be useful in the current discussion.
Consider a set ${\mathcal{S}_{p}\triangleq \lbrace{s_1,...,s_p}\rbrace}$ of ${p<d}$ coordinates ${1 \le s_1 < s_2 < \dots < s_p \le d }$. In addition, ${\mathcal{S}_{p+1} \triangleq \mathcal{S}_{p} \cup \lbrace s_{p+1}\rbrace }$ is a set of $p+1$ coordinates that is formed by adding a new coordinate ${s_{p+1}\in {\lbrace{1,\dots,d}\rbrace \setminus \mathcal{S}_{p}}}$  to $\mathcal{S}_{p}$. 
We also denote here the out-of-sample errors of interest with explicit indications of the underlying sets of coordinates: $\mathcal{E}_{\rm out} ^{\rm unsup}  \left( \widehat{\mtx{U}}_{k} ; \mathcal{S}_{p} \right)$ and $\mathcal{E}_{\rm out} ^{\rm unsup}  \left( \widehat{\mtx{U}}_{k} ; \mathcal{S}_{p+1} \right)$ are the errors induced by forming subspace estimates based on $\mathcal{S}_{p}$ and $\mathcal{S}_{p+1}$, respectively. Now, our goal is to justify the claim that 
\begin{equation}
\label{eq:conjecture justification - error decrease inequality}
\mathcal{E}_{\rm out} ^{\rm unsup}  \left( \widehat{\mtx{U}}_{k} ; \mathcal{S}_{p} \right) \ge \mathcal{E}_{\rm out} ^{\rm unsup}  \left( \widehat{\mtx{U}}_{k} ; \mathcal{S}_{p+1} \right) . 
\end{equation}

Using the error expression in (\ref{eq:linear subspace fitting - out of sample error - proof of proposition 3.2}), we translate the inequality (\ref{eq:conjecture justification - error decrease inequality}) into 
\begin{equation}
\label{eq:conjecture justification - error decrease inequality - eigendecomposition form}
\begin{split}
& \sum_{i\in\widehat{\mathcal{S}}_{p, \rm{max}}^{(k)}}{ \sum_{j=1}^{p} {\lambda_{\mathcal{S}_{p}}^{(j)}} {\left| \left\langle {\vecgreek{\psi}}_{\mathcal{S}_{p}}^{(j)}, \widehat{\vecgreek{\psi}}_{\mathcal{S}_{p}}^{(i)} \right\rangle \right|}^2 } 
\\
& \le \sum_{i\in\widehat{\mathcal{S}}_{p+1, \rm{max}}^{(k)}}{ \sum_{j=1}^{p+1} {\lambda_{\mathcal{S}_{p+1}}^{(j)}} {\left| \left\langle {\vecgreek{\psi}}_{\mathcal{S}_{p+1}}^{(j)}, \widehat{\vecgreek{\psi}}_{\mathcal{S}_{p+1}}^{(i)}  \right\rangle \right| }^2 }.
\end{split}
\end{equation}
Here, the covariance matrix of the $p$-feature vector induced by $\mathcal{S}_{p}$ is $\mtx{C}_{\vec{x},\mathcal{S}_{p}} \triangleq \mathbb{E} \lbrace {   \vec{x}_{\mathcal{S}_{p}} \vec{x}_{\mathcal{S}_{p}}^{T} }  \rbrace $, and its eigenvalues and eigenvectors are $\left\{ { {\lambda_{\mathcal{S}_{p}}^{(j)}} }\right\}_{j=1}^{p}$ and $\left\{ { {\vecgreek{\psi}}_{\mathcal{S}_{p}}^{(j)} }\right\}_{j=1}^{p}$, respectively. 
Similarly, 
the covariance matrix of the $(p+1)$-feature vector stemming from $\mathcal{S}_{p+1}$ is $\mtx{C}_{\vec{x},\mathcal{S}_{p+1}} \triangleq \mathbb{E} \lbrace {   \vec{x}_{\mathcal{S}_{p+1}} \vec{x}_{\mathcal{S}_{p+1}}^{T} }  \rbrace $, and its eigenvalues and eigenvectors are $\left\{ {\lambda_{\mathcal{S}_{p+1}}^{(j)}} \right\}_{j=1}^{p+1}$ and $\left\{ { {\vecgreek{\psi}}_{\mathcal{S}_{p+1}}^{(j)} }\right\}_{j=1}^{p+1}$, respectively. 
To distinguish between the various origins of $\widehat{\mathcal{S}}_{\rm{max}}^{(k)}$, we define here the notation of $\widehat{\mathcal{S}}_{p, \rm{max}}^{(k)}$ as the set of $k$ coordinates utilized based on the $p$-dimensional sample covariance matrix. Correspondingly, the set $\widehat{\mathcal{S}}_{p+1, \rm{max}}^{(k)}$ includes $k$ coordinates selected based on the $(p+1)$-dimensional sample covariance matrix.

For a start, note that the sums in (\ref{eq:conjecture justification - error decrease inequality - eigendecomposition form}) are over non-negative elements. Moreover, the inner summation on the right-hand side of (\ref{eq:conjecture justification - error decrease inequality - eigendecomposition form}) is over $p+1$ terms, whereas its counterpart sum on the left-hand side is over $p$ terms. However, the eigenvalues and eigenvectors in the two sides of (\ref{eq:conjecture justification - error decrease inequality - eigendecomposition form}) are different, as will be explained next.

The $p\times p$ covariance matrix $\mtx{C}_{\vec{x},\mathcal{S}_{p}}$ is a principal submatrix of ${\mtx{C}_{\vec{x},\mathcal{S}_{p+1}}}$, which is the covariance matrix of the $(p+1)$-feature vector induced by $\mathcal{S}_{p+1}$. This can be easily observed by defining the $p\times (p+1)$ matrix $\mtx{Q}$ such that ${\vec{x}_{\mathcal{S}_{p}} = \mtx{Q} \vec{x}_{\mathcal{S}_{p+1}}}$; namely, $\mtx{Q}$ deletes the single feature added to create $\vec{x}_{\mathcal{S}_{p+1}}$ from $\vec{x}_{\mathcal{S}_{p}}$. Then, 
\begin{align}
\label{eq:covariance submatrix relation}
\mtx{C}_{\vec{x},\mathcal{S}_{p}} & =  \mathbb{E} \lbrace {  \left( \mtx{Q}\vec{x}_{\mathcal{S}_{p+1}} \right) \left( \mtx{Q} \vec{x}_{\mathcal{S}_{p+1}}\right)^{T}  }\rbrace
\nonumber \\
& = \mtx{Q} \mathbb{E} \lbrace {   \vec{x}_{\mathcal{S}_{p+1}} \vec{x}_{\mathcal{S}_{p+1}}^{T} }\rbrace \mtx{Q}^{T}
\nonumber \\
& = \mtx{Q} \mtx{C}_{\vec{x},\mathcal{S}_{p+1}} \mtx{Q}^{T}.
\end{align}
This shows that the matrix $\mtx{C}_{\vec{x},\mathcal{S}_{p}}$ can be obtained from $\mtx{C}_{\vec{x},\mathcal{S}_{p+1}}$ by deletion of the row and column (having the same index) corresponding to the added feature. Thus, $\mtx{C}_{\vec{x},\mathcal{S}_{p}}$ is a principal submatrix of $\mtx{C}_{\vec{x},\mathcal{S}_{p+1}}$. 
This relation between the symmetric matrices  $\mtx{C}_{\vec{x},\mathcal{S}_{p}}$ and  $\mtx{C}_{\vec{x},\mathcal{S}_{p+1}}$, lets us apply Cauchy's interlacing theorem for eigenvalues of Hermitian matrices \cite{hwang2004cauchy} to obtain  
\begin{equation}
\label{eq:conjecture justification - interlacing of eigenvalues}
\begin{split}
{\lambda_{\mathcal{S}_{p+1}}^{({\rm sort}[p+1])}} \le {\lambda_{\mathcal{S}_{p}}^{({\rm sort}[p])}} \le {\lambda_{\mathcal{S}_{p+1}}^{({\rm sort}[p])}} \le {\lambda_{\mathcal{S}_{p}}^{({\rm sort}[p-1])} } \le \dots
\\
\dots \le {\lambda_{\mathcal{S}_{p+1}}^{({\rm sort}[2])}} \le {\lambda_{\mathcal{S}_{p}}^{({\rm sort}[1])}} \le {\lambda_{\mathcal{S}_{p+1}}^{({\rm sort}[1])}}
\end{split}
\end{equation}
where the eigenvalues of each of the matrices are referred to in a sorted order, namely, 
\begin{equation}
\label{eq:conjecture justification - sorted eigenvalues - p+1 covariance}
{\lambda_{\mathcal{S}_{p+1}}^{({\rm sort}[p+1])}}  \le {\lambda_{\mathcal{S}_{p+1}}^{({\rm sort}[p])}} \le \dots \le {\lambda_{\mathcal{S}_{p+1}}^{({\rm sort}[2])}}  \le {\lambda_{\mathcal{S}_{p+1}}^{({\rm sort}[1])}}
\end{equation}
are the sorted eigenvalues of $\mtx{C}_{\vec{x},\mathcal{S}_{p+1}}$, and 
\begin{equation}
\label{eq:conjecture justification - sorted eigenvalues - p covariance}
{\lambda_{\mathcal{S}_{p}}^{({\rm sort}[p])}}  \le {\lambda_{\mathcal{S}_{p}}^{({\rm sort}[p-1])}} \le \dots \le {\lambda_{\mathcal{S}_{p}}^{({\rm sort}[2])}}  \le {\lambda_{\mathcal{S}_{p}}^{({\rm sort}[1])}}
\end{equation}
are the sorted eigenvalues of $\mtx{C}_{\vec{x},\mathcal{S}_{p}}$. 

The interlaced structure of the eigenvalue inequalities in (\ref{eq:conjecture justification - interlacing of eigenvalues}) provides an interesting aspect to the analysis of the desired inequality in (\ref{eq:conjecture justification - error decrease inequality - eigendecomposition form}). To see this, we rearrange (\ref{eq:conjecture justification - error decrease inequality - eigendecomposition form}) to rely on the sorted indexing of (\ref{eq:conjecture justification - sorted eigenvalues - p+1 covariance})-(\ref{eq:conjecture justification - sorted eigenvalues - p covariance}) and change the order of the nested summations, namely, the inequality under question (\ref{eq:conjecture justification - error decrease inequality - eigendecomposition form}) becomes 
\begin{equation}
\label{eq:conjecture justification - error decrease inequality - eigendecomposition form - 2}
\begin{split}
\sum_{j=1}^{p} { \alpha_{p}^{(j)} {\lambda_{\mathcal{S}_{p}}^{({\rm sort}[j])}}  } \le  \sum_{j=1}^{p+1} \alpha_{p+1}^{(j)} {\lambda_{\mathcal{S}_{p+1}}^{({\rm sort}[j])}}
\end{split}
\end{equation}
where 
\begin{align}
\label{eq:conjecture justification - error decrease inequality - eigendecomposition form - coefficients definition}
& \alpha_{p}^{(j)} \triangleq \sum_{i\in\widehat{\mathcal{S}}_{p, \rm{max}}^{(k)}} {\left| \left\langle{ {\vecgreek{\psi}}_{\mathcal{S}_{p}}^{({\rm sort}[j])}, \widehat{\vecgreek{\psi}}_{\mathcal{S}_{p}}^{(i)} }\right\rangle \right|}^2  
\nonumber \\
&\text{for}~ j=1,\dots,p, ~\text{and}
\nonumber\\
&\alpha_{p+1}^{(j)} \triangleq { \sum_{i\in\widehat{\mathcal{S}}_{p+1, \rm{max}}^{(k)}}  {\left| \left\langle{ {\vecgreek{\psi}}_{\mathcal{S}_{p+1}}^{({\rm sort}[j])}, \widehat{\vecgreek{\psi}}_{\mathcal{S}_{p+1}}^{(i)} }\right\rangle \right|}^2 }  \nonumber \\ 
&\text{for}~ j=1,\dots,p+1.
\end{align}
The value of $\alpha_{p}^{(j)}$ reflects the quality of approximating the true eigenvector ${\vecgreek{\psi}}_{\mathcal{S}_{p}}^{({\rm sort}[j])}$ by the set of $k$ sample eigenvectors $\left\{ { \widehat{\vecgreek{\psi}}_{\mathcal{S}_{p}}^{(i)} }\right\}_{i\in\widehat{\mathcal{S}}_{p, \rm{max}}^{(k)}}$. The value of $\alpha_{p+1}^{(j)}$ has a similar meaning (with respect to $\mathcal{S}_{p+1}$).

Note that $\alpha_{p}^{(j)}$ and $\alpha_{p+1}^{(j)}$ are values in the range $[0,1]$. However, since (\ref{eq:conjecture justification - error decrease inequality - eigendecomposition form - coefficients definition}) depends on the true and sample eigenvectors of covariance matrices and their submatrices, its characterization is very complex. To generally understand the difficulty in the mathematical analysis of (\ref{eq:conjecture justification - error decrease inequality - eigendecomposition form - coefficients definition}), one can examine the study of the eigenvalue-eigenvector relations provided in \cite{denton2019eigenvectors} that, although being simpler than our case, leads to intricate expressions that are under current research.

The above analysis leads us to choose an empirical approach for justifying our statement on the decay of the out-of-sample error $\mathcal{E}_{\rm out} ^{\rm unsup}  \left( \widehat{\mtx{U}}_{k} ; \mathcal{S}_{p} \right)$ with the increase in the number of features $p$. The experiment settings, referring to the data model provided in Section 2 of the main text, are as follows. 
The data vectors are of dimension $d = 128$ and only $n = 70$ examples are given. 
The linear subspace in the noisy linear data model is of dimension $m = 40$, which is also the number of columns of $\mtx{U}_m$. Each of the experiments below consider one of the following structures for columns of $\mtx{U}_m$: 
\begin{itemize}
    \item The first $m=40$ normalized columns of the $d\times d$ \textit{Hadamard} matrix (these normalized columns are, by definition, orthonormnal).
    \item $m=40$ \textit{random} orthonormal vectors that are a subset of the left singular vectors of a $d\times d$ Gaussian matrix of i.i.d. components $\mathcal{N}(0,1)$.
\end{itemize}
These Hadamard and random constructions are \textit{global} in the sense that they are defined using all the $d$ coordinates of the feature space. However, unlike the random form, the Hadamard form has a deterministic structure.
In all the settings  $ \vec{z} \sim \mathcal{N}\left( \vec{0}, \mtx{I}_{m} \right)$, but we consider two different levels of noise (that is represented by the variable $\epsilon$ in the data model (1)): $\sigma_{\epsilon}=0.1$ and $\sigma_{\epsilon}=0.5$. 

\begin{figure*}[t]
\centering
\subfloat[]{\includegraphics[width=0.65\columnwidth]{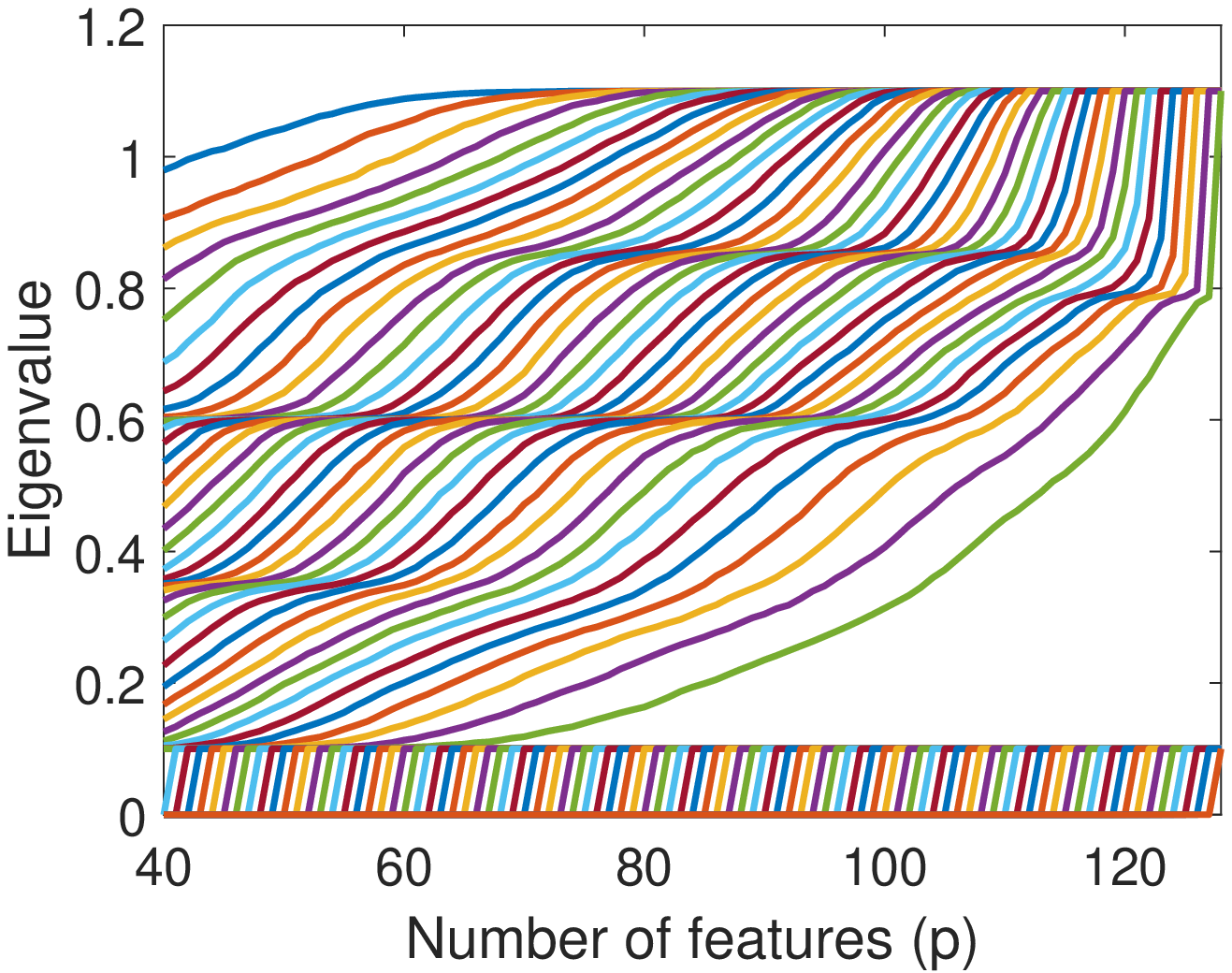}\label{fig_appendix:hadamard_m40_k40_noise01_sorted_eigenvalues}}
\subfloat[]{\includegraphics[width=0.65\columnwidth]{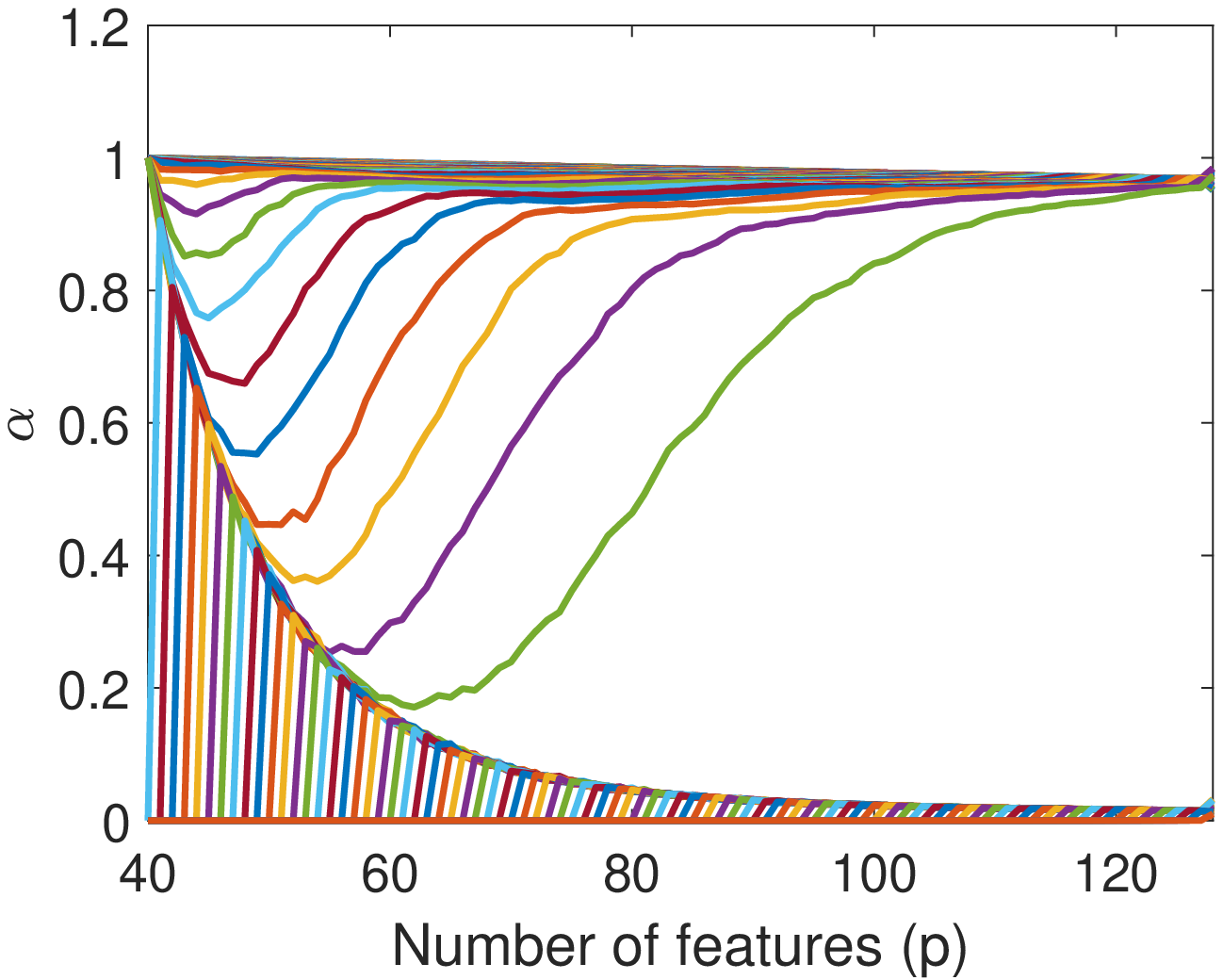}\label{fig_appendix:hadamard_m40_k40_noise01_sorted_coefficients}}
\subfloat[]{\includegraphics[width=0.65\columnwidth]{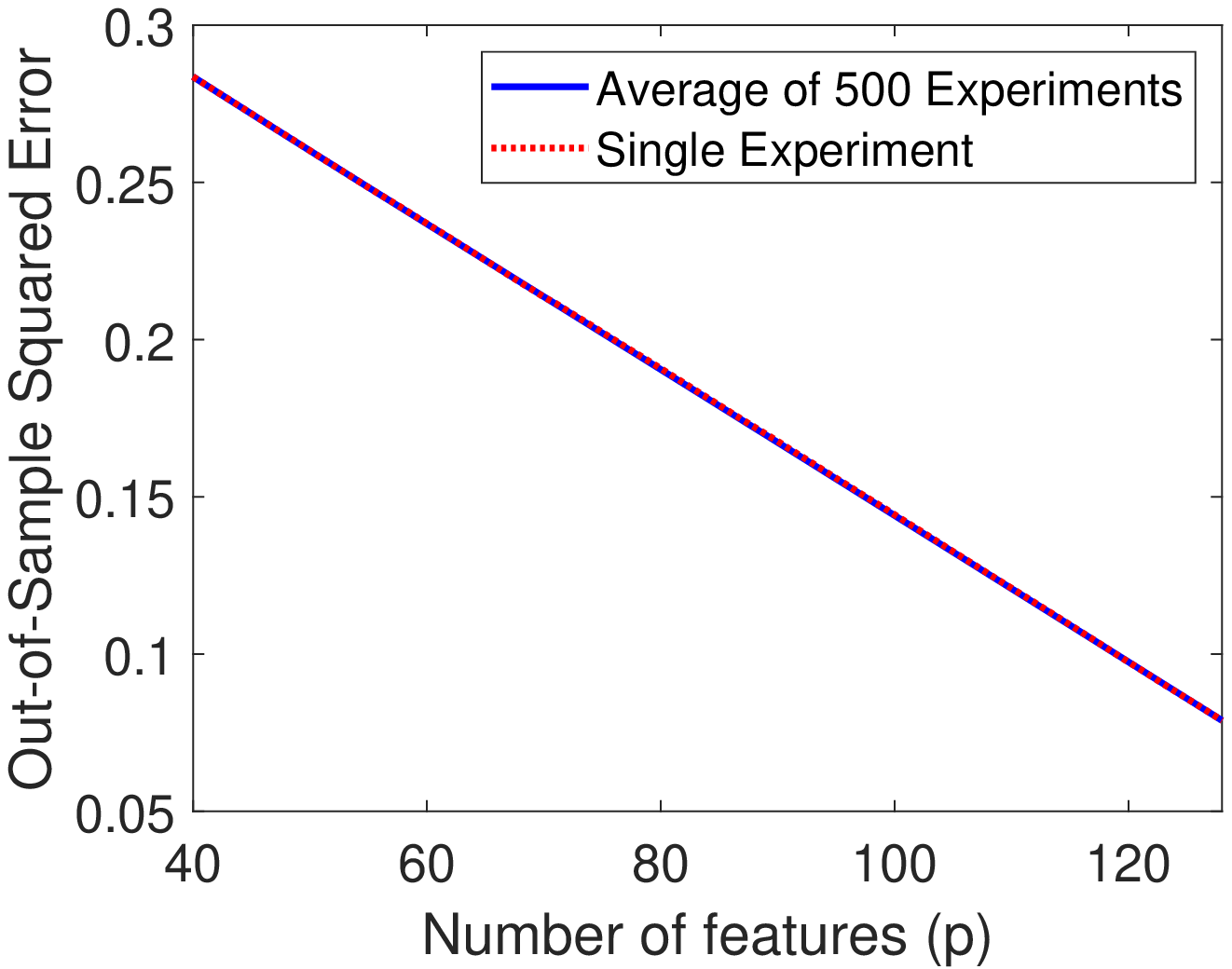}\label{fig_appendix:hadamard_m40_k40_noise01_errors}}
\\
\subfloat[]{\includegraphics[width=0.65\columnwidth]{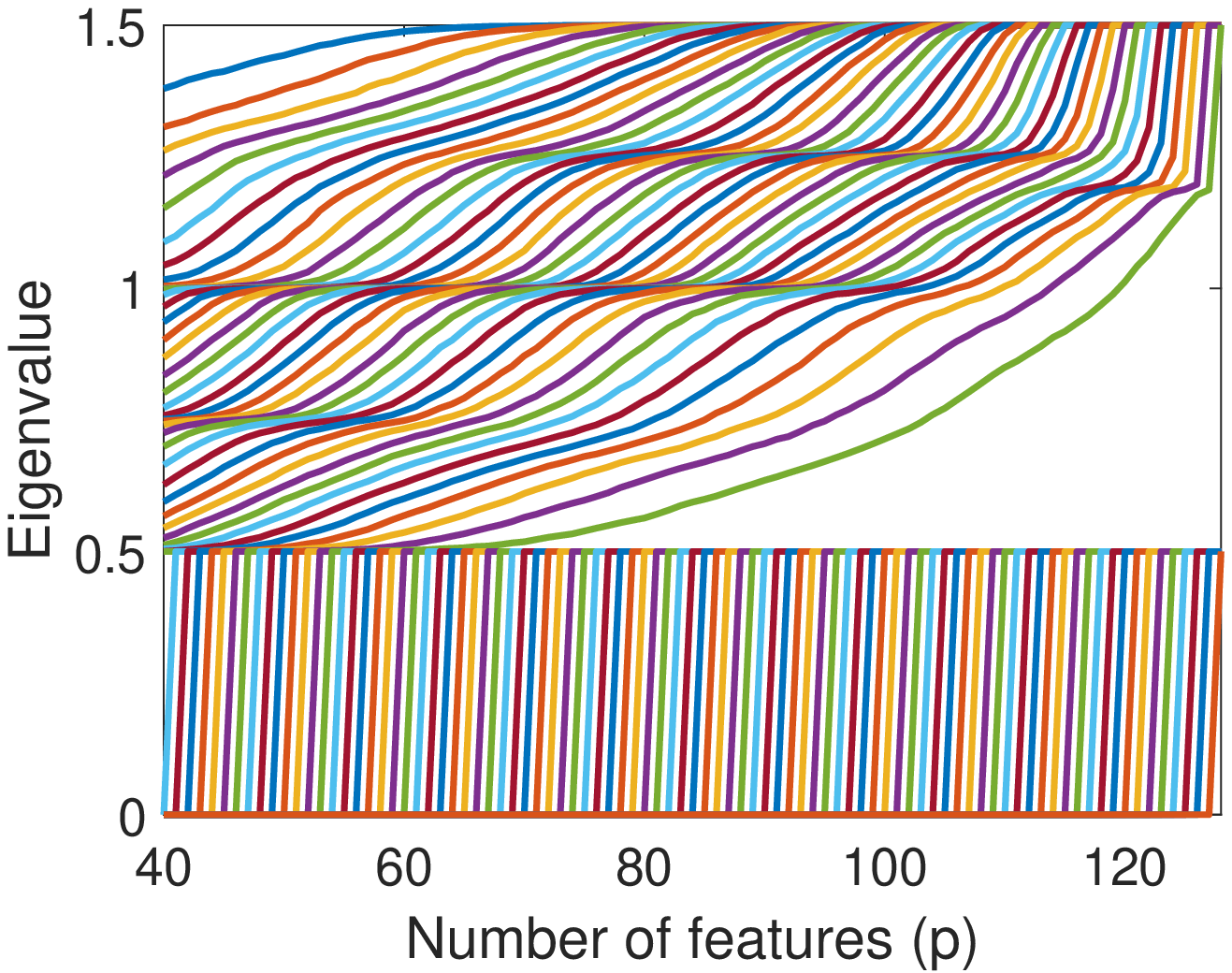}\label{fig_appendix:hadamard_m40_k40_noise05_sorted_eigenvalues}}
\subfloat[]{\includegraphics[width=0.65\columnwidth]{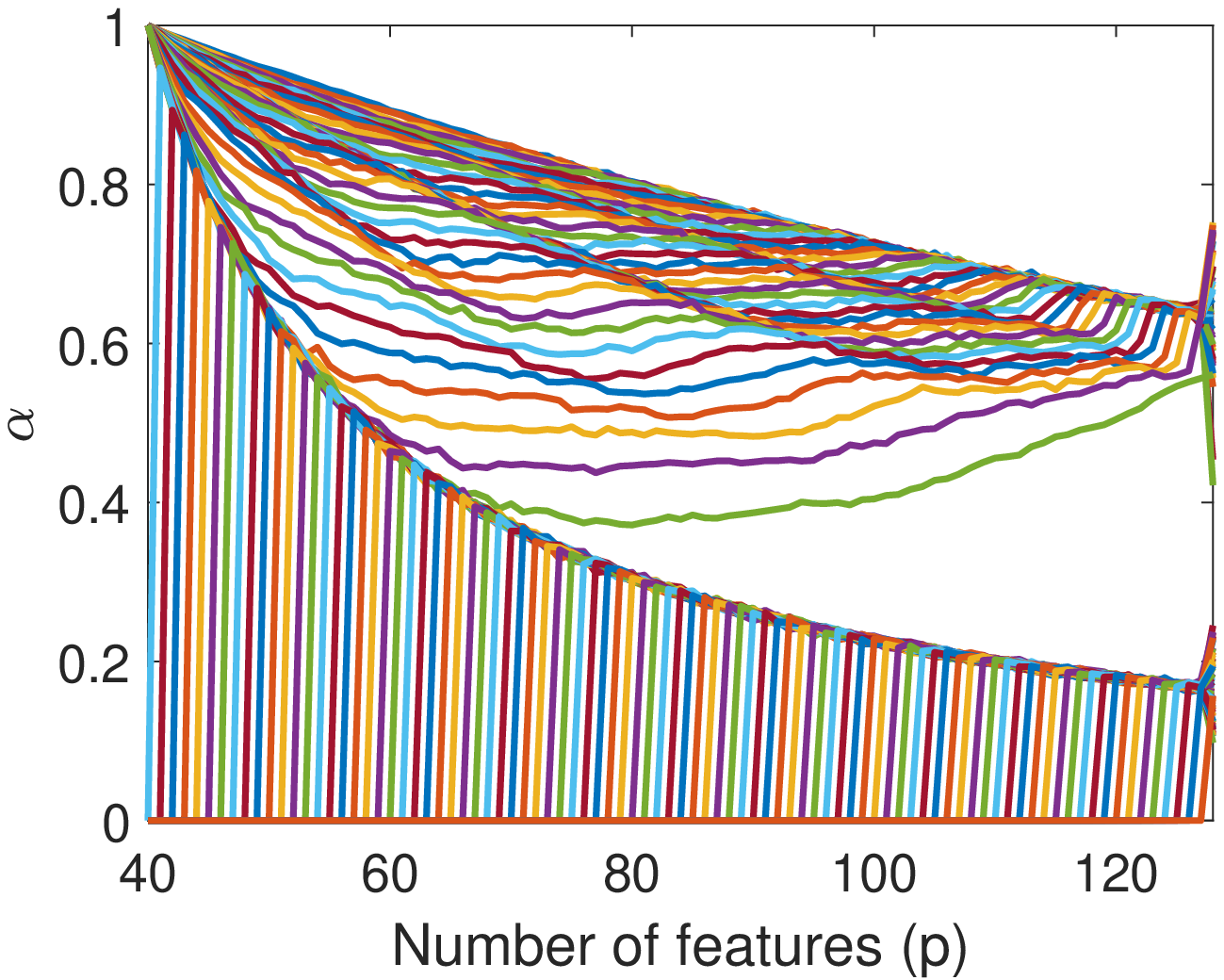}\label{fig_appendix:hadamard_m40_k40_noise05_sorted_coefficients}}
\subfloat[]{\includegraphics[width=0.65\columnwidth]{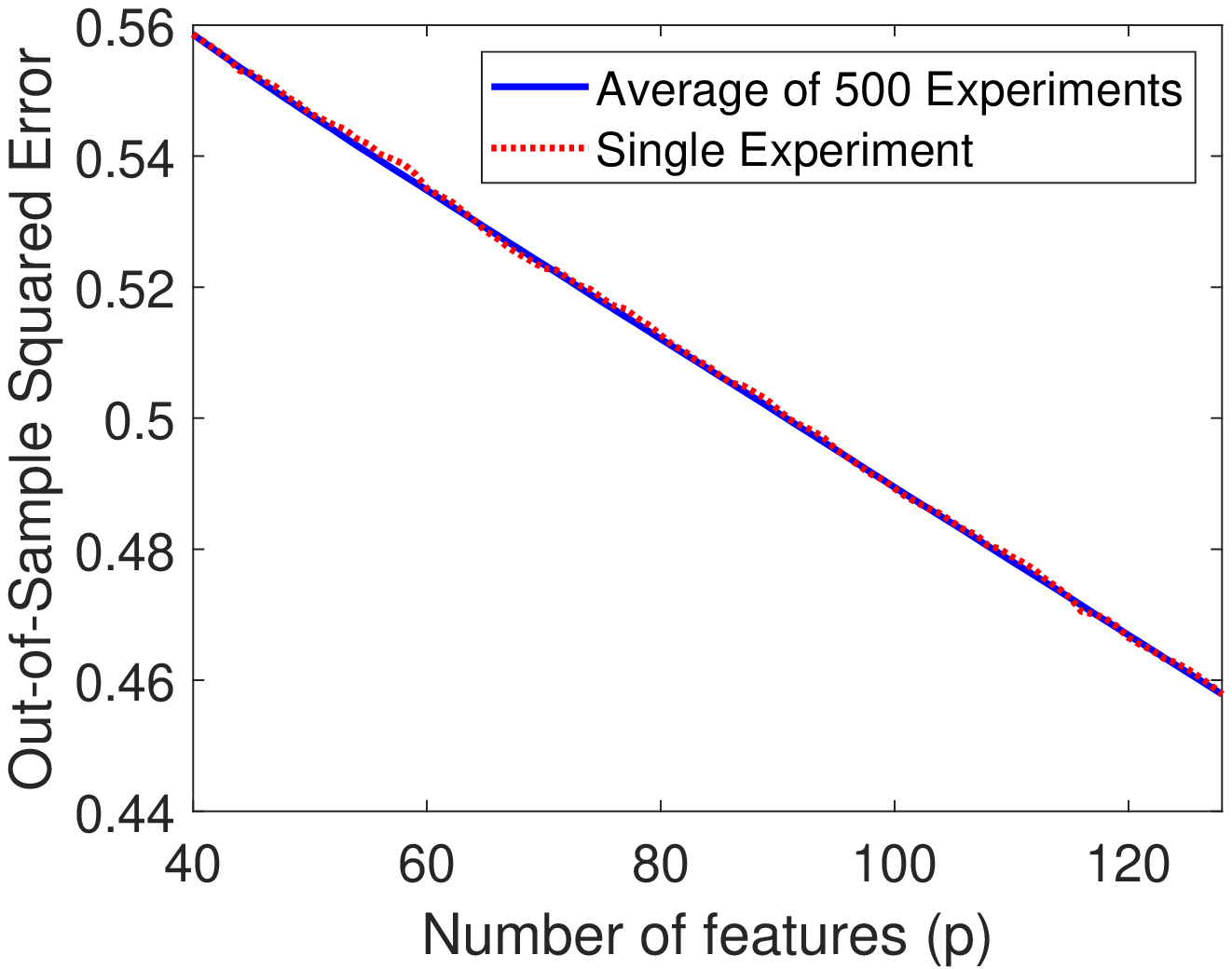}\label{fig_appendix:hadamard_m40_k40_noise05_errors}}
\\
\subfloat[]{\includegraphics[width=0.65\columnwidth]{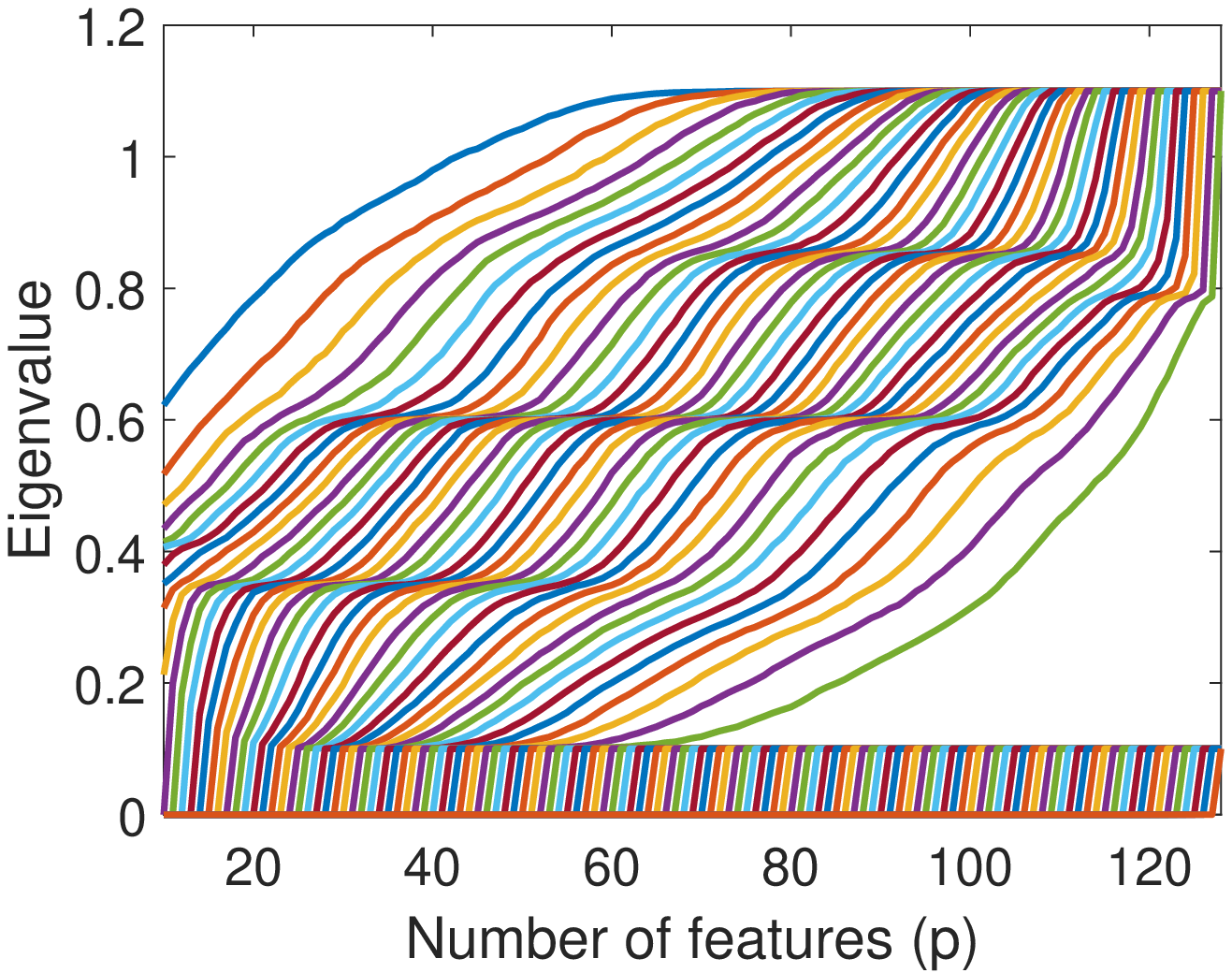}\label{fig_appendix:hadamard_m40_k10_noise01_sorted_eigenvalues}}
\subfloat[]{\includegraphics[width=0.65\columnwidth]{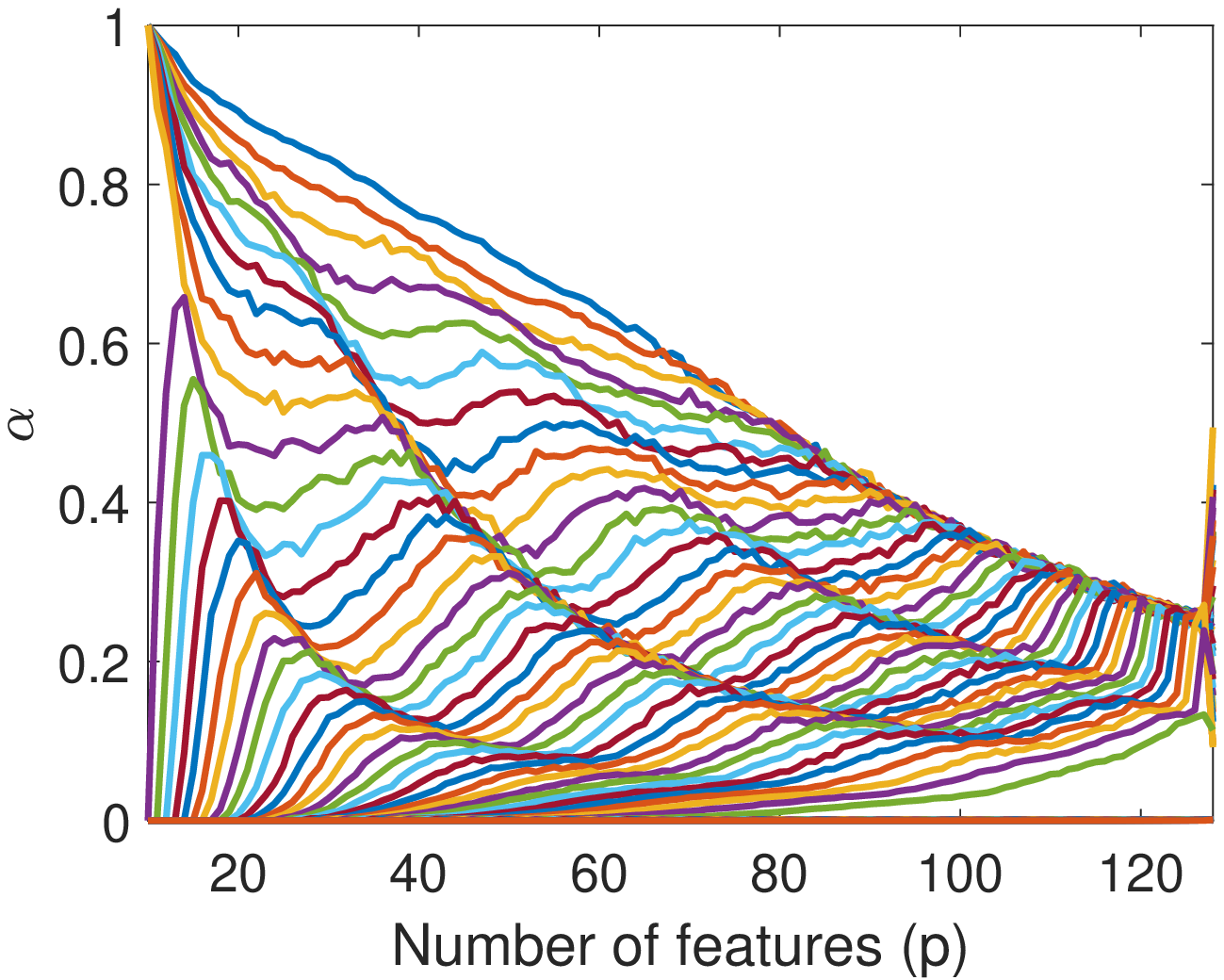}\label{fig_appendix:hadamard_m40_k10_noise01_sorted_coefficients}}
\subfloat[]{\includegraphics[width=0.65\columnwidth]{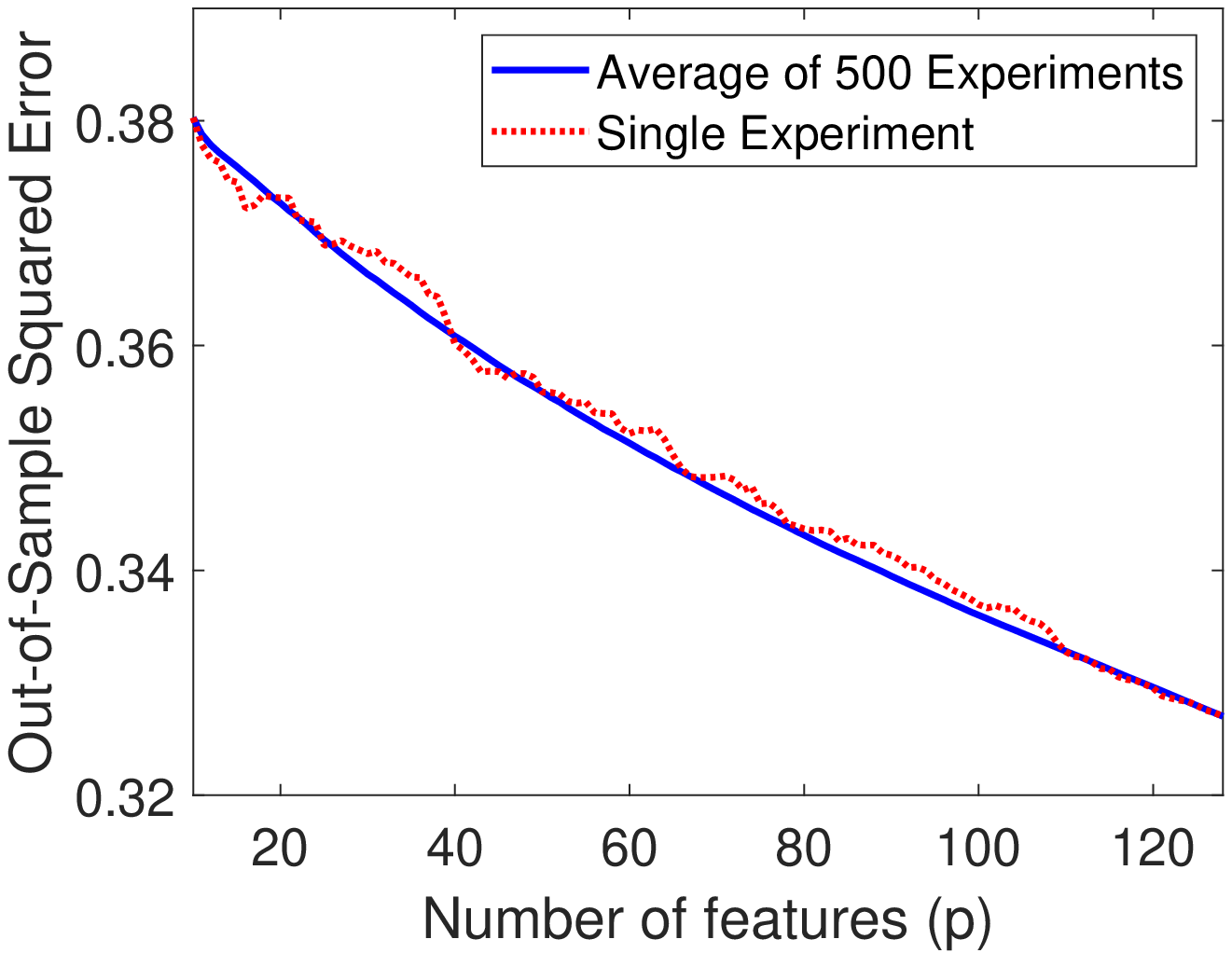}\label{fig_appendix:hadamard_m40_k10_noise01_errors}}
\caption{Empirical demonstration of the evolution of the components in (\ref{eq:conjecture justification - error decrease inequality - eigendecomposition form - 2}) and the corresponding out-of-sample error $\mathcal{E}_{\rm out} ^{\rm unsup}  \left( \widehat{\mtx{U}}_{k} ; \mathcal{S}_{p} \right)$, and their evolution with the increase in the number of features $p$. 
Each line of subfigures corresponds to a different experimental setting, yet, for all of them the true subspace is of the \textbf{Hadamard} form, $d=128$, $m=40$, and $n=70$. The first line of subfigures considers $k=m=40$ and a noise level of  $\sigma_{\epsilon} = 0.1$. The second line of subfigures corresponds to $k=m=40$ and a noise level of $\sigma_{\epsilon} = 0.5$. 
The third line of subfigures corresponds to $k=10$ and a  noise level of  $\sigma_{\epsilon} = 0.1$. 
(a), (d) and (g) present the sorted eigenvalues ${\lambda_{\mathcal{S}_{p}}^{({\rm sort}[j])}}$ of the true covariance matrices corresponding to $p$-feature vectors (each of the curves corresponds to another value of $j$).
(b), (e) and (h) show the (sorted) coefficients $\alpha_{p}^{(j)}$ 
 defined in (\ref{eq:conjecture justification - error decrease inequality - eigendecomposition form - coefficients definition}).
(c), (f) and (i) exhibit the out-of-sample error $\mathcal{E}_{\rm out} ^{\rm unsup}  \left( \widehat{\mtx{U}}_{k} ; \mathcal{S}_{p} \right)$ for a single instance of sequential increase of $\mathcal{S}_p$ (dotted red line) and for average over 500 different orders of sequentially increasing $\mathcal{S}_p$ (solid blue line). 
}
\label{fig_appendix:hadamard_specific_examples}
\end{figure*}

For a start, we exemplify the evolution of the eigenvalues $\left\{ { {\lambda_{\mathcal{S}_{p}}^{({\rm sort}[j])}} }\right\}_{j=1}^{p}$ with $p$. We consider three different settings as described in the caption of Fig.~\ref{fig_appendix:hadamard_specific_examples}.   Figures \ref{fig_appendix:hadamard_m40_k40_noise01_sorted_eigenvalues}, \ref{fig_appendix:hadamard_m40_k40_noise05_sorted_eigenvalues}, \ref{fig_appendix:hadamard_m40_k10_noise01_sorted_eigenvalues} clearly show the monotonic increase explained by the application of Cauchy's interlacing theorem in (\ref{eq:conjecture justification - interlacing of eigenvalues}). The corresponding behavior of $\left\{ { {\alpha_{p}^{(j)}} }\right\}_{j=1}^{p}$ (see Figures \ref{fig_appendix:hadamard_m40_k40_noise01_sorted_coefficients}, \ref{fig_appendix:hadamard_m40_k40_noise05_sorted_coefficients}, \ref{fig_appendix:hadamard_m40_k10_noise01_sorted_coefficients}) is indeed intricate as mentioned above. Specifically, Fig.~\ref{fig_appendix:hadamard_m40_k40_noise05_sorted_coefficients} shows the effect of an increased noise level. Fig.~\ref{fig_appendix:hadamard_m40_k10_noise01_sorted_coefficients} demonstrates the consequence of estimating a subspace of an incorrect dimension. 
Despite the complex behavior of $\left\{ { {\alpha_{p}^{(j)}} }\right\}_{j=1}^{p}$, Figures \ref{fig_appendix:hadamard_m40_k40_noise01_errors}, \ref{fig_appendix:hadamard_m40_k40_noise05_errors}, \ref{fig_appendix:hadamard_m40_k10_noise01_errors} present that the resulting out-of-sample errors monotonically decrease \textit{on average} (where $\mathcal{S}_{p}$ is uniformly chosen at random) with the increase in $p$ (see solid blue curves in Figs. \ref{fig_appendix:hadamard_m40_k40_noise01_errors}, \ref{fig_appendix:hadamard_m40_k40_noise05_errors}, \ref{fig_appendix:hadamard_m40_k10_noise01_errors}). This is explained next.

\begin{figure*}[t]
\centering
\subfloat[]{\includegraphics[width=0.65\columnwidth]{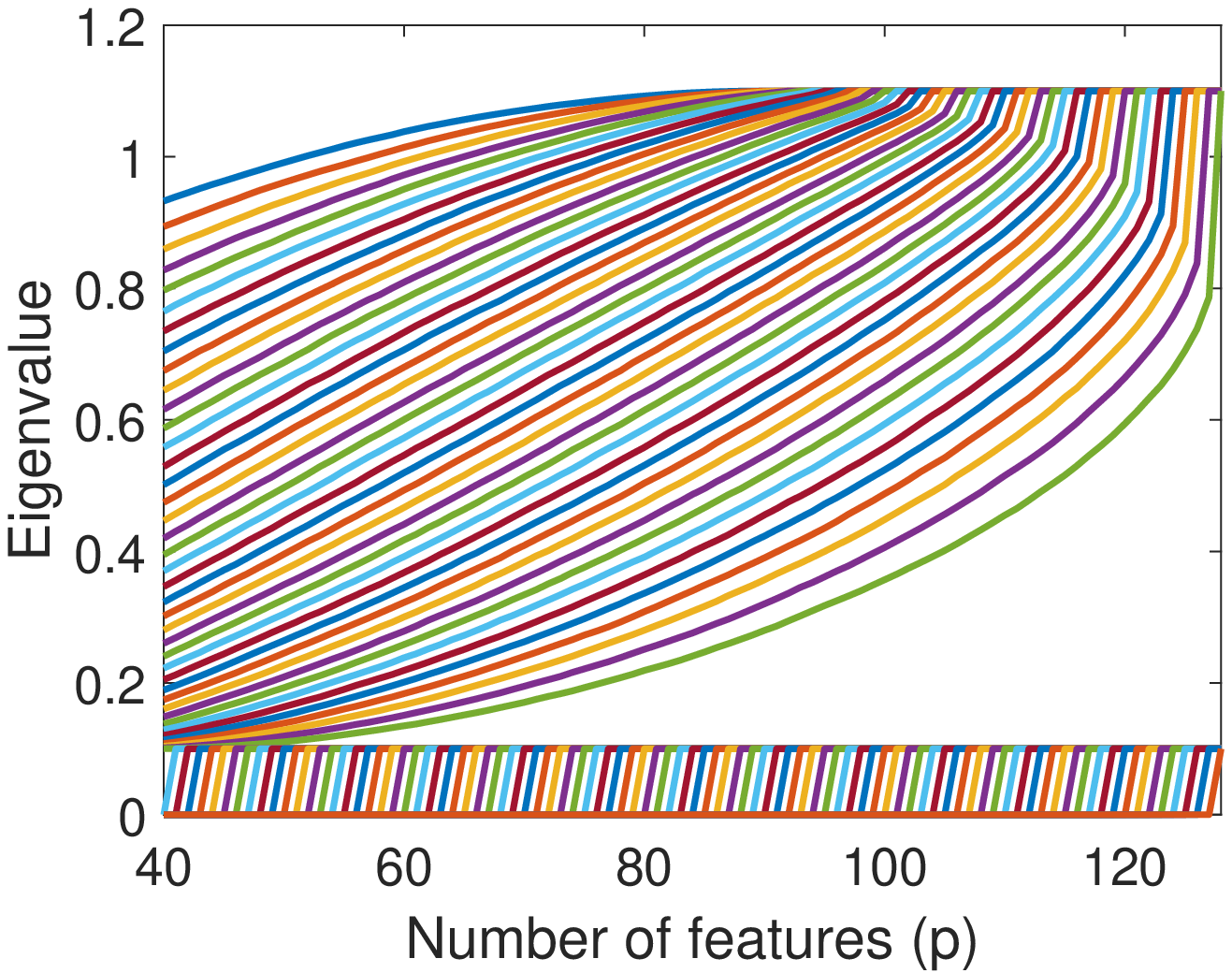}\label{fig_appendix:random_m40_k40_noise01_sorted_eigenvalues}}
\subfloat[]{\includegraphics[width=0.65\columnwidth]{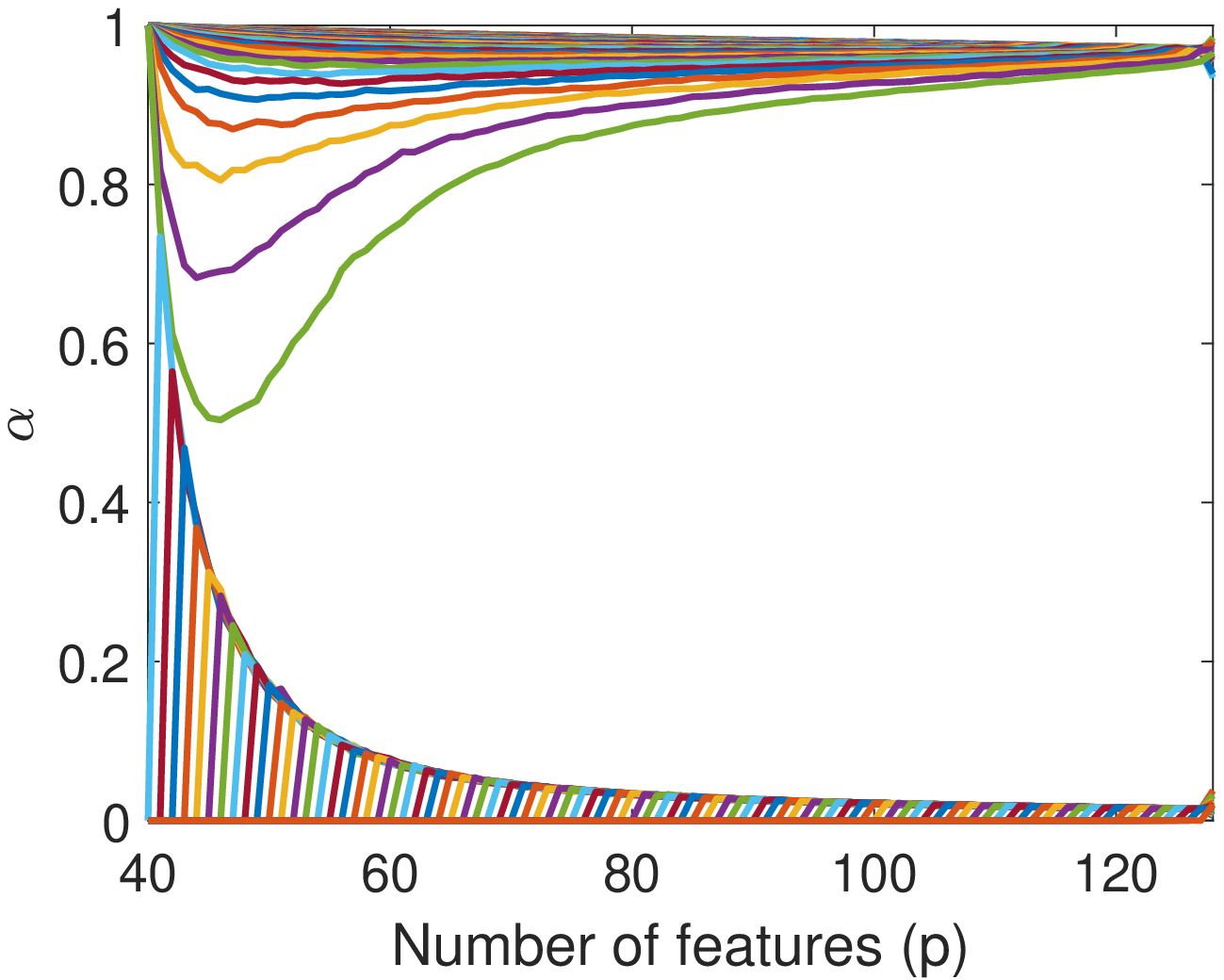}\label{fig_appendix:random_m40_k40_noise01_sorted_coefficients}}
\subfloat[]{\includegraphics[width=0.65\columnwidth]{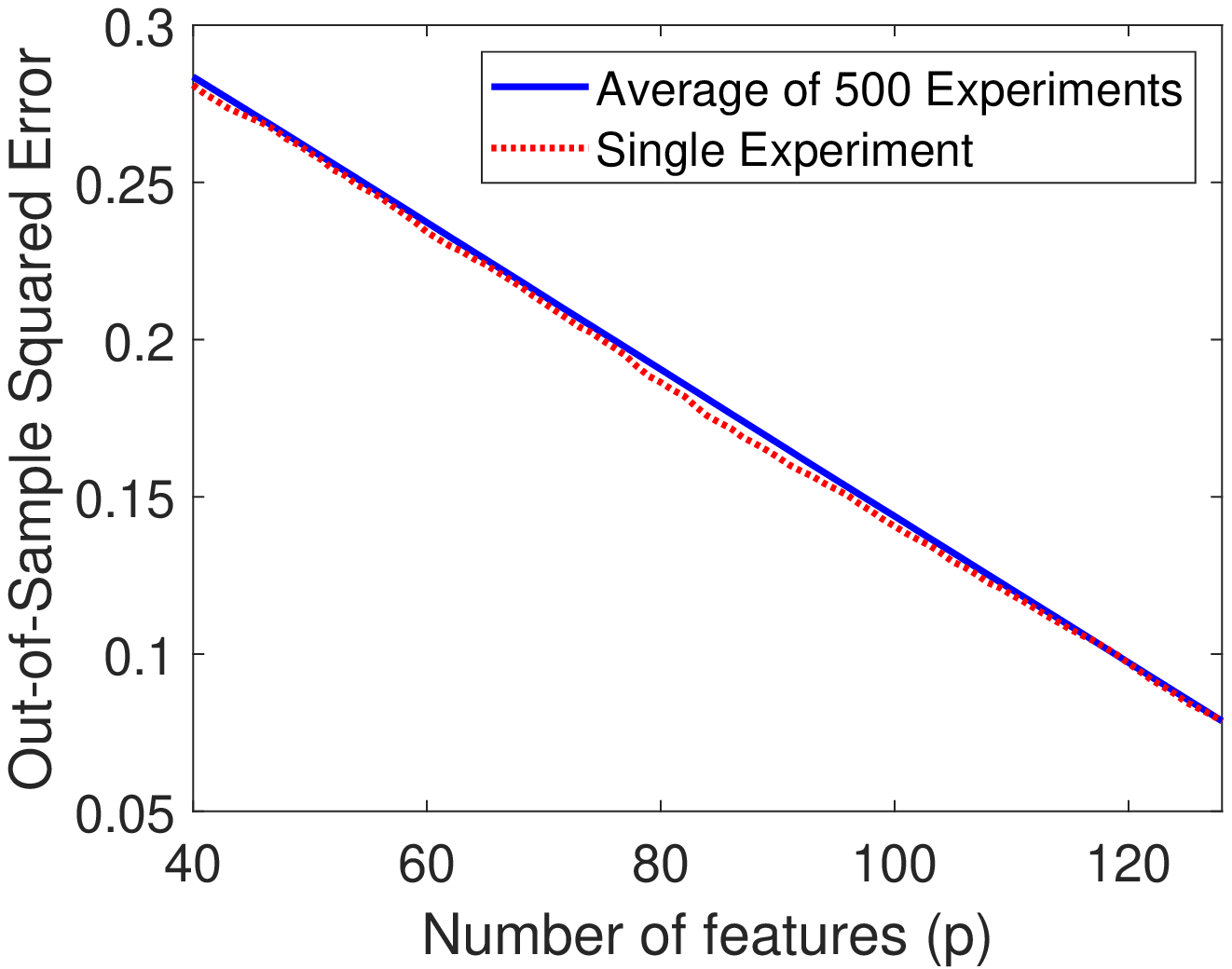}\label{fig_appendix:random_m40_k40_noise01_errors}}
\\
\subfloat[]{\includegraphics[width=0.65\columnwidth]{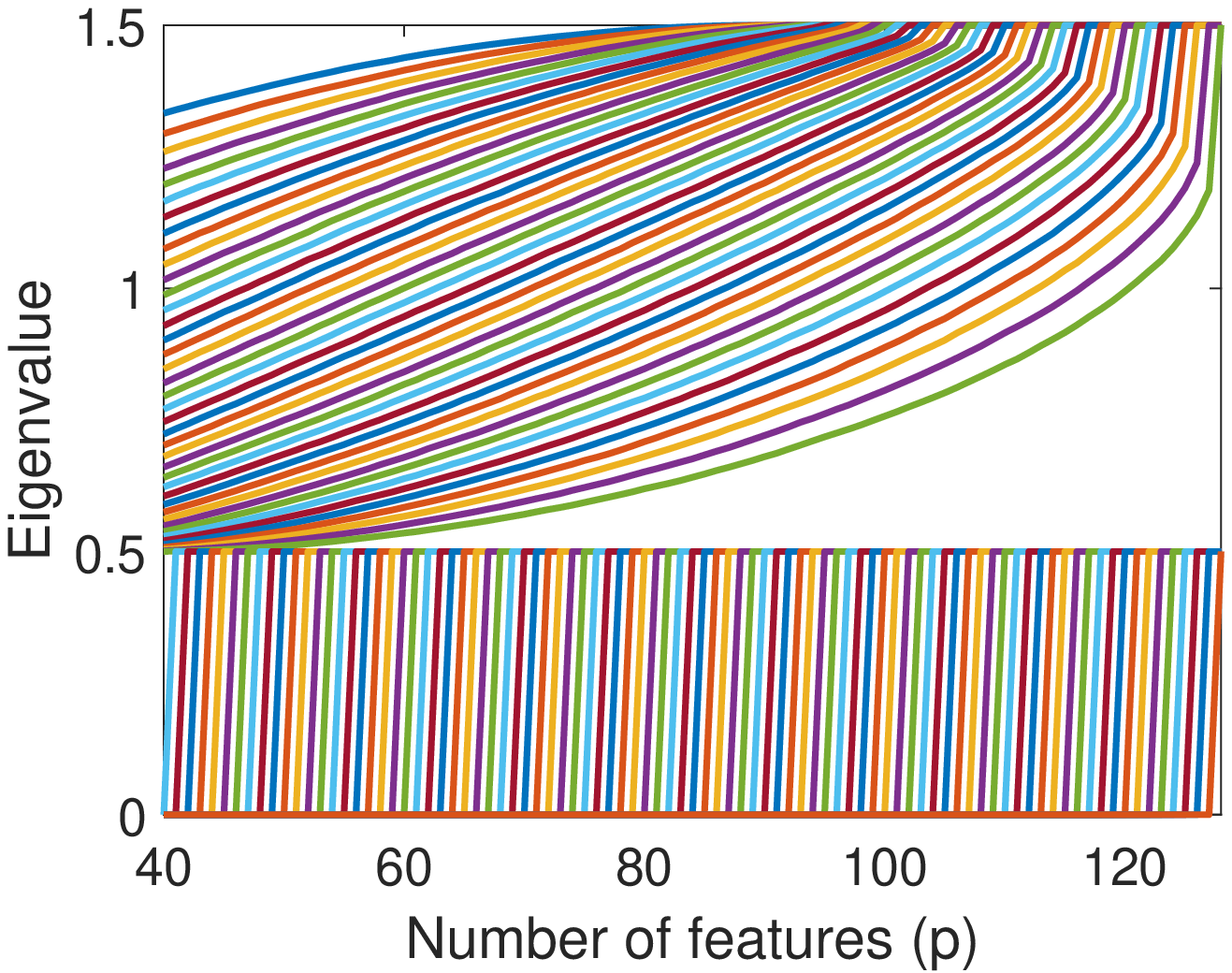}\label{fig_appendix:random_m40_k40_noise05_sorted_eigenvalues}}
\subfloat[]{\includegraphics[width=0.65\columnwidth]{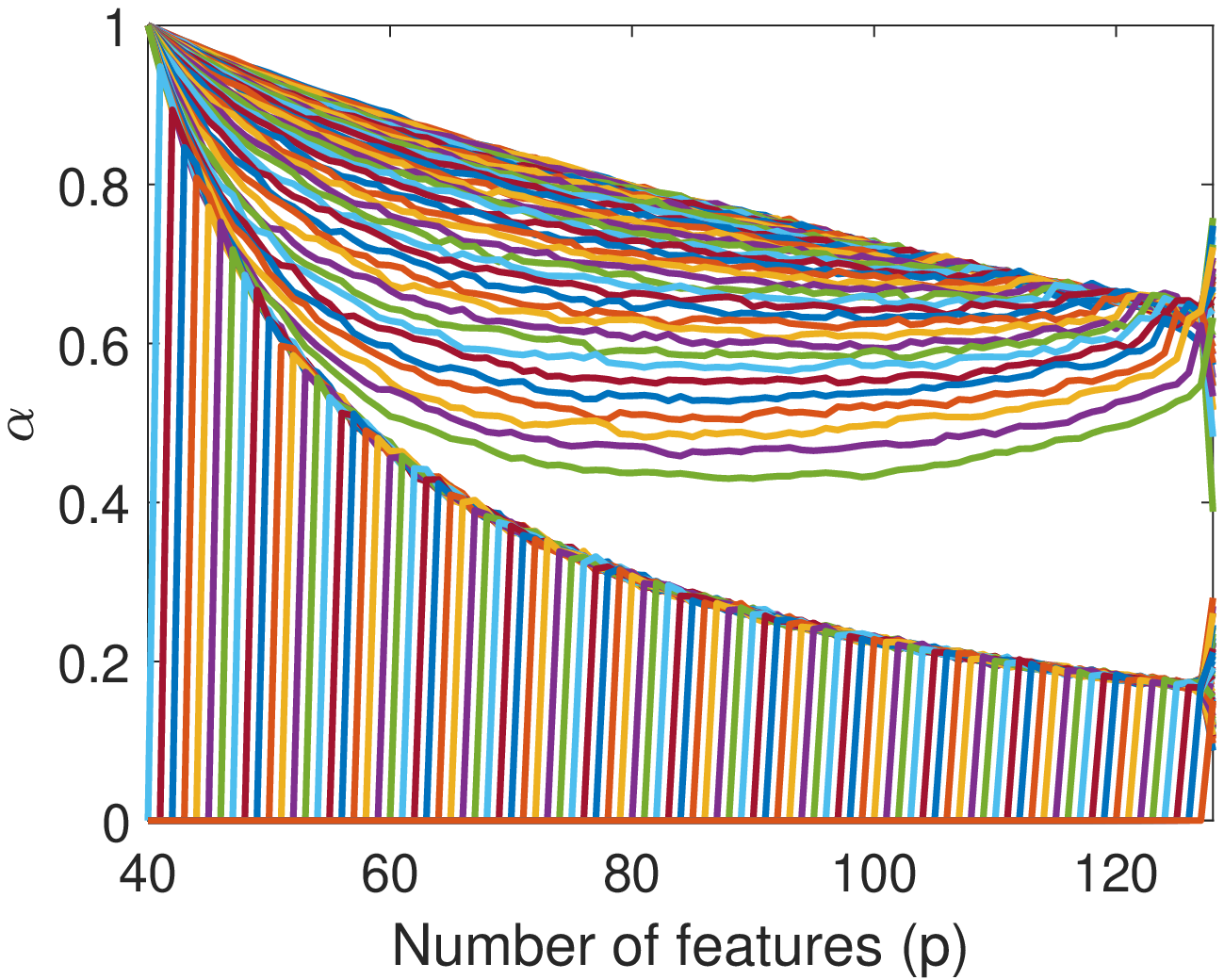}\label{fig_appendix:random_m40_k40_noise05_sorted_coefficients}}
\subfloat[]{\includegraphics[width=0.65\columnwidth]{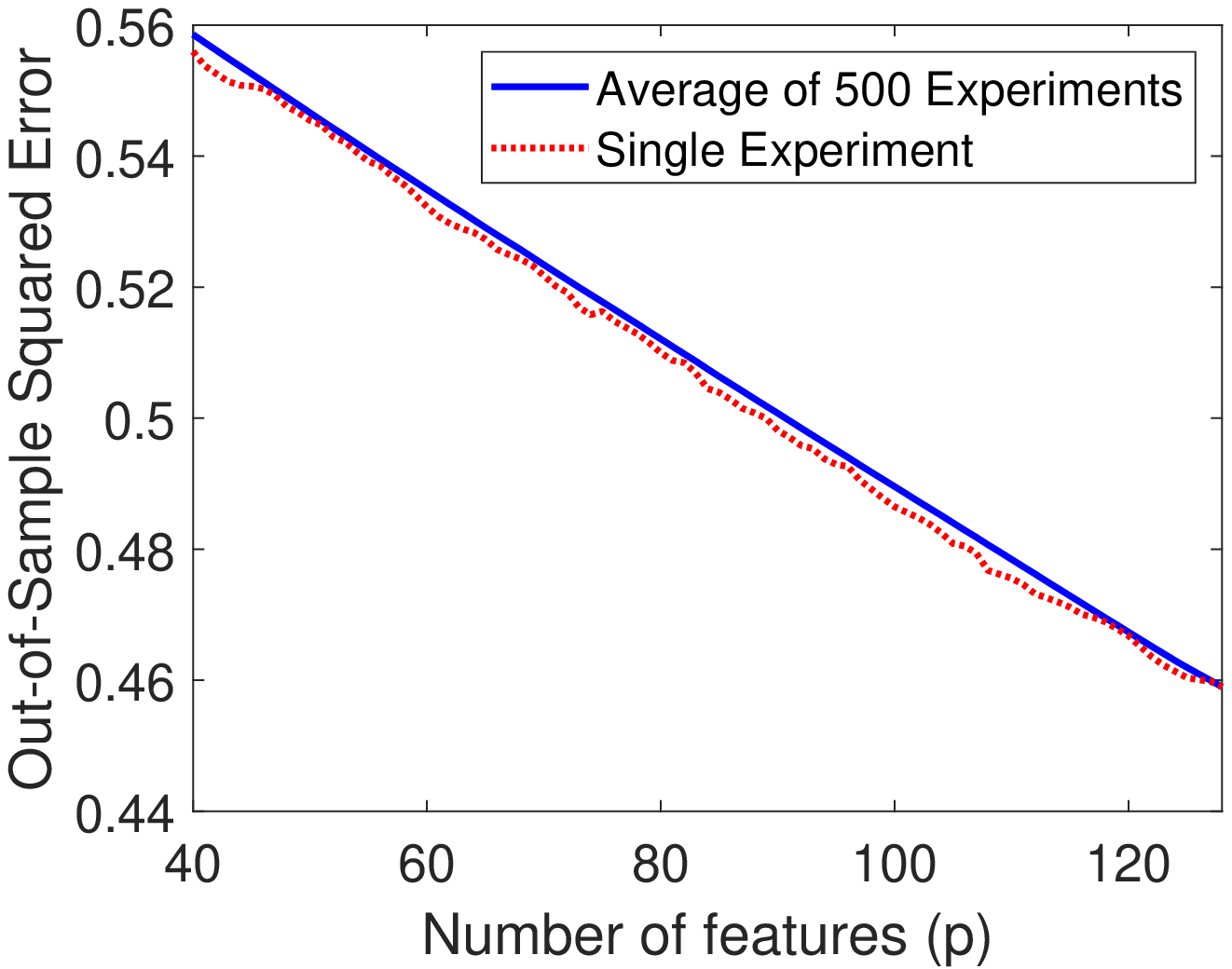}\label{fig_appendix:random_m40_k40_noise05_errors}}
\\
\subfloat[]{\includegraphics[width=0.65\columnwidth]{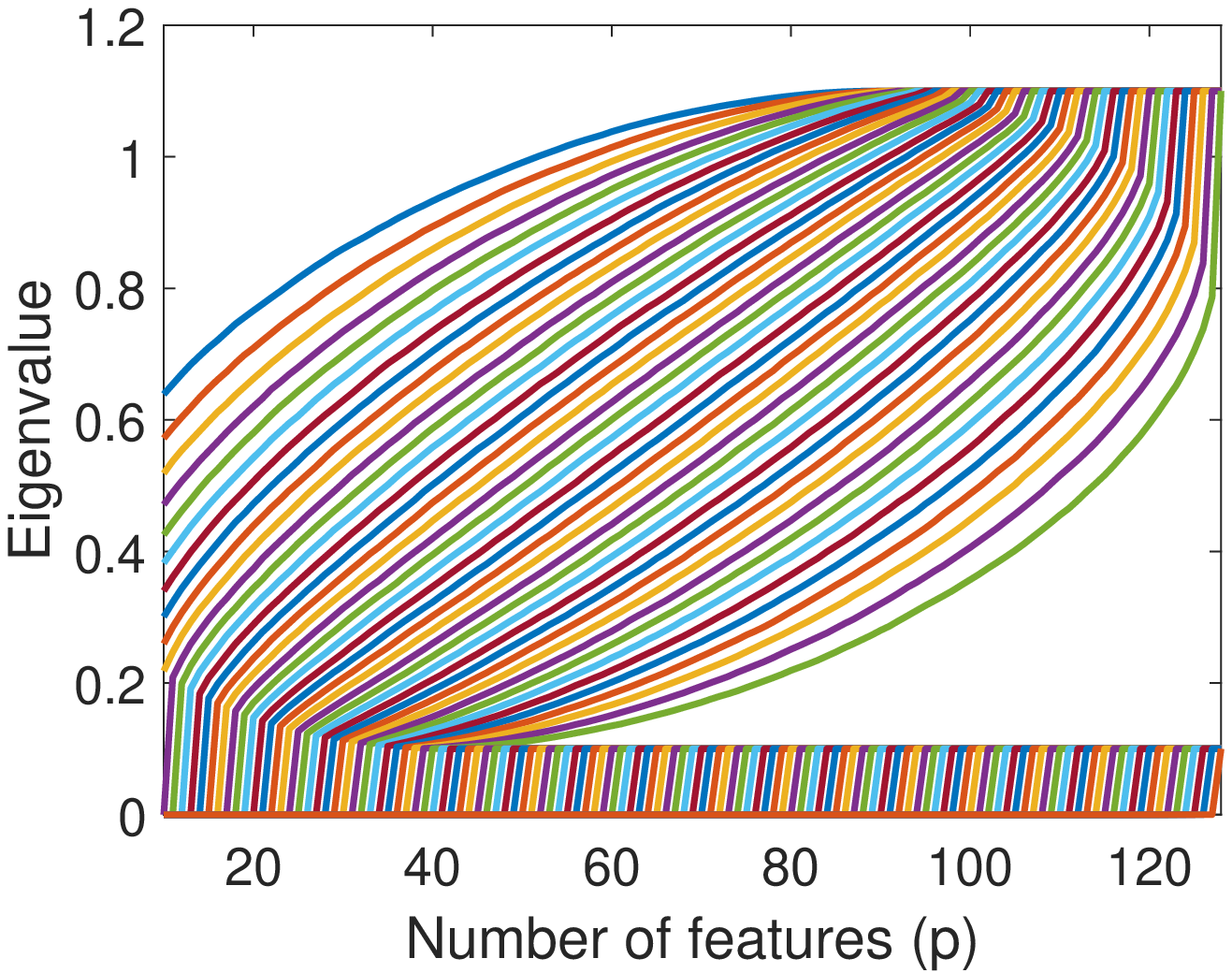}\label{fig_appendix:random_m40_k10_noise01_sorted_eigenvalues}}
\subfloat[]{\includegraphics[width=0.65\columnwidth]{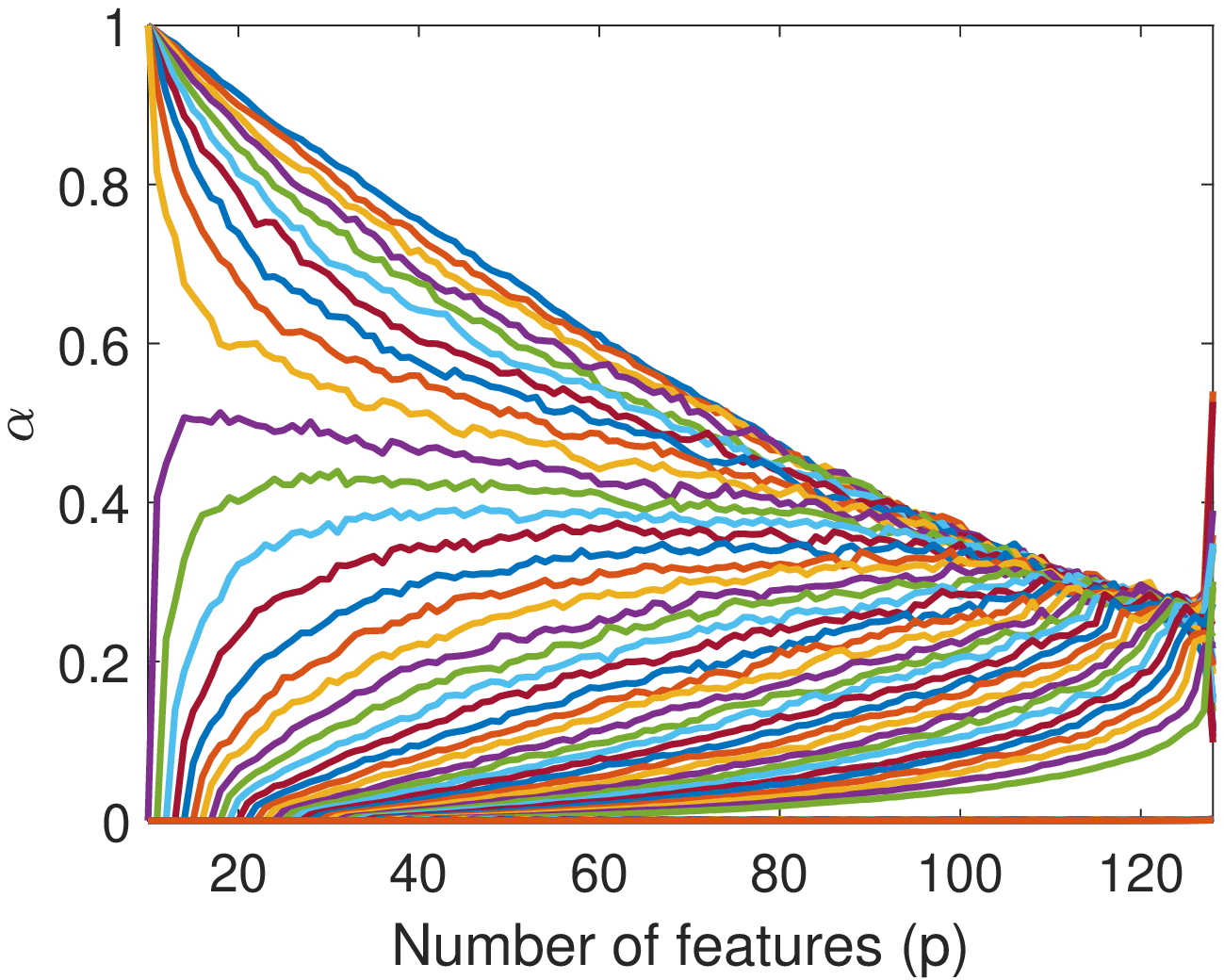}\label{fig_appendix:random_m40_k10_noise01_sorted_coefficients}}
\subfloat[]{\includegraphics[width=0.65\columnwidth]{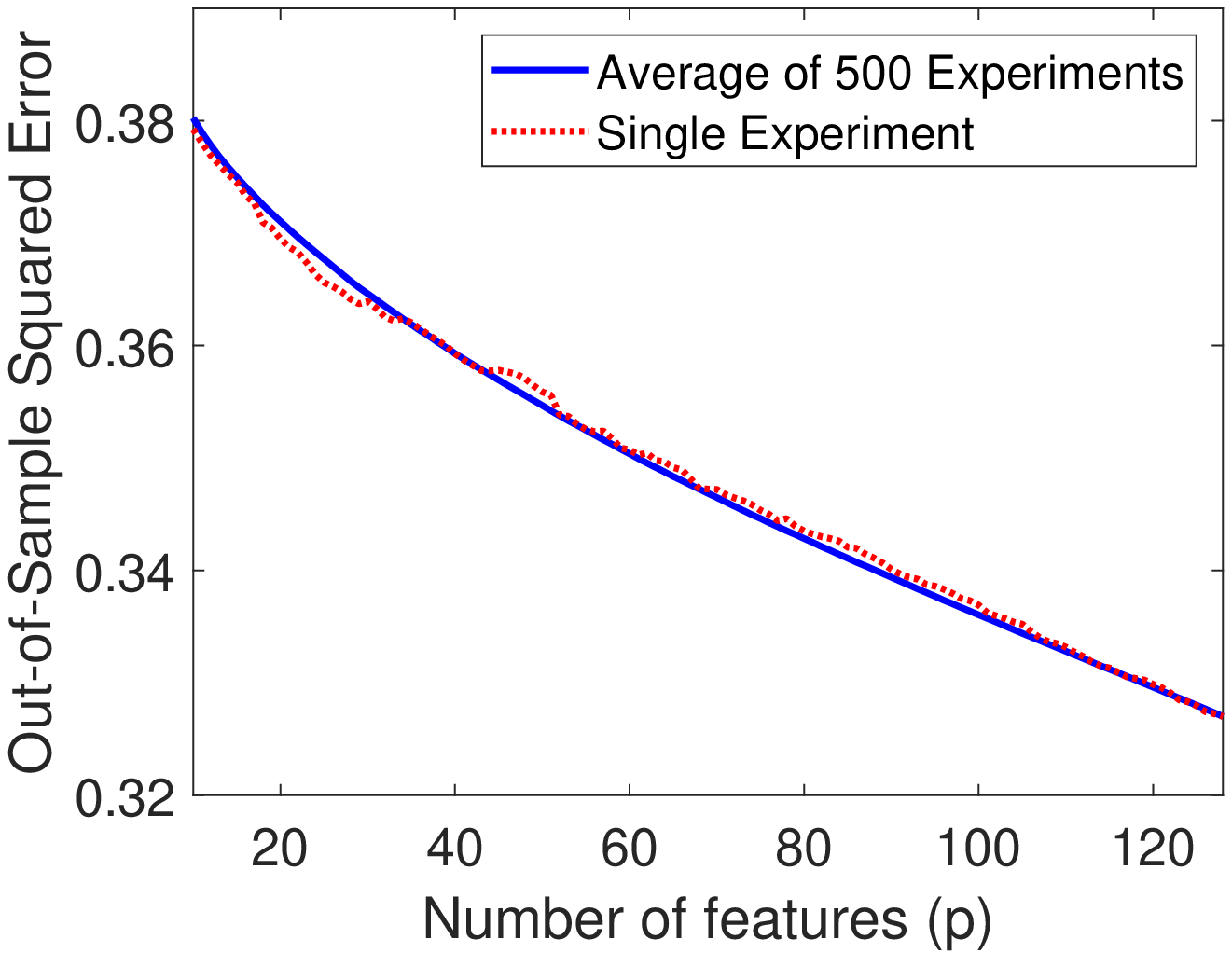}\label{fig_appendix:random_m40_k10_noise01_errors}}
\caption{Empirical demonstration of the evolution of the components in (\ref{eq:conjecture justification - error decrease inequality - eigendecomposition form - 2}) and the corresponding out-of-sample error $\mathcal{E}_{\rm out} ^{\rm unsup}  \left( \widehat{\mtx{U}}_{k} ; \mathcal{S}_{p} \right)$, and their evolution with the increase in the number of features $p$. 
Each line of subfigures corresponds to a different experimental setting, yet, for all of them the true subspace is of the \textbf{random} form, $d=128$, $m=40$, and $n=70$. The first line of subfigures considers $k=m=40$ and a noise level of  $\sigma_{\epsilon} = 0.1$. The second line of subfigures corresponds to $k=m=40$ and a noise level of $\sigma_{\epsilon} = 0.5$. 
The third line of subfigures corresponds to $k=10$ and a  noise level of  $\sigma_{\epsilon} = 0.1$. 
(a), (d) and (g) present the sorted eigenvalues ${\lambda_{\mathcal{S}_{p}}^{({\rm sort}[j])}}$ of the true covariance matrices corresponding to $p$-feature vectors (each of the curves corresponds to another value of $j$).
(b), (e) and (h) show the (sorted) coefficients $\alpha_{p}^{(j)}$ 
 defined in (\ref{eq:conjecture justification - error decrease inequality - eigendecomposition form - coefficients definition}).
(c), (f) and (i) exhibit the out-of-sample error $\mathcal{E}_{\rm out} ^{\rm unsup}  \left( \widehat{\mtx{U}}_{k} ; \mathcal{S}_{p} \right)$ for a single instance of sequential increase of $\mathcal{S}_p$ (dotted red line) and for average over 500 different orders of sequentially increasing $\mathcal{S}_p$ (solid blue line). 
}
\label{fig_appendix:random_specific_examples}
\end{figure*}

We now proceed to the empirical results that explain the decay of the out-of-sample error $\mathcal{E}_{\rm out} ^{\rm unsup}  \left( \widehat{\mtx{U}}_{k} ; \mathcal{S}_{p} \right)$ with the increase in $p$. Figures \ref{fig_appendix:hadamard_m40_k40_noise01_errors}, \ref{fig_appendix:hadamard_m40_k40_noise05_errors} present the evolution of the out-of-sample error for estimated subspaces of dimension $k=m$ (i.e., the true subspace dimension is known) and Fig.~\ref{fig_appendix:hadamard_m40_k10_noise01_errors} corresponds to $k=10<m$ (namely, an incorrect dimension). Each figure contains two curves: the dotted red curves present the sequence of errors $\mathcal{E}_{\rm out} ^{\rm unsup}  \left( \widehat{\mtx{U}}_{k} ; \mathcal{S}_{p} \right)$ induced by a single sequential construction of $\mathcal{S}_p$;  the solid blue curves show the sequence of \textit{averages} over the errors $\mathcal{E}_{\rm out} ^{\rm unsup}  \left( \widehat{\mtx{U}}_{k} ; \mathcal{S}_{p} \right)$ induced by 500 different (and uniformly chosen at random) sequential constructions of $\mathcal{S}_p$.

Figures~\ref{fig_appendix:hadamard_m40_k40_noise01_errors}, \ref{fig_appendix:hadamard_m40_k40_noise05_errors}, \ref{fig_appendix:hadamard_m40_k10_noise01_errors} show that, \textit{on average}, adding features is beneficial and reduces $\mathcal{E}_{\rm out} ^{\rm unsup}  \left( \widehat{\mtx{U}}_{k} ; \mathcal{S}_{p} \right)$. However, for a \textit{specific and arbitrary} order of adding features, there is no guarantee that each added feature is indeed useful (for example, see the dotted red curve in Fig. \ref{fig_appendix:hadamard_m40_k10_noise01_errors} that does not exhibit a monotonic decreasing trend). The results also show that the deviation from monotonicity is larger for higher noise levels and/or significant differences between the dimensions of the estimated and true subspaces.
Corresponding experiments for the \textit{random} subspace setting, are provided in Fig. \ref{fig_appendix:random_specific_examples} and further support the findings of the Hadamard case discussed above.

\begin{figure*}[]
\centering
\subfloat[]{\includegraphics[width=0.5\columnwidth]{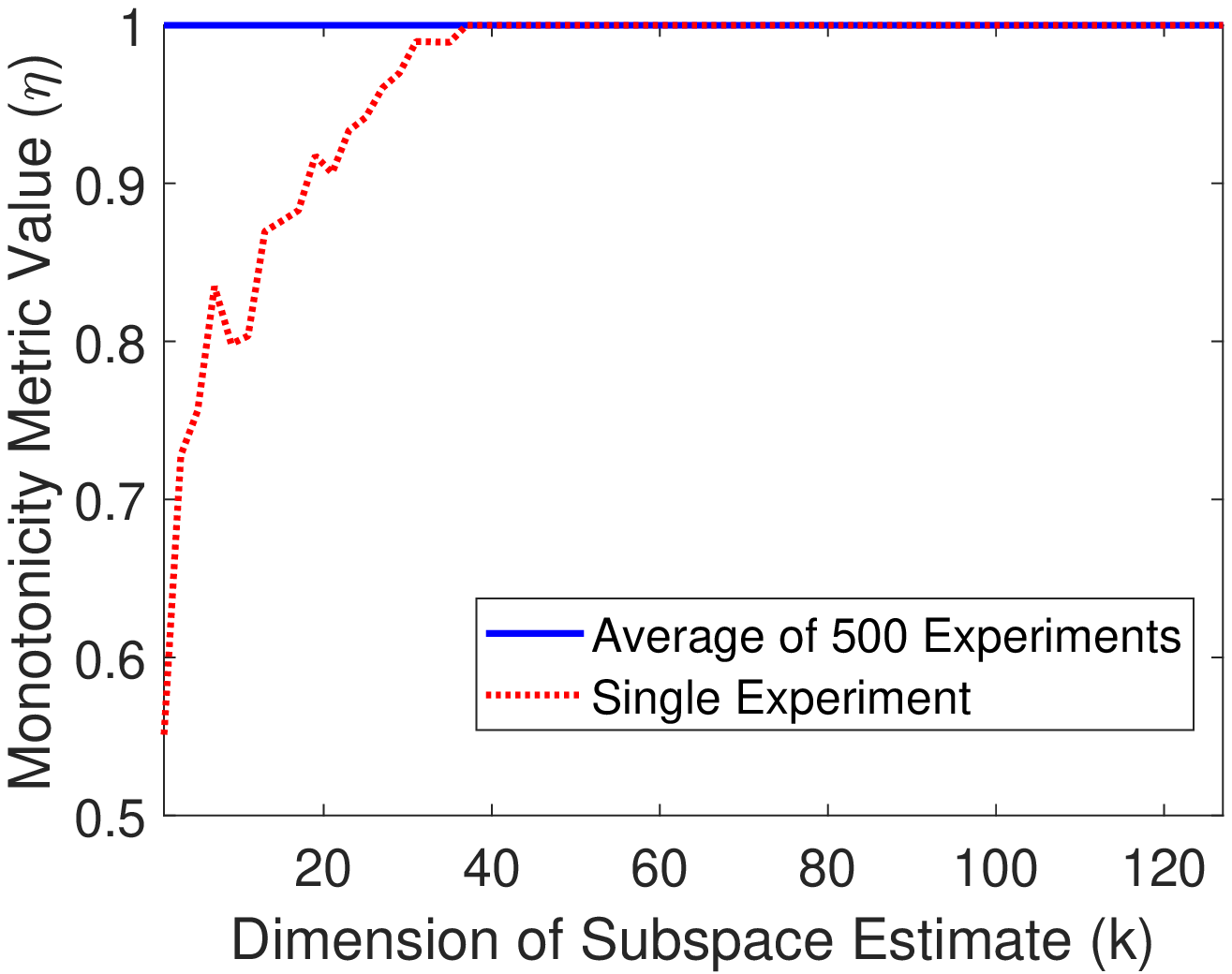}\label{fig_appendix:hadamard_m40_noise01_monotonicity_metric_curve}}
\subfloat[]{\includegraphics[width=0.5\columnwidth]{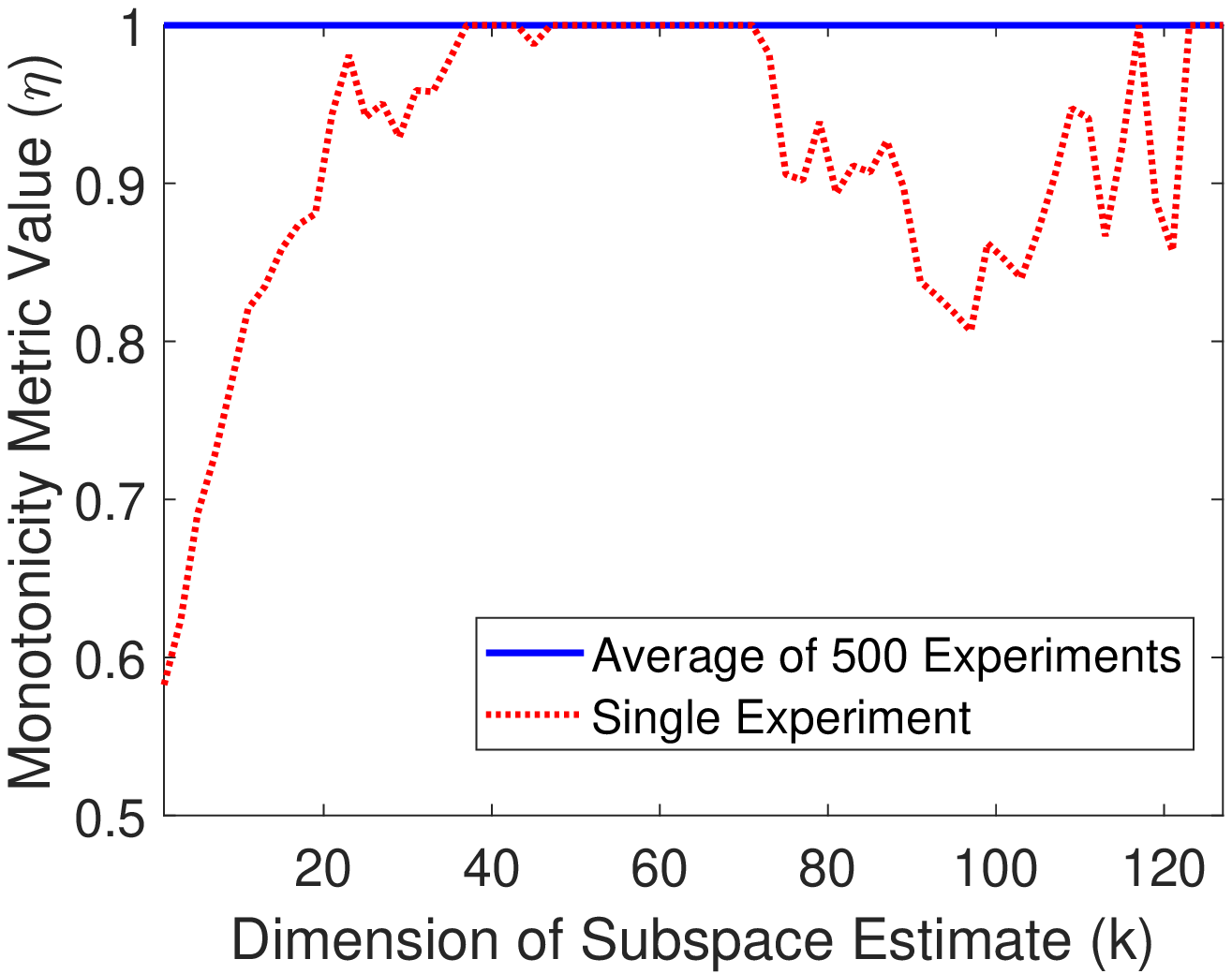}\label{fig_appendix:hadamard_m40_noise05_monotonicity_metric_curve}}
\subfloat[]{\includegraphics[width=0.5\columnwidth]{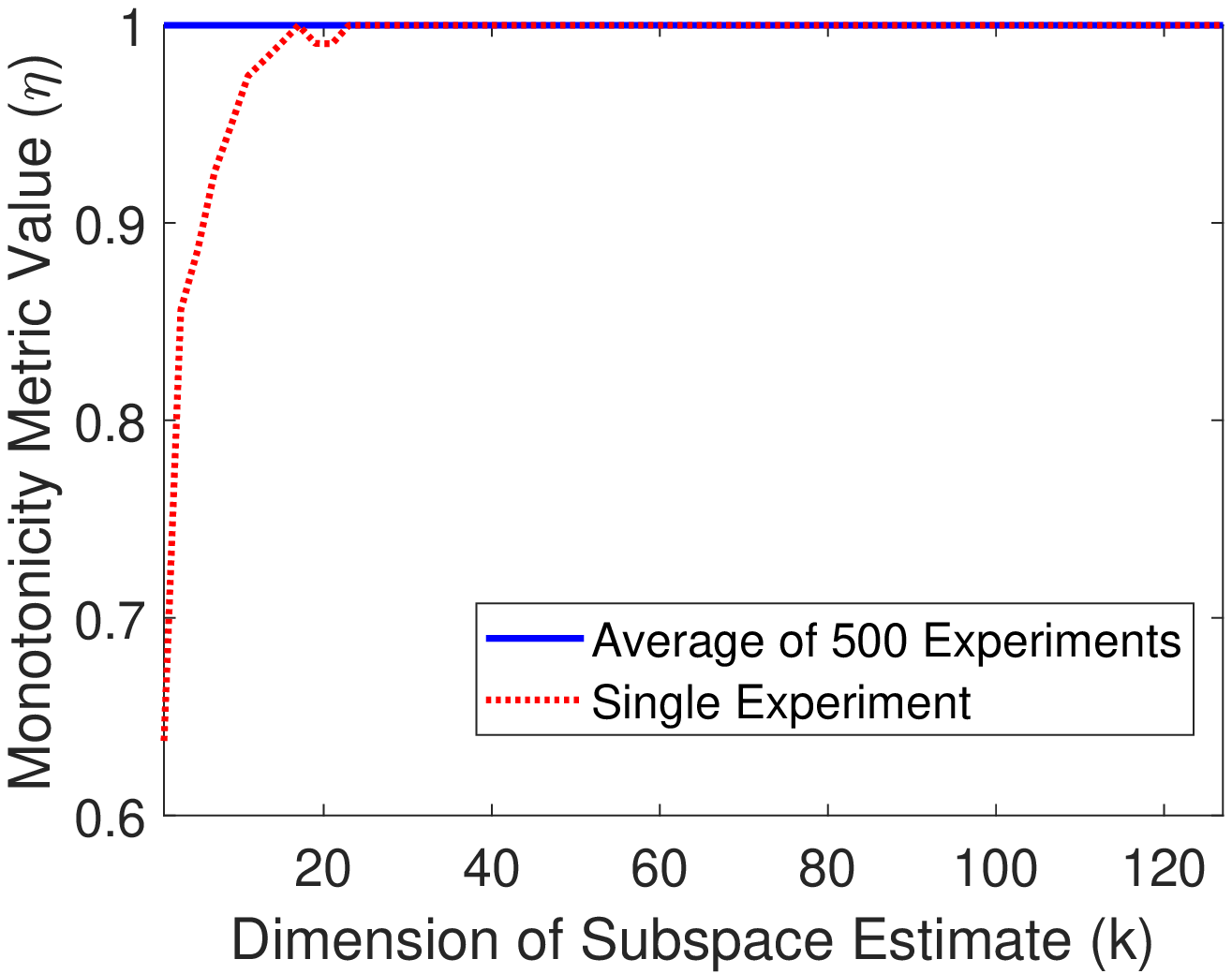}\label{fig_appendix:random_m40_noise01_monotonicity_metric_curve}}
\subfloat[]{\includegraphics[width=0.5\columnwidth]{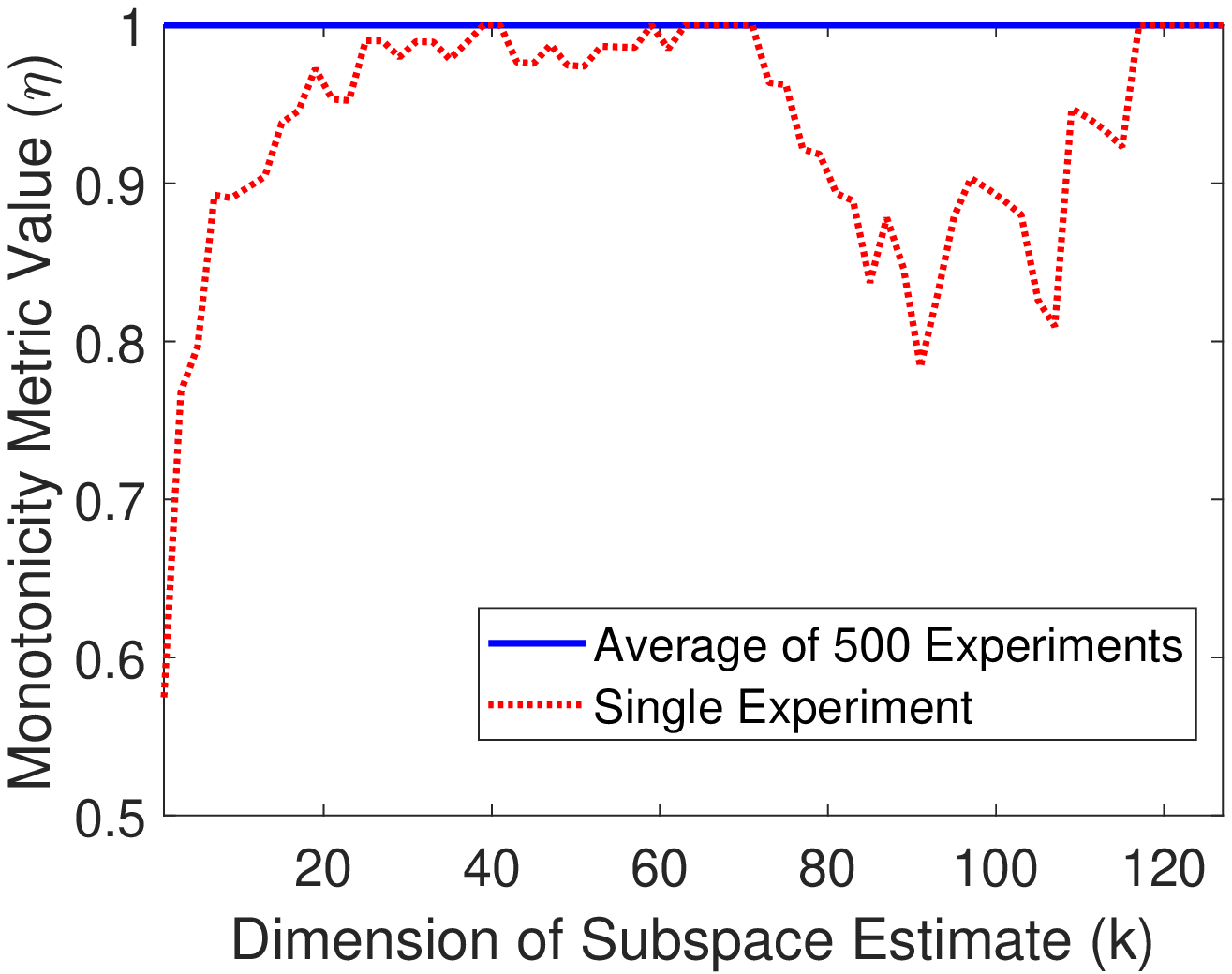}\label{fig_appendix:random_m40_noise05_monotonicity_metric_curve}}
\caption{Empirical evaluation of the monotonicity metric, defined in (\ref{eq:metric for monotonic decrease evaluation}), versus the estimated subspace dimension.
All the evaluated settings correspond to $d=128$, $m=40$, and $n=70$. 
The results in (a) and (b) are for the Hadamard case with noise levels $\sigma_{\epsilon} = 0.1$ and $\sigma_{\epsilon} = 0.5$, respectively. 
The results in (c) and (d) are for the Random subspace construction with noise levels $\sigma_{\epsilon} = 0.1$ and $\sigma_{\epsilon} = 0.5$, respectively. 
The dotted red curves obtained for a single instance of sequential increase of $\mathcal{S}_p$, and the solid blue curves are monotonicity evaluations based on the average out-of-sample errors obtained from 500 different orders of sequentially increasing $\mathcal{S}_p$. 
}
\label{fig_appendix:all_monotonicity_metric_curve}
\end{figure*}

The results in Figures \ref{fig_appendix:hadamard_m40_k40_noise01_errors},\ref{fig_appendix:hadamard_m40_k40_noise05_errors},\ref{fig_appendix:hadamard_m40_k10_noise01_errors} are only for several values of $k$. Therefore, we also present results for the entire range possible for the dimension of the subspace estimate, i.e., $k=1,\dots,d$. This extensive set of experiments is provided in Fig.~\ref{fig_appendix:all_monotonicity_metric_curve} in a summarized form described as follows. We again use the notation emphasizing the dependency of the error on $p$, namely, $\mathcal{E}_{\rm out} ^{\rm unsup}  \left( \widehat{\mtx{U}}_{k} ; \mathcal{S}_{p} \right)$. For the various settings, we are interested in assessing the monotonic decrease of the error curve of $\mathcal{E}_{\rm out} ^{\rm unsup}  \left( \widehat{\mtx{U}}_{k} ; \mathcal{S}_{p} \right)$ over the (discrete) range of $p=k,\dots,d$. Hence, we evaluate the monotonicity of the discrete sequence ${ \left\{ { \mathcal{E}_{\rm out} ^{\rm unsup}  \left( \widehat{\mtx{U}}_{k} ; \mathcal{S}_{j} \right) }\right\}_{j=k}^{d}  }$ by computing the relative number of feature additions that reduced (or kept) the error. Namely, this metric is defined as 
\begin{align}
\label{eq:metric for monotonic decrease evaluation}
& \eta \left( { \left\{ { \mathcal{E}_{\rm out} ^{\rm unsup}  \left( \widehat{\mtx{U}}_{k} ; \mathcal{S}_{j} \right) }\right\}_{j=k}^{d}  } \right) \triangleq 
\nonumber \\ 
& \frac{ \sum_{j=k+1}^{d} { \mathbb{I} \left\{ {  \mathcal{E}_{\rm out} ^{\rm unsup}  \left( \widehat{\mtx{U}}_{k} ; \mathcal{S}_{j} \right) - \mathcal{E}_{\rm out} ^{\rm unsup}  \left( \widehat{\mtx{U}}_{k} ; \mathcal{S}_{j-1} \right) \le 0 } \right\}} }{ d-k }
\end{align}
where $\mathbb{I} \lbrace \cdot \rbrace$ is an indicator function returning $1$ if the condition is applied on is true and $0$ otherwise. Essentially, the metric (\ref{eq:metric for monotonic decrease evaluation}) summarizes the monotonicity of an entire error curve into a single value in the range $\left[0,1\right]$. An error curve with $\eta \left( { \left\{ { \mathcal{E}_{\rm out} ^{\rm unsup}  \left( \widehat{\mtx{U}}_{k} ; \mathcal{S}_{j} \right) }\right\}_{j=k}^{d}  } \right) = 1$ is monotonic decreasing over the entire range of $p$.

In Fig.~\ref{fig_appendix:all_monotonicity_metric_curve} we exhibit the values of the monotonicity metric for a variety of settings, including subspaces in the Hadamard and random forms (note that the horizontal axes represent the dimension of the subspace estimate). 
Each subfigure includes two curves: the dotted red curves present the monotonicity metric values induced by individual sequential constructions of $\mathcal{S}_p$;  the solid blue curves show the monotonicity metric values obtained for curves of errors \textit{averaged} over 500 experiments differing in their sequential constructions of $\mathcal{S}_p$.
Clearly, \text{specific} orders of adding features do not necessarily yield error curves that are purely monotonically decreasing. However, the \textit{averaged} error curves are monotonic decreasing over the entire range of $p$ (and this is the case for any $k$; see blue-colored curves in Fig. \ref{fig_appendix:all_monotonicity_metric_curve}). We take the results of these and numerous similar simulations with other parameter settings as strong experimental evidence that, on average,  $\mathcal{E}_{\rm out} ^{\rm unsup}  \left( \widehat{\mtx{U}}_{k} ; \mathcal{S}_{p} \right)$ decays with the increase in $p$.

\section{Proofs and Additional Details for Section 4}
\label{appsec:Proofs and Additional Details for Section 4}

\subsection{On the Singular Values of Rectangular, Tall Matrices with Orthonormal Columns}
\label{appsec:On the Singular Values of Rectangular, Tall Matrices with Orthonormal Columns}
A tall, rectangular matrix $\mtx{W} \in\mathbb{R}^{p\times m}$ (where $p\ge m$) has orthonormal columns if and only if all
of its singular values equal 1. This is proved next. 

Consider a real matrix $\mtx{W} \in\mathbb{R}^{p\times m}$ (where $p\ge m$) with orthonormal columns. Then, the corresponding SVD is $\mtx{W}=\mtx{\Omega} \mtx{\Sigma} \mtx{\Theta}^{T}$, where $\mtx{\Omega}$ and $\mtx{\Theta}$ are $p\times p$ and $m\times m$ real orthonormal matrices, respectively, and $\mtx{\Sigma}$ is a $p\times m$ real diagonal matrix with $m$ singular values $\left\{ { \sigma_{i}\left( \mtx{W} \right) }\right\}_{i=1}^{m}$ on its main diagonal. Since $\mtx{W}$ has orthonormal columns, we can write $\mtx{W}^T \mtx{W} = \mtx{I}_m$. Using the SVD form we get that 
\begin{equation}
    \mtx{I}_m = \left( \mtx{\Omega} \mtx{\Sigma} \mtx{\Theta}^{T} \right)^T \mtx{\Omega} \mtx{\Sigma} \mtx{\Theta}^{T} = \mtx{\Theta} \mtx{\Sigma}^T \mtx{\Sigma} \mtx{\Theta}^T
\end{equation}
and this can be translated into 
\begin{equation}
\label{eq:singular values and orthonormal columns - condition 1}
    \mtx{I}_m = \mtx{\Sigma}^T \mtx{\Sigma}.
\end{equation}
Since singular values are, by definition, non-negative real values, then Eq. (\ref{eq:singular values and orthonormal columns - condition 1}) implies that ${\sigma_{i}\left( \mtx{W} \right) = 1}$ for ${i=1,\dots,m}$. This proves the left-to-right direction of the statement. 

The second direction is proved as follows. 
Consider a real matrix $\mtx{W} \in\mathbb{R}^{p\times m}$ (where $p\ge m$) with SVD $\mtx{W}=\mtx{\Omega} \mtx{\Sigma} \mtx{\Theta}^{T}$, where $\mtx{\Omega}$ and $\mtx{\Theta}$ are $p\times p$ and $m\times m$ real orthonormal matrices, respectively, and $\mtx{\Sigma}$ is a $p\times m$ real diagonal matrix with $m$ singular values ${\sigma_{i}\left( \mtx{W} \right) = 1}$ for ${i=1,\dots,m}$ on its main diagonal. This means that $\mtx{\Sigma}^T \mtx{\Sigma} = \mtx{I}_m$. Then, 
\begin{equation}
\begin{split}
    \mtx{W}^T \mtx{W}  =  \left( \mtx{\Omega} \mtx{\Sigma} \mtx{\Theta}^{T} \right)^T \mtx{\Omega} \mtx{\Sigma} \mtx{\Theta}^{T} 
    \\
    =  \mtx{\Theta} \mtx{\Sigma}^T \mtx{\Sigma} \mtx{\Theta}^T 
    = \mtx{\Theta} \mtx{\Theta}^T 
    = \mtx{I}_m
\end{split}
\end{equation}
implying that the columns of $\mtx{W}$ are orthonormal. This completes the proof of the entire statement.

\subsection{The Hard Orthonormality-Constraints Projection Operator $T_{\rm hard}$}
\label{appsec:Proof for the Structure of the Hard Constraint Projection Operator}

The operator projecting onto the hard orthonormality constraints was defined in Section 4.1 as follows. 
Consider a matrix $\mtx{W}^{(\text{\rm in} )}\in\mathbb{R}^{p\times m}$ (where $p\ge m$), with the SVD $\mtx{W}^{(\text{\rm in} )}=\mtx{\Omega} \mtx{\Sigma}^{({\rm in})} \mtx{\Theta}^{T}$, where $\mtx{\Omega}$ and $\mtx{\Theta}$ are $p\times p$ and $m\times m$ real orthonormal matrices, respectively, and $\mtx{\Sigma}^{({\rm in})}$ is a $p\times m$ real diagonal matrix with $m$ singular values $\left\{ { \sigma_{i}\left( \mtx{W}^{(\text{\rm in} )} \right) }\right\}_{i=1}^{m}$ on its main diagonal. Then, projecting $\mtx{W}^{(\text{\rm in} )}$ onto the hard-orthonormality constraint via
\begin{equation} 
\label{eq:appendix - hard ortho constraint projection - optimization problem form}
{\mtx{W}^{(\text{\rm out} )}} = \argmin_{\mtx{W}\in\mathbb{R}^{p\times m}:~ \mtx{W}^T \mtx{W} = \mtx{I}_{m}} \left \Vert  \mtx{W} - {\mtx{W}^{(\text{\rm in} )}} \right \Vert _F^2
\end{equation}
induces the mapping  ${\mtx{W}^{(\text{\rm out} )}}\triangleq T_{\text{hard}}\left({\mtx{W}^{(\text{\rm in} )}}\right)$, where ${\mtx{W}^{(\text{\rm out} )}}=\mtx{\Omega} \mtx{\Sigma}^{({\rm out})} \mtx{\Theta}^{T}$ and the singular values along the main diagonal of $\mtx{\Sigma}^{({\rm out})}$ are ${\sigma_{i}\left( \mtx{W}^{(\text{\rm out} )} \right)=1}$ for ${i=1,\dots,m}$. 
A relevant proof is available in \cite{Kahan11} and also in a more general form in \cite{keller1975closest}.

\subsection{The Soft Orthonormality-Constraints Projection Operator $T_{\alpha}$}
\label{appsec:Proof for the Structure of the Soft Constraint Projection Operator}

The projection of a given matrix $\mtx{W}^{(\text{\rm in} )}\in\mathbb{R}^{p\times m}$ (where $p\ge m$) was defined in the main paper (see Eq. (16)) as follows. Consider the SVD $\mtx{W}^{(\text{\rm in} )}=\mtx{\Omega} \mtx{\Sigma}^{({\rm in})} \mtx{\Theta}^{T}$, where $\mtx{\Omega}$ and $\mtx{\Theta}$ are $p\times p$ and $m\times m$ real orthonormal matrices, respectively, and $\mtx{\Sigma}^{({\rm in})}$ is a $p\times m$ real diagonal matrix with $m$ singular values $\left\{ { \sigma_{i}\left( \mtx{W}^{(\text{\rm in} )} \right) }\right\}_{i=1}^{m}$ on its main diagonal. Then, the projection of $\mtx{W}^{(\text{\rm in} )}$ on the soft-orthonormality constraints is defined in its basic form as 
\begin{align} 
\label{eq:appendix-soft ortho constraint projection - optimization problem form}
    &{\mtx{W}^{(\text{\rm out} )}} = \argmin_{\mtx{W}\in\mathbb{R}^{p\times m}} \left \Vert  \mtx{W} - {\mtx{W}^{(\text{\rm in} )}} \right \Vert _F^2
    \\ \nonumber
    &\text{subject to}~~\lvert{  \sigma_{i}^{2}\left( \mtx{W} \right) - 1 }\rvert \le \alpha ~~\text{for }i=1,...,m
\end{align}
is equivalent to the thresholding mapping  ${{\mtx{W}^{(\text{\rm out} )}}\triangleq T_{\alpha}\left({\mtx{W}^{(\text{\rm in} )}}\right)}$ where ${\mtx{W}^{(\text{\rm out} )}}=\mtx{\Omega} \mtx{\Sigma}^{({\rm out})} \mtx{\Theta}^{T}$ and the singular values along the main diagonal of $\mtx{\Sigma}^{({\rm out})}$ are 
\begin{align}
\label{eq:appendix - linear subspace fitting optimization - supervised with soft constraints - singular values thresholding}
    &\sigma_{i}\left( \mtx{W}^{(\text{\rm out} )} \right)=
    \\
    &\begin{cases}
      \sigma_{i}\left( \mtx{W}^{(\text{\rm in} )} \right), \qquad \text{if}\ \sigma_{i}\left( \mtx{W}^{(\text{\rm in} )} \right) \in \left[\tau_{\alpha}^{\rm low},\tau_{\alpha}^{\rm high}\right] 
      \\
       {\tau_{\alpha}^{\rm low}} , ~~~\qquad\qquad \text{if}\ {\sigma_{i}\left( \mtx{W}^{(\text{\rm in} )} \right)} <  {\tau_{\alpha}^{\rm low}} 
      \\
      {\tau_{\alpha}^{\rm high}} , ~~\qquad\qquad \text{if}\ {\sigma_{i}\left( \mtx{W}^{(\text{\rm in} )} \right)} >  {\tau_{\alpha}^{\rm high}}
    \end{cases} \nonumber
\end{align}
for $i=1,...,m$, where the threshold levels are defined by $\tau_{\alpha}^{\rm low} \triangleq {\sqrt{\max{\left\{ {0, 1-\alpha} \right\} }}} $ and $\tau_{\alpha}^{\rm high} \triangleq {\sqrt{1+\alpha}} $. Also recall that singular values are non-negative by their definition. 

The relation between (\ref{eq:appendix - linear subspace fitting optimization - supervised with soft constraints - singular values thresholding}) and (\ref{eq:appendix-soft ortho constraint projection - optimization problem form}) is based on the extension of the case of strict orthonormality constraints that was explained above and proved in \cite{Kahan11}.

\subsection{The Algorithm for Supervised Subspace Fitting with Soft Orthonormality Constraints}
\label{appsec:The Algorithm for Supervised Subspace Fitting with Soft Orthonormality Constraints} 

We present here the explicit form of the method proposed in Section 4.3 for supervised subspace fitting with soft orthonormality constraints, i.e., the numerical procedure to address the problem in (15). We utilize the projected gradient descent technique to obtain the procedure outlined in Algorithm \ref{alg:appendix - Supervised Subspace Fitting via Projected Gradient Descent - Soft Constraints}.

Similar to Algorithm 1, we initialize the iterative process by setting $\mtx{W}^{(i=0)}$ by projecting the closed-form solution of the \textit{unconstrained} supervised problem onto the orthonormality constraint of interest (here using the operator $T_{\alpha}$). The gradient step size $\mu$ is updated in each iteration based on a simple line search mechanism that scales the former step size by finding the best within a set of update factors. This line search approach was also used in the implementation of Algorithm 1. 

One can also implement the proposed Algorithms without the line search mechanism and instead set a fixed gradient step size based on the worst case gradient direction induced by the quadratic cost functions examined in this paper.

\begin{algorithm}[t]
   \caption{Supervised Subspace Fitting via Projected Gradient Descent: \textbf{Soft} Orthonormality Constraints}
   \label{alg:appendix - Supervised Subspace Fitting via Projected Gradient Descent - Soft Constraints}
\begin{algorithmic}
   \STATE {\bfseries Input:} a dataset $\mathcal{D}^{\rm sup} _{\mathcal{S}} = \left \{ \left( \vec{x}^{(\ell)}_{\mathcal{S}},\vec{z}^{(\ell)} \right) \right\}_{\ell=1}^n$, a coordinate subset $\mathcal{S}$, and a threshold level $\alpha \ge 0$
   \STATE {\bfseries Initialize} $\mtx{W}^{(t=0)}= T_{\alpha}\left( \left( \mtx{Z} \mtx{X}_{\mathcal{S}}^{+} \right)^T \right)$, $t=0$
   \REPEAT
   \STATE $t \leftarrow t+1$
   \STATE $\mtx{Y}^{(t)} = \mtx{W}^{(t-1)} - \mu \mtx{X}_{\mathcal{S}} \left(  \left(\mtx{W}^{(t-1)}\right)^T \mtx{X}_{\mathcal{S}} - \mtx{Z} \right)^T$
   \STATE $\mtx{W}^{(t)} = T_{\alpha}\left( { \mtx{Y}^{(t)} } \right) $ 
   \UNTIL{stopping criterion is satisfied}
   \STATE Set $\widehat{\mtx{U}}_{m,\mathcal{S}} = \mtx{W}^{(t)}$
   \STATE Create $\widehat{\mtx{U}}_{m}$ based on $\widehat{\mtx{U}}_{m,\mathcal{S}}$ and zeros at rows corresponding to $\mathcal{S}_c$
   \STATE {\bfseries Output:} $\widehat{\mtx{U}}_{m}$
   
\end{algorithmic}
\end{algorithm}

\subsection{Additional Details on the Experiments in Section 4 (Supervised Settings)}
\label{appsubsec:Additional Details on the Experiments in Section 4}

In Section 4 of the main paper we present fully-supervised subspace fitting problems that are categorized into three types: strict orthonormally constrained (Section 4.1), unconstrained (essentially, a regression problem form, see Section 4.2), and soft orthonotmally constrained (Section 4.3). The empirical measurements of the out-of-sample errors of the various supervised settings are provided together in Fig.~3b (in the main text). We here elaborate on the settings of the experiments presented in Fig.~3. 

Since the problems are supervised, then the dimension $m$ of the true subspace is known. Accordingly, the results are only for estimation of $m$-dimensional representations. As usual, the data model is based on (1). Here the dimension of the entire space is $d=64$, the true subspace dimension is $m=20$, the number of examples is $n=32$, and the noise in the model corresponds to $\sigma_{\epsilon}=0.5$. Each of the curves in Fig.~3b presents the values $\mathcal{E}_{\rm out} ^{\rm unsup}  \left( \widehat{\mtx{U}}_{k} ; \mathcal{S}_{p} \right)$ versus $p$, which is the number of features used for the actual learning. The increase in $p$ refers to a sequential extension of $\mathcal{S}_{p}$ to include additional coordinates of features to be utilized. 

\begin{figure*}[]
\centering
\subfloat[]{\includegraphics[width=0.8\columnwidth]{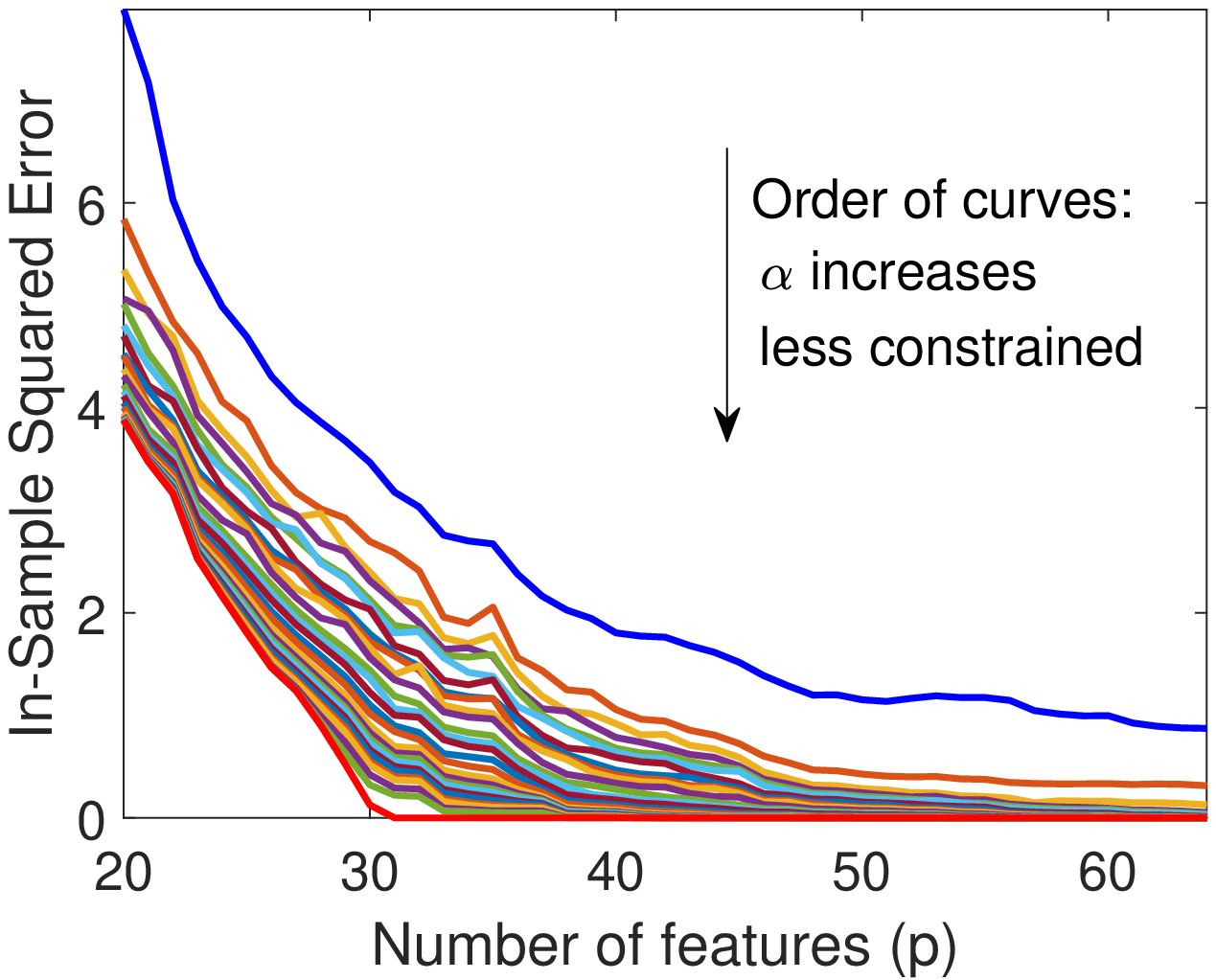}}
\subfloat[]{\includegraphics[width=0.8\columnwidth]{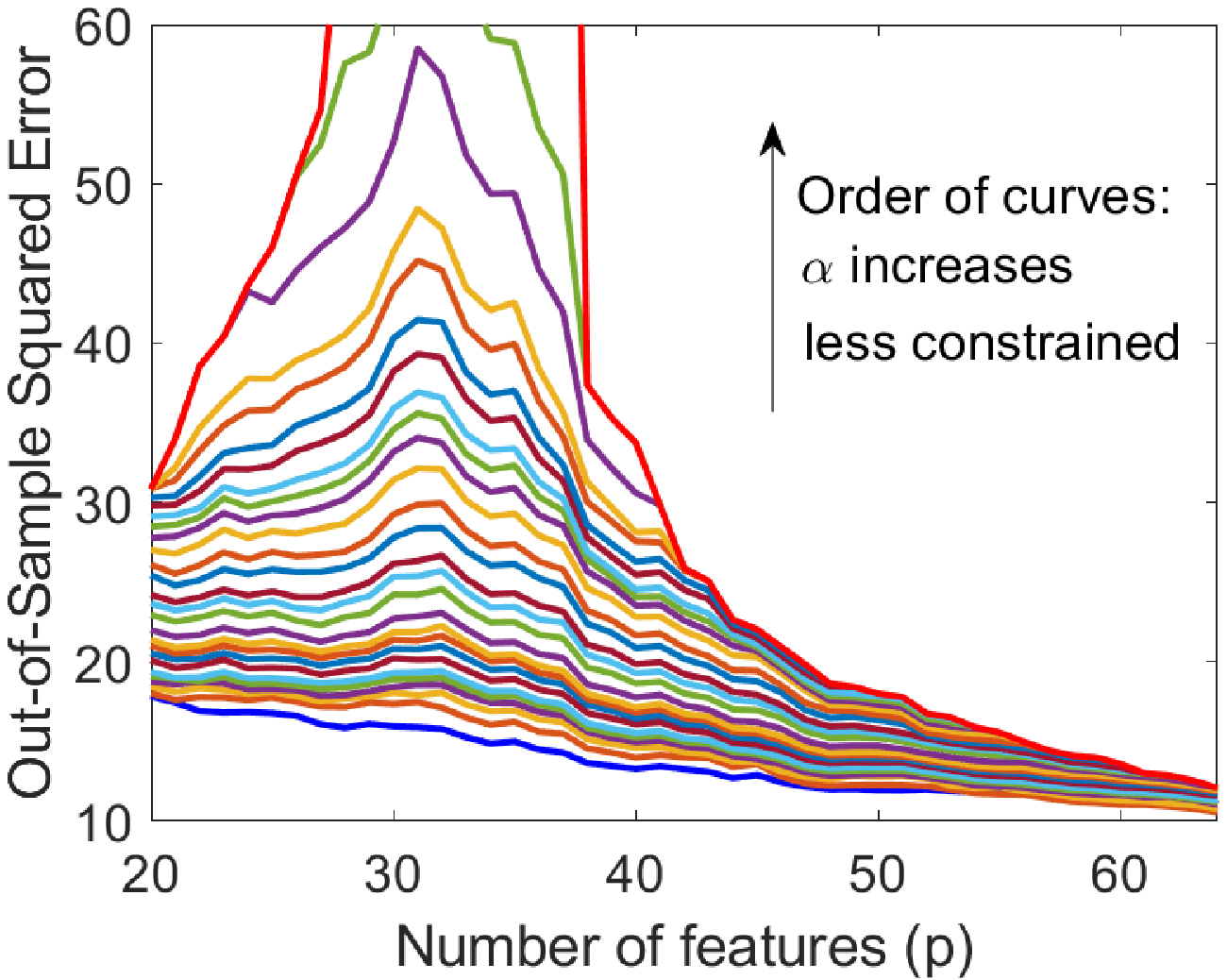}\label{fig_appendix:fully_supervise_varying_soft_ortho_constraints__out_of_sample_curves}}
\caption{The \textbf{(a) in-sample} errors $\mathcal{E}_{\rm in} ^{\rm sup}  \left( \widehat{\mtx{U}}_{m} \right)$ and \textbf{(b) out-of-sample} errors $\mathcal{E}_{\rm out} ^{\rm sup}  \left( \widehat{\mtx{U}}_{m} \right)$ of fully-supervised learning versus the number of parameters $p$. The errors correspond to a single sequential construction of $\mathcal{S}_{p}$. Here $d=64$, $m=20$, $n=32$, and $\sigma_{\epsilon}=0.5$. Each curve presents the results for a different level $\alpha$ of orthonormality constraints. The results here correspond to problems located along the yellow-colored border line in Fig.~1.}
\label{fig_appendix:fully_supervise_varying_soft_ortho_constraints}
\end{figure*}

The results in Fig.~3b present smooth curves by conducting the corresponding experiments 10 times with different sequential constructions of $\mathcal{S}_{p}$ and then averaging the induced errors.  
We present in Fig. \ref{fig_appendix:fully_supervise_varying_soft_ortho_constraints} the corresponding non-smooth curves by conducting these experiments for a specific (but arbitrary) order of adding features (i.e., without averaging over multiple experiments). 

Clearly, for the less orthonormally constrained settings (see the upper curves in Fig.~3b), the shape of the error curves resemble the double descent behavior, where the peak of each of these curves is obtained for $p=n-1$ (the minus 1 is due to the centering of the $n$ examples given). Importantly, after reaching the peak values, the out-of-sample errors start to decrease as the number of features increases and eventually achieving significantly lower error values than in the underparameterized range (i.e., for $p<n-1$). This exemplifies the benefits of overparameterization in subspace fitting problems that are fully supervised and may have soft orthonormality constraints. 

The settings that are nearly or (completely) orthonormally constrained (see the lower curves in Fig.~3b) present trends of decrease over the entire range of $p$. This may resemble the results presented above for unsupervised and strictly constrained subspace fitting. While these errors do not follow the double descent trend, they do exhibit the benefits of overparameterization even when the problem includes strict (or nearly strict) orthonormality constraints.

\section{Additional Details for Section 5: The Algorithm for Semi-Supervised Subspace Fitting}
\label{appsec:Additional Details for Section 5}

Section 5 established an approach for semi-supervised subspace fitting with a flexible level of orthonormality constraints. The basic optimization problem is presented in (18) and does not have a closed-form solution. The following extends the details provided in the main text about the numerical procedure for addressing (18) using the projected gradient descent technique. Recall that in this semi-supervised setting there are two datasets in use: a supervised set of examples $\widetilde{\mathcal{D}}^{\rm sup} _{\mathcal{S}} = \left \{ \left( \vec{x}^{(\ell)}_{\mathcal{S}},\vec{z}^{(\ell)} \right) \right\}_{\ell=1}^{n^{\rm sup}}$, and an unsupervised set of examples $\widetilde{\mathcal{D}}^{\rm unsup} _{\mathcal{S}} = \left \{  \vec{x}^{(\ell)}_{\mathcal{S}}  \right\}_{\ell=n^{\rm sup}+1}^{n}$.

The proposed method is presented in Algorithm \ref{alg:appendix - Semi-Supervised Subspace Fitting via Projected Gradient Descent}.
In contrast to Algorithms 1 and \ref{alg:appendix - Supervised Subspace Fitting via Projected Gradient Descent - Soft Constraints} that address fully supervised settings, in the semi-supervised case the evolving solution is initialized to a random matrix, which contains i.i.d.~Gaussian components with zero mean and variance $1/p$, that is projected onto the relevant orthonormality constraint (via the operator $T_{\alpha}$ that for $\alpha=0$ is equivalent to $T_{\rm hard}$). 
The data from the unsupervised examples, $ \mtx{X}_{\mathcal{S}}^{\rm unsup}  \triangleq \left[ \vec{x}^{(n^{\rm sup}  + 1)}_{\mathcal{S}} , \dots, \vec{x}^{(n)}_{\mathcal{S}} \right]$, is used in conjunction with the supervised examples in the gradient descent steps throughout the iterations of the algorithm. 

Since (18) extends (15) only with respect to the optimization cost, then Algorithm \ref{alg:appendix - Semi-Supervised Subspace Fitting via Projected Gradient Descent} simply extends Algorithm \ref{alg:appendix - Supervised Subspace Fitting via Projected Gradient Descent - Soft Constraints} by updating the gradient used in the descent stage of the $t^{\rm th}$ iteration with 
\begin{align} 
\label{eq:linear subspace fitting optimization - semi-supervised gradient descent step}
& G^{\rm semisup}\left( {\mtx{W}^{(t)} } \right) \triangleq \mtx{X}_{\mathcal{S}}^{\rm sup}  \left(  \left(\mtx{W}^{(t)}\right)^T  \mtx{X}_{\mathcal{S}}^{\rm sup}  - \mtx{Z}^{\rm sup} \right)^T 
\nonumber \\
&  - 2\mtx{X}_{\mathcal{S}}^{\rm unsup}  \left( \mtx{X}_{\mathcal{S}}^{\rm unsup}  \right)^T \mtx{W}^{(t)} 
\nonumber \\
& + \mtx{X}_{\mathcal{S}}^{\rm unsup} \left(\mtx{X}_{\mathcal{S}}^{\rm unsup} \right)^T \mtx{W}^{(t)} \left(\mtx{W}^{(t)}\right)^T \mtx{W}^{(t)} 
\nonumber \\
& + \mtx{W}^{(t)} \left(\mtx{W}^{(t)}\right)^T \mtx{X}_{\mathcal{S}}^{\rm unsup} \left(\mtx{X}_{\mathcal{S}}^{\rm unsup} \right)^T  \mtx{W}^{(t)} 
\end{align}
that was obtained by differentiation of the semi-supervised cost function of (18), ${\left \Vert  \mtx{Z}^{\rm sup}  - \mtx{W}^T \mtx{X}_{\mathcal{S}}^{\rm sup}   \right \Vert _F^2 + \left \Vert  \left( {\mtx{I}_{p} - \mtx{W} \mtx{W}^T } \right) \mtx{X}_{\mathcal{S}}^{\rm unsup}  \right \Vert _F^2}$, with respect to $\mtx{W}$.

The gradient step size $\mu$ is updated in each iteration based on a simple line search approach that was described above for Algorithm \ref{alg:appendix - Supervised Subspace Fitting via Projected Gradient Descent - Soft Constraints}.

\begin{algorithm}[t]
   \caption{Semi-Supervised Subspace Fitting via Projected Gradient Descent (Soft Orthonormality Constraints)}
   \label{alg:appendix - Semi-Supervised Subspace Fitting via Projected Gradient Descent}
\begin{algorithmic}
   \STATE {\bfseries Input:} datasets $\widetilde{\mathcal{D}}^{\rm sup} _{\mathcal{S}} = \left \{ \left( \vec{x}^{(\ell)}_{\mathcal{S}},\vec{z}^{(\ell)} \right) \right\}_{\ell=1}^{n^{\rm sup}}$ and $\widetilde{\mathcal{D}}^{\rm unsup} _{\mathcal{S}} = \left \{  \vec{x}^{(\ell)}_{\mathcal{S}}  \right\}_{\ell=n^{\rm sup}+1}^{n}$, a coordinate subset $\mathcal{S}$, and a threshold level $\alpha \ge 0$
   \STATE {\bfseries Initialize} $\mtx{W}^{(t=0)}= T_{\alpha}\left( \mtx{H} \right)$ where $\mtx{H}$ is a $p\times m$ random Gaussian matrix of i.i.d.~components $\mathcal{N}(0,1/p)$, $t=0$
   \REPEAT
   \STATE $t \leftarrow t+1$
   \STATE $\mtx{Y}^{(t)} = \mtx{W}^{(t-1)} - \mu\cdot G^{\rm semisup}\left( {\mtx{W}^{(t-1)} } \right)$
   \STATE $\mtx{W}^{(t)} = T_{\alpha}\left( { \mtx{Y}^{(t)} } \right) $ 
   \UNTIL{stopping criterion is satisfied}
   \STATE Set $\widehat{\mtx{U}}_{m,\mathcal{S}} = \mtx{W}^{(t)}$
   \STATE Create $\widehat{\mtx{U}}_{m}$ based on $\widehat{\mtx{U}}_{m,\mathcal{S}}$ and zeros at rows corresponding to $\mathcal{S}_c$
   \STATE {\bfseries Output:} $\widehat{\mtx{U}}_{m}$
   
\end{algorithmic}
\end{algorithm}

\begin{figure*}[]
\centering
\subfloat[]{\includegraphics[width=0.8\columnwidth]{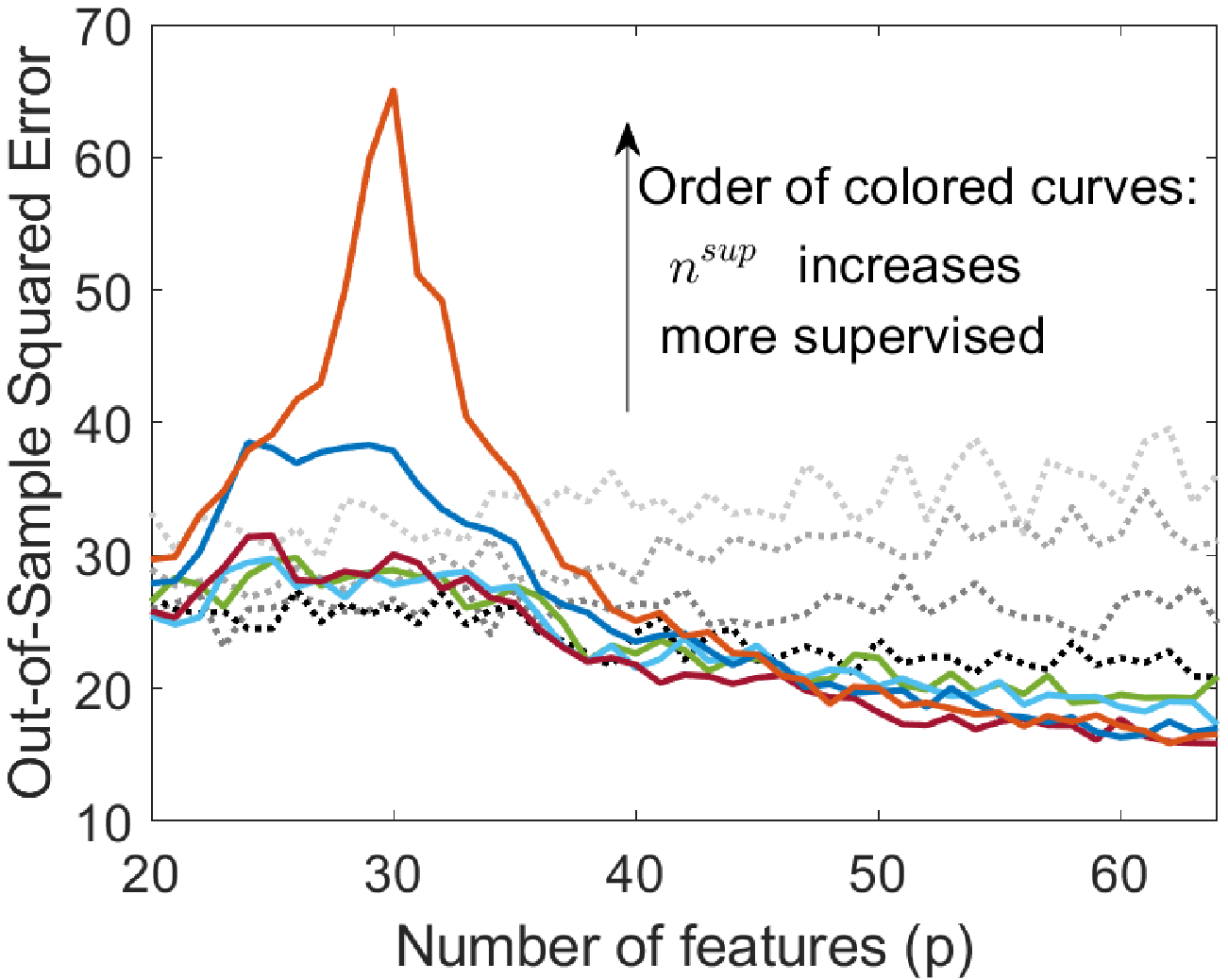}\label{fig:semi supervised with soft ortho constraints - out-of-sample errors}}
\subfloat[]{\includegraphics[width=0.8\columnwidth]{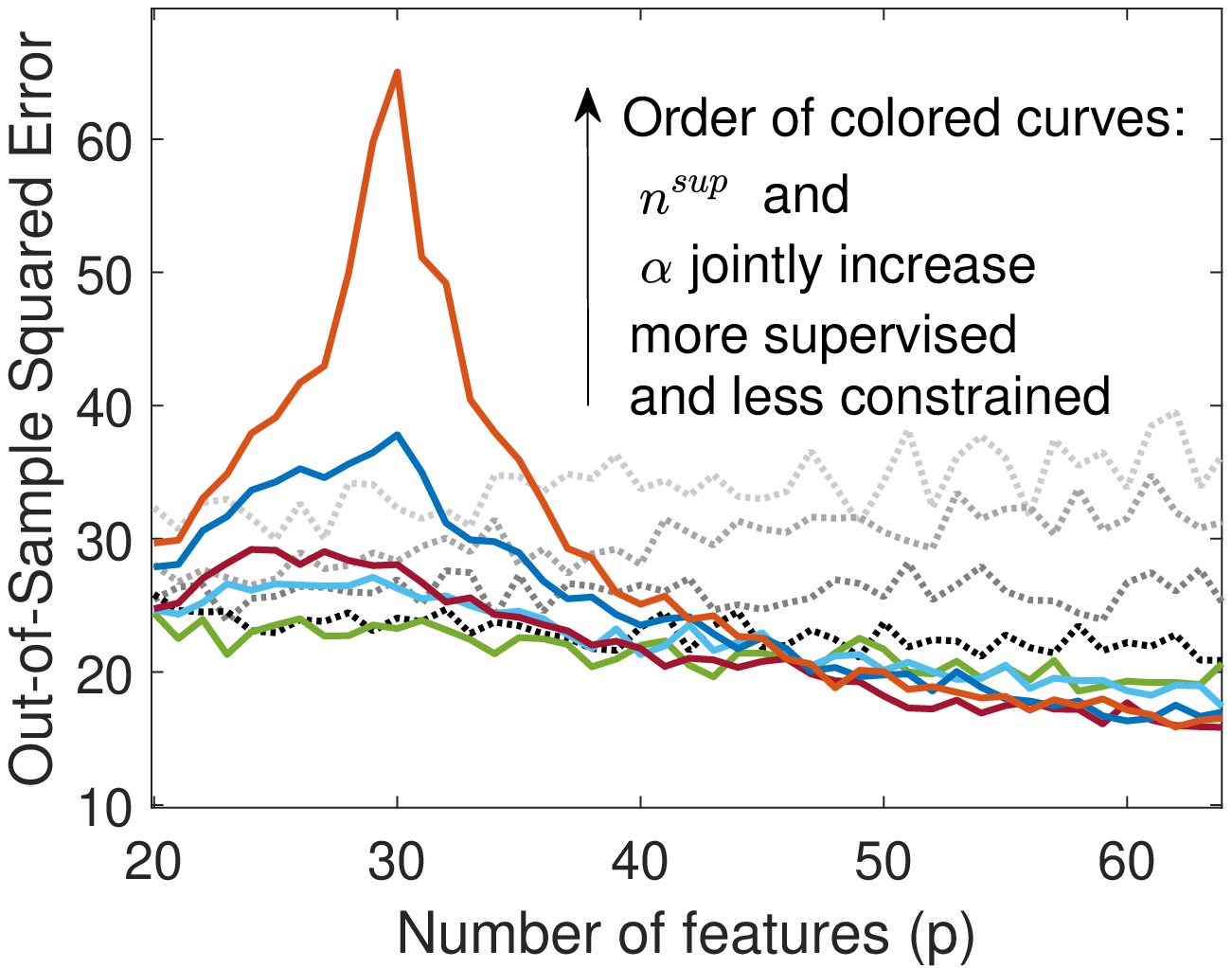}\label{fig:diagonal direction - out-of-sample errors}}
\caption{The out-of-sample errors, $\mathcal{E}_{\rm out} ^{\rm sup}  \left( \widehat{\mtx{U}}_{m} \right)$ versus the number of parameters $p$. The errors correspond to a single experiment with a single sequential order of adding coordinates to $\mathcal{S}$. Here $d=64$, $m=20$ and $n=32$. 
\textbf{(a) Unconstrained} settings ($\alpha\rightarrow \infty$):  Each curve presents the results for a different supervision level, $n^{\rm sup} \in \left\{ {0,4,8,12, 16,20,24,28,n=32} \right\}$.
\textbf{(b)} Problems residing at the supervision-orthonormality plane along the \textbf{diagonal trajectory} connecting the standard subspace fitting and the pure regression. Each curve presents the results for a different pair of supervision and orthonormality constraint levels that jointly increase.
In both subfigures, the gray dotted curves correspond to $n^{\rm sup} \in \left\{ {0,4,8,12} \right\}$.}
\end{figure*}

The error curves in Figures \ref{fig:semi supervised with soft ortho constraints - out-of-sample errors - AVERAGE} and \ref{fig:diagonal direction - out-of-sample errors - AVERAGE} are smooth due to averaging over 25 experiments with different sequential orders of adding coordinates to $\mathcal{S}$. In Figures \ref{fig:semi supervised with soft ortho constraints - out-of-sample errors} and \ref{fig:diagonal direction - out-of-sample errors} we provide the corresponding error curves obtained from a single experiment (i.e., for a single order of adding coordinates to $\mathcal{S}$). 
Note that Fig.~\ref{fig:diagonal direction - out-of-sample errors - AVERAGE} considers settings with soft orthonormality constraints (at various levels) that reduce the error levels compared to the corresponding unconstrained settings presented in Fig~\ref{fig:semi supervised with soft ortho constraints - out-of-sample errors - AVERAGE}.

\section{Unsupervised Subspace Fitting with Soft Orthonormality Constraints}
\label{appsec:Unsupervised Subspace Fitting with Soft Orthonormality Constraints}

In Section 3.1 we defined the unsupervised form of the linear subspace fitting problem that included a strict orthonormality constraint and solved it via PCA.  
Now, we can define the corresponding range of unsupervised problems with flexible levels of orthonormality constraints, namely,  
\begin{align} 
\label{eq:appendix - linear subspace fitting optimization - unsupervised with soft constraints}
    &\widehat{\mtx{U}}_{m,\mathcal{S}} = \argmin_{\mtx{W}\in\mathbb{R}^{p\times m}}  \left \Vert  \left( {\mtx{I}_{p} - \mtx{W} \mtx{W}^T } \right) \mtx{X}_{\mathcal{S}}  \right \Vert _F^2
    \nonumber\\ 
    &\text{subject to} ~\lvert{  \sigma_{i}^{2}\left( \mtx{W} \right) - 1 }\rvert \le \alpha ~~\text{for }i=1,...,m, 
\end{align}
where we assume that $m$ is known, $ \mtx{X}_{\mathcal{S}} \triangleq \left[ \vec{x}^{(1)}_{\mathcal{S}} , \dots, \vec{x}^{(n )}_{\mathcal{S}} \right]$ is the data matrix corresponding to the (unsupervised) dataset that was considered in Section 3, and $\alpha$ determines the orthonormality constraint level. The optimization cost in (\ref{eq:appendix - linear subspace fitting optimization - unsupervised with soft constraints}) reflects the unsupervised aspect of the problem.  
We address (\ref{eq:appendix - linear subspace fitting optimization - unsupervised with soft constraints}) using the projected gradient descent method and get the process described in Algorithm \ref{alg:appendix - Unsupervised Subspace Fitting via Projected Gradient Descent}. As before, the soft orthonormality constraints induce a projection stage that uses the soft-threshold projection $T_{\alpha}$ from (\ref{eq:appendix - linear subspace fitting optimization - supervised with soft constraints - singular values thresholding}). Importantly, unlike (\ref{eq:appendix - linear subspace fitting optimization - supervised with soft constraints - singular values thresholding}) we set the lower threshold to $\tau_{\alpha}^{\rm low} \triangleq {\sqrt{\max{\left\{ {10^{-16}, 1-\alpha} \right\} }}} $ that avoids clipping of singular values to zero, and the upper threshold remains the same, i.e., $\tau_{\alpha}^{\rm high} \triangleq {\sqrt{1+\alpha}} $.
Avoiding clipping the singular values to zero is important for maintaining the full rank of the evolving solution matrix throughout the (projected) gradient descent process. Unlike  the supervised and semi-supervised settings, we empirically found that avoiding clipping singular values to zero is a crucial aspect in the unsupervised problems when optimized via projected gradient descent. 
The gradient descent step (in the $t^{\rm th}$ iteration) is based on the gradient of the unsupervised cost of (\ref{eq:appendix - linear subspace fitting optimization - unsupervised with soft constraints}), i.e., 
\begin{align} 
\label{eq:linear subspace fitting optimization - unsupervised gradient descent step}
G^{\rm unsup}\left( {\mtx{W}^{(t)} } \right) & \triangleq   - 2\mtx{X}_{\mathcal{S}}   \mtx{X}_{\mathcal{S}}^T \mtx{W}^{(t)} 
\nonumber \\
& + \mtx{X}_{\mathcal{S}} \mtx{X}_{\mathcal{S}}^T  \mtx{W}^{(t)} \left(\mtx{W}^{(t)}\right)^T \mtx{W}^{(t)} 
\nonumber \\
& + \mtx{W}^{(t)} \left(\mtx{W}^{(t)}\right)^T \mtx{X}_{\mathcal{S}} \mtx{X}_{\mathcal{S}}^T  \mtx{W}^{(t)} . 
\end{align}
Note that due to the unsupervised form of the problem we cannot initialize the process using the unconstrained linear regression solution (as we did in the Algorithms developed above for the fully supervised settings with soft orthonormality constraints). Therefore, the initialization in Algorithm \ref{alg:appendix - Unsupervised Subspace Fitting via Projected Gradient Descent} sets $\mtx{W}^{(i=0)}$ to a $p\times m$ matrix with i.i.d.~Gaussian entries $\mathcal{N}(0,1/p)$. 
\begin{algorithm}[t]
   \caption{Unsupervised Subspace Fitting via Projected Gradient Descent (Soft Orthonormality Constraints)}
   \label{alg:appendix - Unsupervised Subspace Fitting via Projected Gradient Descent}
\begin{algorithmic}
   \STATE {\bfseries Input:} a dataset  ${\mathcal{D}}_{\mathcal{S}} = \left \{ \vec{x}^{(\ell)}_{\mathcal{S}}  \right\}_{\ell=1}^{n}$, a coordinate subset $\mathcal{S}$, $m$, and a threshold level $\alpha \ge 0$
   \STATE {\bfseries Initialize} $\mtx{W}^{(t=0)}= T_{\alpha}\left( \mtx{H} \right)$ where $\mtx{H}$ is a $p\times m$ random Gaussian matrix of i.i.d.~components $\mathcal{N}(0,1/p)$, $t=0$
   \REPEAT
   \STATE $t \leftarrow t+1$
   \STATE $\mtx{Y}^{(t)} = \mtx{W}^{(t-1)} - \mu\cdot G^{\rm unsup}\left( {\mtx{W}^{(t-1)} } \right)$
   \STATE $\mtx{W}^{(t)} = T_{\alpha}\left( { \mtx{Y}^{(t)} } \right) $ 
   \UNTIL{stopping criterion is satisfied}
   \STATE Set $\widehat{\mtx{U}}_{m,\mathcal{S}} = \mtx{W}^{(t)}$
   \STATE Create $\widehat{\mtx{U}}_{m}$ based on $\widehat{\mtx{U}}_{m,\mathcal{S}}$ and zeros at rows corresponding to $\mathcal{S}_c$
   \STATE {\bfseries Output:} $\widehat{\mtx{U}}_{m}$
   
\end{algorithmic}
\end{algorithm}

The empirical results obtained using Algorithm \ref{alg:appendix - Unsupervised Subspace Fitting via Projected Gradient Descent} for a range of $\alpha$ values from zero (strictly constrained) to infinity (unconstrained) showed that all the respective solutions accurately follow the PCA solution obtained for the unsupervised problem with a strict orthonormality constraint (i.e., the solution obtained in Section 3 for $k=m$).


\bibliography{subspace_fitting_references}

\begin{thebibliography}{23}
\providecommand{\natexlab}[1]{#1}
\providecommand{\url}[1]{\texttt{#1}}
\expandafter\ifx\csname urlstyle\endcsname\relax
  \providecommand{\doi}[1]{doi: #1}\else
  \providecommand{\doi}{doi: \begingroup \urlstyle{rm}\Url}\fi

\bibitem[Belkin et~al.(2019{\natexlab{a}})Belkin, Hsu, Ma, and
  Mandal]{belkin2019reconciling}
Belkin, M., Hsu, D., Ma, S., and Mandal, S.
\newblock Reconciling modern machine-learning practice and the classical
  bias--variance trade-off.
\newblock \emph{Proceedings of the National Academy of Sciences}, 116\penalty0
  (32):\penalty0 15849--15854, 2019{\natexlab{a}}.

\bibitem[Belkin et~al.(2019{\natexlab{b}})Belkin, Hsu, and Xu]{belkin2019two}
Belkin, M., Hsu, D., and Xu, J.
\newblock Two models of double descent for weak features.
\newblock \emph{arXiv preprint arXiv:1903.07571}, 2019{\natexlab{b}}.

\bibitem[Breiman \& Freedman(1983)Breiman and Freedman]{breiman1983many}
Breiman, L. and Freedman, D.
\newblock How many variables should be entered in a regression equation?
\newblock \emph{Journal of the American Statistical Association}, 78\penalty0
  (381):\penalty0 131--136, 1983.

\bibitem[Denton et~al.(2019)Denton, Parke, Tao, and
  Zhang]{denton2019eigenvectors}
Denton, P.~B., Parke, S.~J., Tao, T., and Zhang, X.
\newblock Eigenvectors from eigenvalues: {A} survey of a basic identity in
  linear algebra.
\newblock \emph{arXiv preprint arXiv:1908.03795}, 2019.

\bibitem[Geiger et~al.(2019)Geiger, Jacot, Spigler, Gabriel, Sagun, d'Ascoli,
  Biroli, Hongler, and Wyart]{geiger2019scaling}
Geiger, M., Jacot, A., Spigler, S., Gabriel, F., Sagun, L., d'Ascoli, S.,
  Biroli, G., Hongler, C., and Wyart, M.
\newblock Scaling description of generalization with number of parameters in
  deep learning.
\newblock \emph{arXiv preprint arXiv:1901.01608}, 2019.

\bibitem[Gower et~al.(2004)Gower, Dijksterhuis, et~al.]{gower2004procrustes}
Gower, J.~C., Dijksterhuis, G.~B., et~al.
\newblock \emph{Procrustes {P}roblems}, volume~30.
\newblock Oxford University Press on Demand, 2004.

\bibitem[Hastie et~al.(2019)Hastie, Montanari, Rosset, and
  Tibshirani]{hastie2019surprises}
Hastie, T., Montanari, A., Rosset, S., and Tibshirani, R.~J.
\newblock Surprises in high-dimensional ridgeless least squares interpolation.
\newblock \emph{arXiv preprint arXiv:1903.08560}, 2019.

\bibitem[Hwang(2004)]{hwang2004cauchy}
Hwang, S.-G.
\newblock Cauchy's interlace theorem for eigenvalues of {H}ermitian matrices.
\newblock \emph{The American Mathematical Monthly}, 111\penalty0 (2):\penalty0
  157--159, 2004.

\bibitem[Johnstone \& Lu(2009)Johnstone and Lu]{johnstone2009consistency}
Johnstone, I.~M. and Lu, A.~Y.
\newblock On consistency and sparsity for principal components analysis in high
  dimensions.
\newblock \emph{Journal of the American Statistical Association}, 104\penalty0
  (486):\penalty0 682--693, 2009.

\bibitem[Jolliffe(1972)]{jolliffe1972discarding}
Jolliffe, I.~T.
\newblock Discarding variables in a principal component analysis. i: Artificial
  data.
\newblock \emph{Journal of the Royal Statistical Society: Series C (Applied
  Statistics)}, 21\penalty0 (2):\penalty0 160--173, 1972.

\bibitem[Jolliffe(1973)]{jolliffe1973discarding}
Jolliffe, I.~T.
\newblock Discarding variables in a principal component analysis. ii: Real
  data.
\newblock \emph{Journal of the Royal Statistical Society: Series C (Applied
  Statistics)}, 22\penalty0 (1):\penalty0 21--31, 1973.

\bibitem[Kahan(2011)]{Kahan11}
Kahan, W.
\newblock The nearest orthogonal or unitary matrix, August 2011.
\newblock "URL:
  \url{https://people.eecs.berkeley.edu/~wkahan/Math128/NearestQ.pdf}. Last
  visited on 2020/02/06".

\bibitem[Keller(1975)]{keller1975closest}
Keller, J.~B.
\newblock Closest unitary, orthogonal and {H}ermitian operators to a given
  operator.
\newblock \emph{Mathematics Magazine}, 48\penalty0 (4):\penalty0 192--197,
  1975.

\bibitem[Mei \& Montanari(2019)Mei and Montanari]{mei2019generalization}
Mei, S. and Montanari, A.
\newblock The generalization error of random features regression: Precise
  asymptotics and double descent curve.
\newblock \emph{arXiv preprint arXiv:1908.05355}, 2019.

\bibitem[Nie et~al.(2010)Nie, Xu, Tsang, and Zhang]{nie2010flexible}
Nie, F., Xu, D., Tsang, I. W.-H., and Zhang, C.
\newblock Flexible manifold embedding: A framework for semi-supervised and
  unsupervised dimension reduction.
\newblock \emph{IEEE Transactions on Image Processing}, 19\penalty0
  (7):\penalty0 1921--1932, 2010.

\bibitem[Paul(2007)]{paul2007asymptotics}
Paul, D.
\newblock Asymptotics of sample eigenstructure for a large dimensional spiked
  covariance model.
\newblock \emph{Statistica Sinica}, pp.\  1617--1642, 2007.

\bibitem[Shen et~al.(2016)Shen, Shen, and Marron]{shen2016general}
Shen, D., Shen, H., and Marron, J.
\newblock A general framework for consistency of principal component analysis.
\newblock \emph{The Journal of Machine Learning Research}, 17\penalty0
  (1):\penalty0 5218--5251, 2016.

\bibitem[Spigler et~al.(2018)Spigler, Geiger, d'Ascoli, Sagun, Biroli, and
  Wyart]{spigler2018jamming}
Spigler, S., Geiger, M., d'Ascoli, S., Sagun, L., Biroli, G., and Wyart, M.
\newblock A jamming transition from under-to over-parametrization affects loss
  landscape and generalization.
\newblock \emph{arXiv preprint arXiv:1810.09665}, 2018.

\bibitem[Sugiyama(2006)]{sugiyama2006local}
Sugiyama, M.
\newblock Local fisher discriminant analysis for supervised dimensionality
  reduction.
\newblock In \emph{Proceedings of the 23rd International Conference on Machine
  Learning}, pp.\  905--912, 2006.

\bibitem[Ulfarsson \& Solo(2011)Ulfarsson and Solo]{ulfarsson2011vector}
Ulfarsson, M.~O. and Solo, V.
\newblock Vector $ l\_0 $ sparse variable {PCA}.
\newblock \emph{IEEE Transactions on Signal Processing}, 59\penalty0
  (5):\penalty0 1949--1958, 2011.

\bibitem[Xu \& Hsu(2019)Xu and Hsu]{xu2019number}
Xu, J. and Hsu, D.~J.
\newblock On the number of variables to use in principal component regression.
\newblock In \emph{Advances in Neural Information Processing Systems}, pp.\
  5095--5104, 2019.

\bibitem[Yang et~al.(2006)Yang, Fu, Zha, and Barlow]{yang2006semi}
Yang, X., Fu, H., Zha, H., and Barlow, J.
\newblock Semi-supervised nonlinear dimensionality reduction.
\newblock In \emph{Proceedings of the 23rd International Conference on Machine
  Learning}, pp.\  1065--1072, 2006.

\bibitem[Zhang et~al.(2007)Zhang, Zhou, and Chen]{zhang2007semi}
Zhang, D., Zhou, Z.-H., and Chen, S.
\newblock Semi-supervised dimensionality reduction.
\newblock In \emph{Proceedings of the 2007 {SIAM} International Conference on
  Data Mining}, pp.\  629--634. SIAM, 2007.

\end{thebibliography}
\bibliographystyle{icml2020}

\end{document}